\documentclass[11pt]{article}
\usepackage{multirow}
\newcommand{\secondbest}[1]{\textcolor{blue}{#1}}

\usepackage{microtype}
\usepackage{graphicx}
\usepackage{subcaption}
\usepackage{booktabs} % for professional tables

\usepackage{hyperref}
\usepackage{times}

\usepackage[T1]{fontenc} 

\usepackage[preprint]{acl}

\usepackage{amsmath}
\usepackage{amssymb}
\usepackage{mathtools}
\usepackage{amsthm}

\usepackage[capitalize,noabbrev]{cleveref}
\usepackage{xspace}

\usepackage{amsmath}
\usepackage{amssymb}
\usepackage{mathtools}
\usepackage{amsthm}

\usepackage[capitalize,noabbrev]{cleveref}

\usepackage{hyperref}
\usepackage{url}

\usepackage{packages/boxmath}
\DeclareMathOperator{\LSE}{LSE}

\usepackage{subcaption}

\usepackage{algorithm}
\usepackage{algorithmic}
\usepackage{multirow}
\usepackage{enumitem}
\usepackage{arydshln}
\usepackage{wrapfig}
\usepackage{booktabs}
\usepackage{graphicx}
\usepackage[normalem]{ulem}
\useunder{\uline}{\ul}{}
\theoremstyle{plain}

\theoremstyle{definition}

\theoremstyle{remark}

\usepackage[textsize=tiny]{todonotes}

\newcommand{\method}{\textsc{Prompt2Box}\xspace}
\newcommand{\VisMethod}{\textsc{Box-SNE}\xspace}
\newcommand{\heading}[1]{\vspace{5pt}\noindent\underline{\textsc{#1}}}

\title{\method: Improving LLM Weakness Discovery and Specificity Estimation by Uncovering Entailment Structure among Prompts}

\author{
  \textbf{Neeladri Bhuiya\textsuperscript{1,3}
    \thanks{The work was primarily done while both authors were at UMass Amherst.}}
  \quad
  \textbf{Shib Sankar Dasgupta\textsuperscript{2}\footnotemark[1]}
  \quad
  \textbf{Andrew McCallum\textsuperscript{1}}
  \quad
  \textbf{Haw-Shiuan Chang\textsuperscript{1}}
  \\
  \textsuperscript{1}CICS, University of Massachusetts Amherst, USA
  \\
  \textsuperscript{2}Amazon AWS AI
  \qquad
  \textsuperscript{3}A10 Networks
  \\
  \texttt{nbhuiya@a10networks.com, shibdg@amazon.com,}
  \\
  \texttt{\{mccallum,hschang\}@cs.umass.edu}
}

\begin{document}

\maketitle

\begin{abstract}

To discover the weaknesses of LLMs, researchers often embed prompts into a vector space and cluster them to extract insightful patterns. However, vector embeddings primarily capture topical similarity; as a result, prompts that share a topic but differ in specificity, and consequently in difficulty, are often represented similarly, making fine-grained weakness analysis difficult. To address this limitation, we propose \method, which embeds prompts into a box embedding space using a trained encoder. The encoder, trained on existing and synthesized datasets, outputs box embeddings that capture not only semantic similarity but also specificity relations between prompts (e.g., ``\textit{writing an adventure story}'' is more specific than ``\textit{writing a story}''). We further develop a novel dimension reduction technique for box embeddings to facilitate dataset visualization and comparison. Our experiments demonstrate that box embeddings consistently capture prompt specificity better than vector baselines and achieve 45\% error reduction on average in predicting specificity compared to the prompt length baseline. On the downstream task of creating hierarchical clustering trees for 17 LLMs from the UltraFeedback dataset, \method can identify 13.5\% more LLM weaknesses than vector baselines and achieves an approximately 33\% stronger correlation between hierarchical depth and instruction specificity. The code is available at \url{https://github.com/zawedcvg/box_embeddings}.

% \todo{add the specificity result in the last sentence}

\end{abstract}

\section{Introduction}
\label{sec:intro}

\begin{figure}[t!]
  \vskip 0.2in
  \begin{center}
    \centerline{\includegraphics[width=\columnwidth]{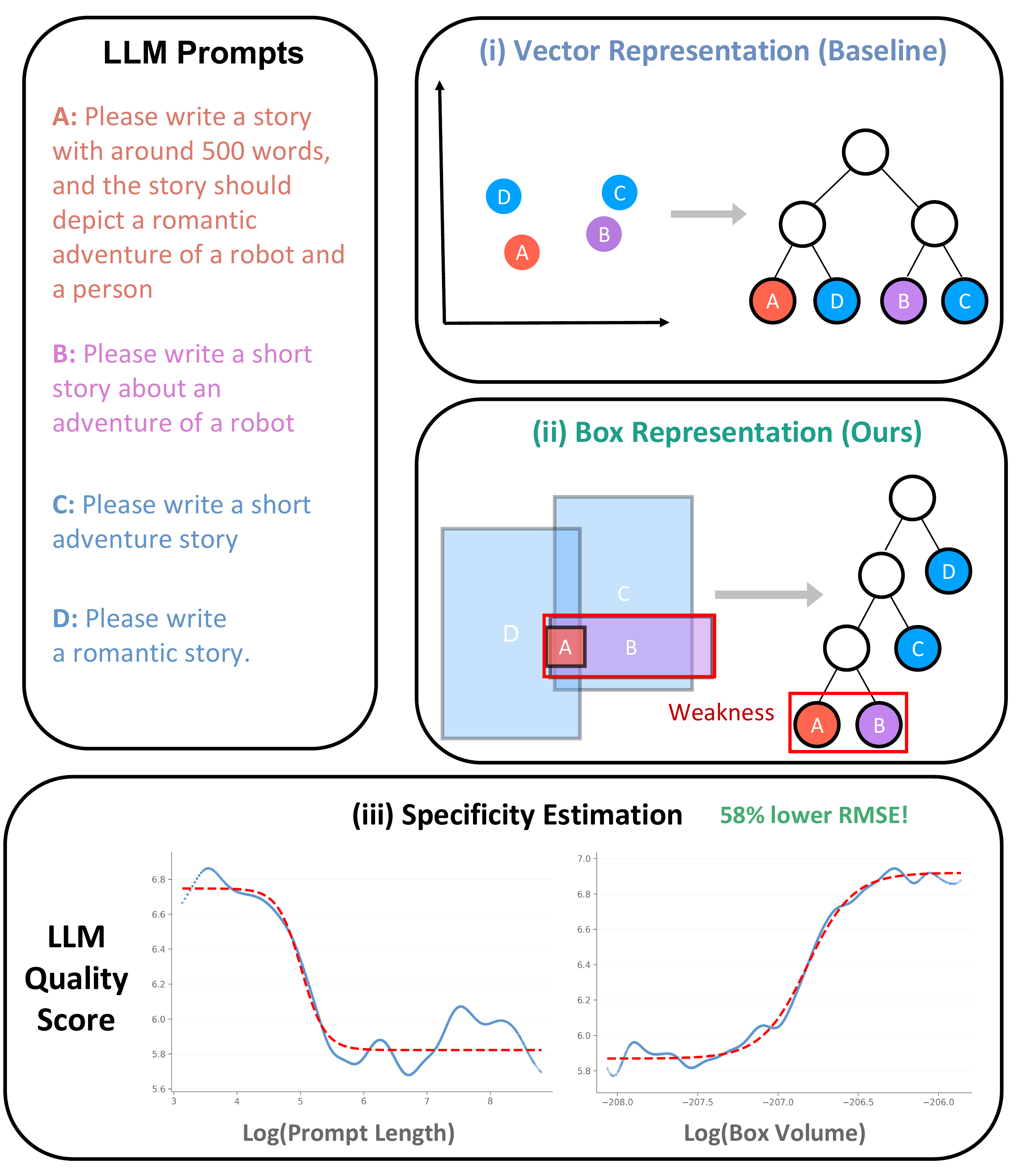}}
    \caption{Comparison between the widely-used vector representation in (i) and our box representation in (ii) for analyzing the performance of an LLM on four prompts. Blue means that the LLM achieves a high performance on the prompt while red means the opposite. Our approach correctly highlights "writing robot-adventure stories" as a weakness of the LLM by clustering prompt \textit{A} and \textit{B}. In (iii), we visualize the scores of LLama-2-13b-Chat's responses versus specificity estimates based on length and box volume }
    \label{First_fig}
  \end{center}
\end{figure}

For large language model (LLM) developers, it is crucial to identify an LLM's weaknesses to guide the next round of collection of additional high-quality training data. When developers observe a bad response from an LLM to a prompt, there are two main possible causes: 1) the LLM has a weakness in that area or topic, or 2) the prompt itself is very specific, which makes LLM unable to extrapolate from its training data. This induces two challenging questions: 1) how to discover those weakness topics, and 2) how to measure specificity. 

Several prior works~\citep{jiang-etal-2024-followbench,tamkin2024clio,zeng2025evaltree,tian2025skillverse} address the first challenge by embedding every prompt into a vector and hierarchically clustering the prompts based on vector similarity. By analyzing performance differences across these clusters in a two-dimensional projection of the embedding space, LLM developers can diagnose systematic weaknesses relative to competing models. However, the vector-based methods rely on the assumption that LLMs perform similarly for similar prompts, without accounting prompt specificity or difficulty. Consequently, vectors can group two prompts that are topically similar but have different levels of difficulty together. Ignoring prompt specificity loses an entire mode of failure, making it difficult to attribute the actual reason of a weakness.

%it unclear whether the LLM performs poorly on the underlying topic in general, or whether it struggles only with the more specific or difficult prompts within that cluster.
%brings undesired ambiguities for interpreting the LLMs' performance using the vector-based approach. For example, when the LLM achieves a low average score for a prompt cluster, it is unclear whether the LLM performs poorly on the underlying topic in general, or whether it struggles only with the more specific or difficult prompts within that cluster.

Recent studies~\citep{sun2024conifer,atmakuru2024cs4,lu2025benchmarking,jaroslawicz2025many,zhang2025cfbench} demonstrate that adding constraints to a prompt reduces the solution space, making the prompt more specific and more difficult for an LLM to answer.\footnote{Notice that we do not consider the under-specified prompts as in \citet{kim2025detail,zi2025more}.} However, there is no standard method to measure prompt specificity. Although the simple length baseline often highly correlates with specificity, \citet{li2015fast,gao2019predicting,kapur2026more} show that there is usually room of improvements in each particular application.

\Cref{First_fig} (i) illustrates a limitation of vector-based methods. Prompt \textit{A} is semantically similar to prompt \textit{D}, but considerably more specific. In a hierarchical clustering based on vector embeddings, the low score of \textit{A} drags down the average score of the cluster containing both \textit{A} and \textit{D}, causing that cluster to be flagged as a weakness of the LLM. However, this conclusion is misleading: since the cluster broadly captures the ability to \textit{"write a romantic story"}, the high score of \textit{D} already demonstrates that the LLM is competent at this general task. The poor performance on \textit{A} reflects a failure on a more specific variant instead of the broader capability that the cluster represents. This example underscores a key limitation of relying solely on vector similarity for LLM weakness analysis.

To address this problem, we propose \method, which embeds each prompt into a high-dimensional box embedding space. Unlike vector embeddings, box embeddings~\citep{Boxlattice} can naturally represent asymmetric semantic relationships like entailment, making them well-suited for modeling hierarchical structure among prompts. 
%Box embeddings, as introduced in \citet{Boxlattice}, represent concepts as hyper-rectangles and have been shown to effectively capture such asymmetries.
Conceptually, each textual prompt is represented by a box parameterized by a center vector and a size vector. The center vector captures the semantic location of the prompt, such that semantically similar prompts are mapped to nearby centers. The size vector controls the semantic scope of the prompt: more general prompts are represented by larger boxes, while more specific prompts correspond to smaller boxes. This geometry allows entailment relationships to be expressed through box containment.

Specifically, if the box of one prompt is contained within the box of another prompt, we interpret this as an entailment relation, where the more specific prompt entails the more general one. For example, in \Cref{First_fig} (ii), box \textit{A} is almost entirely contained within box \textit{D}, reflecting that the prompt ``\textit{writing a romantic adventure story}'' (prompt \textit{A}) semantically entails the more general prompt ``\textit{writing a romantic story}'' (prompt \textit{D}).

The box of a prompt can also be interpreted as the space of its valid responses. In this interpretation, strong performance for a prompt implies the existence of a high-quality response in this solution space that can be produced by this LLM. For example, the LLM does worse on prompts \textit{A} and \textit{B} than for prompt \textit{D}, which suggests that the LLM is not good at ``\textit{writing an adventure story about a robot}'', especially when the adventure involves some \textit{romantic} elements, but this LLM is probably good at writing other kinds of \textit{romantic stories}. Our box representation demonstrates that the weakness of the LLM lies in the region of box \textit{A} and \textit{B}. Vectors, on the other hand, cannot support such conclusions because of their lack of specificity information.

% we propose \method, which embeds each prompt into a box space. 

%We show that the box of a prompt could also be interpreted as the space of its valid responses. In this interpretation, a good performance of an LLM for a prompt means that there exists a good response in this box that could be outputted by this LLM, but the LLM is still likely to output some other bad responses in this box. For example, the LLM does badly for prompt \textit{A} and \textit{B} but not for prompt \textit{D}, which suggests that the LLM is not good at ``\textit{writing an adventure story of a robot}'', especially when the adventure involves some \textit{romantic} relations, but this LLM is probably good at writing other kinds of \textit{romantic stories}. Our box representation demonstrates that the weakness of the LLM lies on the region of box \textit{A} and \textit{B}, while we cannot get the similar conclusion from the vector baseline because of its lack of specificity information.

%well for prompt \textit{D} but not for prompt \textit{A} in \Cref{First_fig}, which suggests that the LLM 
%box D  (ii)

%Not doing well in portraying an adventure
%the LLM outputs a good point inside the box, but it does not mean the LLM .
%An LLM performs well for a prompt

% box 
% entailment
% constraint subset
% solution subset

% For example

% LLM does worse for the prompt A, which means LLM is not good at writing romantic adventure of a robot and a person rather than writing a romantic story

%This bring the ambiguity of the interpretation of the 

%\hs{experiments}
To discover the entailment structure among prompts, we leverage existing entailment datasets and synthesize entailment relations between prompts. Next, we train an encoder to map every prompt into a box. Furthermore, we propose a new dimension reduction method and a new hierarchical clustering method for box embeddings. Our experiments show that our box embeddings predict the entailment relations much better than the vector baselines, which allows us to better analyze the weaknesses of LLMs through a 2D box embedding space and our specificity-aware hierarchical clustering method. Furthermore, even though the model is never trained to predict specificity of a prompt, we find that the box volume is a better specificity estimator than the length of the prompt. \Cref{First_fig} (iii) shows that LLM scores are much more correlated with box volume than prompt lengths. %longer prompts do not necessarily lead to lower LLM scores, while the score decay much more smoothly as the box volume decreases.

%existing entailment datasets
%synthesize 
%Compare LLMs, and datasets
%introduce two applications
%better hierarchical cluster and identify 

%\subsection{Main Contributions}
\heading{Main Contributions}
\begin{itemize}[itemsep=2pt, parsep=0pt, topsep=5pt]
    \item We propose \method, which uses a box embedding-based representation to capture the entailment relation among prompts. We also propose novel methods to synthesize entailment datasets for training an encoder that maps each LLM prompt to a box embedding.
    \item We propose \VisMethod, a novel dimensionality-reduction method for box embeddings, and a new hierarchical clustering algorithm for box embeddings.
    \item We propose new evaluation metrics to assess similarity, entailment, and specificity in prompt representations. The experiments demonstrate large improvements of \method over strong baselines and how box embeddings can be used to analyze LLMs and LLM evaluation benchmarks.
    %\hs{todo}
\end{itemize}

%as we add more constraints into a prompt, the prompt would become more difficult for LLMs because the number of valid .

%vector, cluster and analyze the tree or the 2D projection of the embedding space

%In the LLM development process
%In industry, we often need to know the weakness of LLMs and try to fix them. To know what training data we should prepare for training and testing LLMs. 

% LLM Evaluation
% LLM weakness finding

% In traditional vector-based representation, each prompt is only a point in the space. 
%It assumes that the LLMs tend to have similar performance in similar prompts.
% only capture the similarity. So very specific/difficult prompts might be close to general/easy prompts.

% Also, we are not sure if one does not do well in a prompt because the LLM has a weakness in this topic or this prompt is very difficult

% structure

% the quality degrades as the prompts become more specific (cite CS4 and the paper it cites).
% this is because there are many possible responses that satisfy the constraints in the prompt and LM could output the best response from them. 

% For example, 
%  and LLM might have seen some  training

\section{Related Work}
%% Box
%% Include original work (vilnis et al) then talk about gumbel box (dasgupta etal) say we are using this version of box. Different task that got help by asymmetric relation modeling: KG, Task hierarchy, Word2Box, Recommendation systems. Various domains prove that boxes are better at capturing asymmetry hierarchical 
Box embeddings \cite{Boxlattice}, a form of region-based embeddings, have been shown to outperform other region-based representations such as Order Embeddings \cite{order_embedding} and Poincaré Embeddings \cite{poincare} in modeling asymmetric relationships~\cite{NEURIPS2021_88d25099}.
%Despite their effectiveness, training box embeddings is challenging due to gradient flattening arising from their piecewise-constant, region-based geometry. To address this issue, \cite{gumbel_box} proposes a Gumbel-distribution-based smoothing of the box landscape, which enables stable gradient-based optimization. We adopt this parameterization in our work. 
Box embeddings have been successfully applied to model hierarchical and structured semantic relationships across multiple domains. In computer vision, \citet{daroya2024task2box} used box embeddings to represent task-level hierarchies. In the context of knowledge bases, box embeddings effectively capture hierarchical graph structures such as WordNet \cite{akbc, box-to-box} and OWL ontologies \cite{owl-ontology}. Furthermore, \citet{query2box, box-to-box} introduce box-embedding-based formulations for knowledge graph query answering, where the logical structure of a query is directly encoded in the embedding space. As far as we know, no prior work uses box embeddings to analyze prompts or LLMs' weaknesses.

As identifying LLMs' weaknesses has become increasingly important, more and more benchmark/prompt analysis methods are being proposed. Examples include Clio~\citep{tamkin2024clio}, SkillVerse~\citep{tian2025skillverse}, and EvalTree~\citep{zeng2025evaltree}. Moreover, many recent studies leverage LLMs to discover taxonomy and categories from a corpus~\citep{hsu2024chime,tian2024generic,zhang2025llmtaxo,kargupta2025taxoadapt,zhong2025hicode,gao2025science,chirkova2025llm}. Although different papers use different clustering methods or leverage LLMs in different ways, most of them conduct (hierarchical) clustering based on the vector embedding space. Our paper discovers that boxes perform better than vectors in terms of identifying LLM weakness clusters, and thus can potentially improve over the aforementioned related works. 

%LLM evaluation
%prompt analysis or weakness finding
%taxonomy and category discovery on text

%based on 

%In a similar spirit, our work applies box embeddings to the prompt space of large language models, using their geometric containment properties to model entailment and hierarchical relationships among prompts.

\section{Method}
\label{sec:method}
 
\begin{figure*}[t!]
  %\vskip 0.2in
  \begin{center}
    \centerline{\includegraphics[width=1\textwidth]{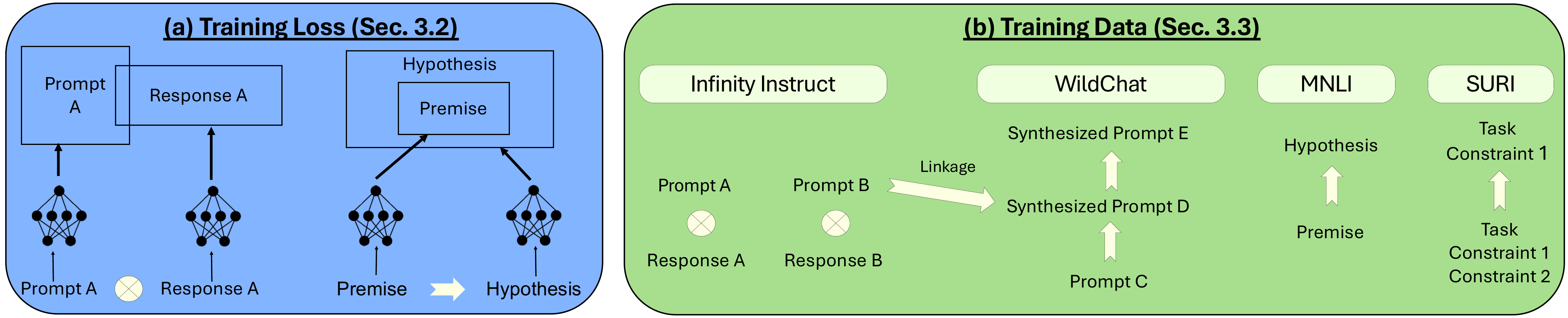}}
    \caption{Illustration of our encoder training method. White $\Rightarrow$ means entailment and $\bigotimes$ means intersection. (a) An encoder is trained to take a prompt and output a box. Our loss function encourages its output box to overlap with the box of its corresponding response and to be contained by the box of the prompt it entails. (b) We use Infinity Instruct to encourage similar prompts to intersect with each other, and use WildChat, MNLI, and SURI to create positive and negative examples for learning entailment relationships between prompts.}
    \label{fig:Training}
  \end{center}
\end{figure*}

We first introduce our definition of entailment between prompts and establish the connection among constraint space, entailment, and solution space in \Cref{sec:def}. We describe the methods of optimizing our encoder in \Cref{sec:opt}. Finally, \Cref{sec:data_syn} explains how the training data are curated. We expand on how we parameterize box embeddings and compute the intersection size and entailment probability in Appendix~\ref{sec:box_rep}. 

%our box embedding space could mean 
%definition 
%box representation
%Optimization\Cref{sec:opt}
%Data Curation\Cref{sec:data_syn}

%\Cref{fig:Training}

%\hs{Our method is summarized at \Cref{fig:Training} (a) and (b)}

\subsection{Definition of Terms}
\label{sec:def}
Let $\mathcal X$ denote the space of instructions. In this paper, we focus on modeling both prompt similarity and prompt specificity, which we define as the number of constraints, both implicit and explicit, present in the instruction.

 %Let $\mathcal U$ denote the universe of all possible constraints. %For any instruction $x \in \mathcal X$, let $\mathcal C : \mathcal X \rightarrow 2^{\mathcal U}$
% be a mapping from an instruction to the set of constraints it induces, where
% \begin{equation}
% \small
% \mathcal C(x) = \{\, c \mid c\in \mathcal U \text{ is satisfied by all valid responses to } x \,\}.
% \end{equation}
%A prompt with a greater number of constraints is more specific than one with fewer constraints.
% \begin{equation}
% \text{$a$ is more specific than $b$} \quad\Longleftrightarrow\quad \mathcal C(a)\supset\mathcal C(b).
% \end{equation}
This notion of specificity closely mirrors the traditional logical concept of entailment. Recall that for any two statements $c$ and $d$, if $c$ entails $d$ (written $c \models d$), then whenever $c$ is true, $d$ must also be true. In other words, $c$ imposes a stronger condition than $d$; it is more specific. One can thus view specificity as a specialisation of entailment to the space $\mathcal{X}$. %Formally, constraint inclusion implies entailment:
% \begin{equation}
% \forall\, a,b \in \mathcal X,\qquad
% \mathcal C(a)\supset\mathcal C(b)
% \;\Longrightarrow\;
% a \models b .
% \end{equation}
% Taking the example present in Figure~\ref{First_fig}, we have two instructions: $a =$ ``Please write a short story about an adventure of a robot'' and $b =$ ``Please write a short adventure story.''  Since it is not possible to exhaustively list out all the set of possible constraints, one can informally say that C{b} belongs c(a) if c(a) can be written as b with additional constraint/constraints. We can thus write them as

% $\mathcal{C}(a)$: Please write a short adventure story; Make the story about a robot. \\

% We see that $\mathcal{C}(a) \supset \mathcal{C}(b)$ , thus based on definition a $a \models b$

% Let $\mathcal{Y}$ denote the universe of all possible solutions. For any instruction $x \in \mathcal{X}$, let $\mathcal{S} : \mathcal{X} \rightarrow 2^{\mathcal{Y}}$
% be a mapping from an instruction to the set of valid solutions that satisfy the constraints imposed by $x$, where $\mathcal{S}(x) \subseteq \mathcal{Y}$.

For a pair of prompts $a, b \in \mathcal{X}$, if $a$ contains more constraints than $b$, then any valid solution to $a$ must satisfy a stricter set of conditions. Consequently, the set of valid solutions for $a$ is smaller than that for $b$. % This establishes that inclusion in the constraint space induces reverse inclusion in the solution space:
% \begin{equation}
% \mathcal C(a)\supset\mathcal C(b)\quad\Longrightarrow\quad \mathcal S(a)\subset\mathcal S(b).
% \end{equation}
This perspective offers an explanation for the empirically observed degradation in LLM performance as the number of constraints\footnote{Note that we only consider non-redundant constraints} increases \citep{jiang-etal-2024-followbench, atmakuru2024cs4}. As constraints accumulate, the valid solution space contracts, increasing task difficulty by requiring the model to generate responses from a progressively smaller region of admissible outputs. Consequently, failures under highly constrained prompts expose systematic weaknesses in an LLM’s ability to jointly satisfy multiple interacting constraints.

\subsection{Optimization}
\label{sec:opt}

%\hs{talking about figures. training encoder. bi-directional encoder}
%\Cref{fig:Training}
\heading{Similarity and Entailment for boxes:}
Let the box representations of a and b be defined with $\mathrm{Box}(a)$ and $\mathrm{Box}(b)$. We define the volume of $\mathrm{Box}(a)$ as $\operatorname{Vol}(a)$. We model \emph{prompt similarity} as the volume of the intersection between their boxes, i.e., $\operatorname{VolInt(a, b)} = \operatorname{Vol}(\mathrm{Box}(a) \cap \mathrm{Box}(b))$.

%We model \emph{asymmetric interactions} between prompts using entailment between box embeddings. %In particular, when prompt $a$ entails prompt $b$, we expect $\mathrm{Box}(b)$ to contain $\mathrm{Box}(a)$. In this case, the intersection volume equals the volume of $\mathrm{Box}(a)$, i.e., $\operatorname{VolInt}(a,b) =
%\operatorname{Vol}(\mathrm{Box}(a))$.

\emph{Entailment score} is defined as the conditional probability
\begin{equation}
\label{eq:entailment}
p(b \mid a)
\coloneqq
\frac{
\operatorname{VolInt}(a, b)
}{
\operatorname{Vol}(\mathrm{Box}(a))
}
%\tag{ii}
\end{equation}

By construction, $p(b \mid a) = 1$ when $a$ fully entails $b$, and $p(b \mid a) < 1$ otherwise, providing a principled measure of prompt entailment.

\heading{Learnable Parameters.}
For each prompt $a$, the box embedding is parameterized by a center vector $a_{\text{center}} \in \mathbb{R}^D$ and a width vector $a_\delta \in \mathbb{R}_+^D$. These parameters are produced by passing the prompt embedding from a Sentence Transformer through two separate MLP heads. The Sentence Transformer and both MLPs are trained jointly.\footnote{MLP leads to a 2\% increase in the number of parameters over the baselines.}

\heading{Gumbel Box Formulation:}
Optimizing objectives involving hard $\min$ and $\max$ operators is challenging due to their non-differentiability \citep{softbox, gumbel_box, min-max-box}. We adopt the Gumbel Box formulation \citep{gumbel_box}, which replaces hard interval endpoints with Gumbel-distributed random variables and yields smooth, differentiable approximations to box intersection and containment. We present more details of the method in Appendix \ref{appendix:gumbel_specific}.

\heading{Contrastive Training Objective.}
We train the model using contrastive learning objectives for both prompt similarity and entailment. Positive and negative prompt pairs are constructed for symmetric similarity and asymmetric entailment relations as shown in \Cref{fig:Training}. We provide training details in the Appendix~\ref{sec:training_details}.

%We use the Multiple Negatives Loss~\cite{henderson2017efficientnaturallanguageresponse} boosted by GradCache~\citep{gao-etal-2021-scaling} to allow for a large batch size while training. Each batch only contains the training samples from one dataset and each dataset is selected according to its data size percentage and a round-robin scheduling.

%round-robin
%\hs{Kinda it is selected based on a round Robin fashion  based on the data percentage}

%To effectively train our box representations to capture both standard relevance and specificity, we carefully curate datasets that jointly support these two objectives.
%Specifically, we combine similarity based data to supervise the learning of box intersections, which model semantic relevance, with constraint-based instruction data and existing entailment datasets to train box volumes and containment relationships, which encode instruction specificity. While lemma a showed that specificity implies entailment and not otherwise, we see that including existing entailment datasets helps with performance.

%Our training data is constructed by combining existing instruction-response corpora with a synthesized hierarchical instruction dataset. The overall objective of the data curation process is to expose the model to hierarchical instruction structures, while ensuring relevance and semantic coherence across instruction levels.

\subsection{Data Curation}
\label{sec:data_syn}

Finding sufficient entailment data to train a box encoder is a very challenging task, which limits the adoption of box representation. Fortunately, the high accuracy and flexibility of recent LLMs make synthesizing entailment data at large scale feasible. \Cref{fig:Training} illustrates how we use synthesized and existing entailment datasets to train our box encoder and we will describe the curation steps for each dataset below.

%insufficient entailment data
%LLM 
%\hs{todo. Existing dataset and synthesize new data}
%\Cref{sec:infinite_inst}
%\Cref{sec:mnli}
%\Cref{sec:wildchat}
%\Cref{sec:suri}
%\Cref{sec:links}

\heading{Semantic Relevance:}
%\label{sec:infinite_inst}
To capture relevance, we gather instruction response pairs from Infinity Instruct~\citep{li2025infinityinstructscalinginstruction} and retain only the English samples. Our contrastive learning encourages similar prompts to overlap with each other by maximizing the intersection (i.e., $\operatorname{VolInt(a, b)}$) between the box of a prompt and the box of its response, while penalizing the other negatively sampled intersections ~\citep{Mikolov2013EfficientEO}. 

%We train box intersection using a contrastive learning objective. This is designed to encourage semantic alignment between instructions and their corresponding responses, while separating unrelated instruction–response pairs. \hs{This is done so that prompts with similar responses are indirectly brought close to each other similar to how word2vec is trained....}

\heading{Entailment Data from MultiNLI:}
\label{sec:mnli}

We leverage MultiNLI~\citep{N18-1101}, which contains sentence pairs labeled as entailment, contradiction, or neutral. We apply a preprocessing step to transform these pairs into triplets of (anchor, positive, negative) for contrastive learning. For each anchor sentence, the entailed hypothesis serves as the positive example, while hypotheses labeled as neutral or contradiction are used as negative examples, as they do not express an entailment relationship. This dataset allows the model to learn sentence-level entailment relationships between text pairs.

%In section~\ref{} we talk about specificity implying entailment, and not the other way round. However based on experiments, we see that adding entailment data helps improve performance

% We use data from MultiNLI to further enrich the entailment dataset. In section~\ref{} we talk about how specificity implies entailment, but not the other way around. However we find from experiments that entailment data helps improve the performance. We create transform the pairs dataset in MultiNLI 

\heading{Synthesized Hierarchy from WildChat:}
\label{sec:wildchat}
To learn the entailment relation among instructions, we synthesize hierarchical instructions on WildChat~\citep{zhao2024wildchat}. Specifically, we prompt OpenAI's GPT-4.1 to iteratively rewrite each instruction in WildChat into progressively more general forms, producing a hierarchy spanning multiple levels of specificity. We obtain 20K hierarchical instruction groups, each containing 4-10 levels. %This dataset teaches the model direct parent-child relationships across multiple levels of instruction specificity.

%We first filter the data by retaining only single turn interactions written in English and only include The instructions to those containing between 8 and 150 words; this range is chosen empirically to exclude trivial prompts and overly verbose instructions.

%To reduce redundancy, we compute sentence embeddings using all-mpnet-base-v2~\citep{NEURIPS2020_c3a690be} and remove instructions with cosine similarity greater than 0.9, eliminating near-duplicate or semantically equivalent prompts. The remaining instructions are then passed to GPT-4.1 using an in-context learning prompt (shown in Section XYZ), which generates multiple levels of general instructions for each prompt. This process produces hierarchical instruction trees with varying levels of specificity.

%We obtain 20,000 synthesized hierarchical instruction groups, each containing between 4 and 10 hierarchical levels.

\heading{Sibling Relationships from SURI:}
\label{sec:suri}
While the previous datasets teach direct parent-child relationships, they do not capture sibling relationships, cases where two instructions share a common parent but differ in their specific constraints, and thus do not entail one another. To address this, we leverage SURI~\citep{pham-etal-2024-suri}. Each datapoint in SURI consists of a main goal summarizing the original text, accompanied by approximately ten constraints covering stylistic and semantic elements. We construct instruction trees by combining the main goal with various subsets of constraints. Instructions sharing the same parent but with different constraint combinations are treated as sibling nodes and used as hard negatives in our contrastive learning objective. %as they should exhibit no entailment relationship. 
This 
%complements the previous parent-child entailment relationships by teaching 
teaches the model to distinguish between related but non-entailing instructions.

%To model hierarchical instruction structures, we leverage SURI~\citep{pham-etal-2024-suri}. SURI consists of human written long-form texts backtranslated into structured instructions. Each example instruction consists of a "Main Goal", an instruction summarizing the original text, followed by about ten constraints that focus on stylistic elements, semantic elements, or both.

%We construct hierarchical training instances by combining the main goal with subsets of its constraints, corresponding to different depths of an instruction tree. Higher levels of the hierarchy consist of the main goal with fewer constraints, while lower levels contain subsets with more constraints. Different combinations of constraints are treated as sibling nodes within the hierarchy as shown in figure.... (Here add a figure of main goal + different constraints and also the box representations of the same that we are trying to learn). We get a total of X.... instructions..? We treat the sibling nodes as hard negatives when learning for entailment.

\heading{Dataset Entailment Linkage:}
\label{sec:links}
After initially training with the above datasets, we observed that while the model performed well on entailment-based metrics%(see appendix~\ref{sec:eval}
, it exhibited a noticeable drop in semantic relevance. We hypothesize that the objectives associated with the different datasets are learnt separately, leading to a disconnect in the embedding space. To mitigate this issue, we explicitly connect Infinity Instruct to our synthesized hierarchical dataset using WildChat~\cite{zhao2024wildchat}. For each sampled query prompt from Infinity Instruct, we use MPNet-Base~\citep{NEURIPS2020_c3a690be} model to retrieve similar prompts from WildChat. We then ask GPT-4.1 to find the most specific synthesized prompt entailed by the query prompt.
The aim of this linkage is to force the model to learn a shared representation across the different objectives. %We see in table~\ref{tab:main_results
%~\citep{NEURIPS2020_c3a690be}

%We first use all\_mpnet\_base\_v2~\citep{NEURIPS2020_c3a690be} to search for top-k prompts similar to infinity instruct. Then using gpt 4.1 we find a good point to insert, with the prompts found in the previous step, such that the ensuing instruction group has the correct ordering based on specificity in ABC. This integration encourages shared representations across similarity and entailment based dataset, preventing the model from over-specializing to hierarchical structure at the expense of general relevance and prevents them from being disjoint features the model learns. The combined dataset thus balances hierarchical reasoning with robust instruction-following performance. We present the results of no-links vs otherwise in table ABC. We get 3138 datapoints with it.

%\subsection{Loss Function/Training}

%To train the box embeddings, we build on the Sentence Transformers framework [cite]. We utilize the framework’s custom similarity function interface to define our box-based similarity and entailment loss functions. This enables end-to-end training of box parameters using standard contrastive and ranking-based objectives, while replacing traditional vector similarity (e.g., cosine similarity) with geometry-aware box interactions.
\section{Embedding Space Evaluation}
\label{sec:exp_emb}
We initialize both the box-embedding model and the vector-based baseline from MPNet-base~\citep{NEURIPS2020_c3a690be}. Because vector embeddings cannot naturally represent entailment or partial orders, we treat all entailment relations as similarity for the default vector baseline. Concretely, instruction pairs that exhibit entailment are encouraged to have high cosine similarity, without imposing any directional or containment structure.

In contrast, the box model explicitly separates these notions: similarity is modeled via Eq.~\eqref{eq:box_similarity}, while entailment is captured through the containment-based objective in Eq.~\eqref{eq:entailment}.

Our training set comprises 203,138 samples drawn from the different sources with the following distribution: prompt-response pairs from Infinity Instruct (50K), SURI-based entailment dataset (50K), Synthetic hierarchical instructions from WildChat (50K), MNLI triplets (50K), and the linkage dataset (3,138).

For ablation, we additionally train box models without the linkage dataset (w/o links), as well as both box and vector models without the entailment datasets (w/o entails) (i.e., trained only on pairs from Infinity Instruct). To evaluate the effect of synthetic data, we also include a model trained without linkage dataset and hierarchical instructions from WildChat (w/o synth).

To compare against other embedding methods capable of modeling asymmetry, we include a model trained using the CSDelta metric~\citep{chang-etal-2018-distributional} as well as a hyperbolic embedding model based on squared Lorentzian distance~\citep{pmlr-v97-law19a}. Further implementation and training details are provided in the Appendix~\ref{sec:training_details}.%\todo{describe hyperbolic}

%Under this metric, entailment from $a$ to $b$ is defined as cosine similarity scaled by the difference in vector magnitudes:
%\begin{equation}
%p(b \mid a) = \frac{\mathbf{w}_a^{\top}\mathbf{w}_b}{\|\mathbf{w}_a\|_2 \, \|\mathbf{w}_b\|_2} \cdot \left( \|\mathbf{w}_a\|_1 - \|\mathbf{w}_b\|_1 \right),
%\end{equation}

%where $\mathbf{w}_a$ is the vector representation of $a$ and $\mathbf{w}_b$ is the representation of $b$. Semantic similarity is modeled using cosine similarity. The purpose of this metric is to see if vector norms can encode model entailment.

%All these are trained using the CachedMultipleNegativeLoss from Sentence transformers with a batch size of 2048 and min batch size of 8. We use the following tempoatuies. Intersection temperature = 0.001 and volume temperature = 1

%In addition to this baseline, we evaluate an existing entailment-aware embedding method, CS-Delta [cite], which explicitly models entailment using vector differences. This allows us to compare box embeddings not only against similarity-based representations, but also against prior approaches designed for entailment.

\subsection{Evaluation Metrics}
We evaluate the training process using two complementary metrics: a semantic similarity metric derived from STS-B~\citep{cer2017semeval} and an entailment metric computed on a held-out subset of the SURI dataset. Full results are reported in the appendix.

\heading{Retrieval with FollowBench:}
To evaluate a model’s ability to retrieve more specific yet relevant instructions, we leverage FollowBench~\citep{jiang-etal-2024-followbench}, which consists of instruction groups organized by increasing constraint levels. %Each instruction group contains variants of the same instruction with increasing constraint countsz.
Given a query at level $\mathcal{L}$, the model is tasked to retrieve another query from the same semantic group but $\mathcal{L' > L}$. Success requires the model to correctly prompts that are both semantically similar and also more specific. We evaluate on 688 queries. %comparing against two baselines: random corpus-wide retrieval and random sampling within the correct instruction group (achieving ~50\% accuracy by construction). 
%Table~\ref{tab:results} shows that vector embeddings achieve 64.0\% accuracy, modestly outperforming the in-group random baseline. Box embeddings substantially improve to 73.7\% (+9.7pp) when considering links and entailment data, and 77\% without links, demonstrating superior modeling of constraint hierarchies.

\heading{Score Prediction on UltraFeedback:}
A good embedding space should put the prompts that induce similar response scores close to each other, which allows us to run a kNN (k nearest neighbor) regressor on the embedding space to predict the response scores of unseen prompts. We compare the regressor performance using different embedding spaces on the UltraFeedback~\citep{cui2024ultrafeedbackboostinglanguagemodels} dataset, which contains instructions paired with responses from 17 LLMs and associated quality scores. We split each set of scores and responses into a 70/30 retrieval–test corpus split.

% Since each instruction is evaluated by only a subset of models, we construct 17 model-specific instruction sets. 
% We split each set is split 70/30 into a training/retrieval corpus and a test set. 

For each test instruction, we retrieve the top-5 corpus examples and predict the response score of the test prompt by averaging the scores from the training corpus, reporting root mean squared error (RMSE) against the gold score.

\subsection{Results}

\heading{Retrieval Followbench: }
% CSDelta while dominating all other models on the SURI benchmark but performs abysmally on FollowBench. We hypothesize that this discrepancy arises from the fundamental differences between the two evaluation setups. The held-out SURI evaluation set is based on triplets, which primarily test whether a model can make correct distinctions within the pairs in the triplet. In contrast, FollowBench involves a retrieval-style task and therefore depends more strongly on the global structure of the embedding space. In this scenario, CSDelta struggles to determine whether a high entailment score arises from semantic similarity or simply from a large difference in vector norms. As a result, the model lacks sufficient information to reliably distinguish truly relevant items from unrelated ones.
From Table~\ref{tab:combined_results} we see that in general, the box variants perform better at retrieving children prompts than all the other embedding models, which means box embeddings are more effective at modeling entailment. \textbf{Box} is better than \textbf{Box w/o synth}. Furthermore, \textbf{Box} wins over \textbf{Box w/o entail} and \textbf{Box w/o entail} is better than \textbf{Vector w/o entail}. This suggests that the gains in entailment performance stem from two complementary factors: (1) the synthesized entailment training data, and (2) the representational capacity and inductive bias of box embeddings. 

\heading{Score Prediction Metrics:}
In~\Cref{tab:combined_results} the lowest RMSE comes from the box based models, which suggests the specificity encoded by box volume correlates with the difficulty of the prompts. \textbf{Hyperbolic} and \textbf{CSDelta} achieve similar results when compared with the \textbf{Vector} baseline. 
%Among boxes, \textbf{box w/o links} has the lowest RMSE, which also performs best in FollowBench. \todo{This paragraph should be merged with the last paragraph}

%From the evaluation metricsAs we will see in the test on the downstream tasks later on, we need a combination of both semantic relevance and also hierarchical stuff. Thus we consider the model trained with links as the main model, but we provide ... in the later sections to (satisfy word)/strengthen this claim. Thus, we use the model trained with links dataset in the rest of the experiments.

%From these results, we draw two main conclusions. First,  Second, 

These gains are achieved with essentially the same setup as the baselines, aside from a minor $\sim$2\% increase in parameters, which is insufficient to explain the performance gap, suggesting that the improvements stem from the representational advantages of box embeddings. %The slight improvements over \textbf{Box w/o synth} show the synthesized datasets provide a small but consistent improvement for box models.

Although the \textbf{Box w/o links} achieves the strongest performances in \Cref{tab:combined_results} among the box variants, \Cref{tab:sts_suri_results} in Appendix~\ref{sec:intrinsic_metrics} shows that this comes at the cost of reduced semantic similarity performance. By incorporating link data, \textbf{Box} achieves a better balance between entailment modeling and semantic similarity, which not only is important for visualization in \Cref{sec:visualization} but also improves downstream applications such as hierarchical clustering in \Cref{sec:hier_clustering} and specificity estimation in \Cref{sec:spec_est}.

\begin{table}[t]
\centering
\small
\setlength{\tabcolsep}{5pt}
\resizebox{\linewidth}{!}{%
\begin{tabular}{lc cc}
\toprule
\textbf{Model} & \textbf{FollowBench} & \multicolumn{2}{c}{\textbf{UltraFeedback}} \\
\cmidrule(lr){3-4}
              & \textbf{Accuracy↑}  & \textbf{Avg.\ RMSE↓} & \textbf{Improv.\ (\%)↑} \\
\midrule
Random            & ---                & 1.8293              & 0.0 \\
Hyperbolic        & 0.659              & 1.6260                 & 11.11 \\
CSDelta           & 0.012              & 1.5955              & 12.80 \\
Vector            & 0.640              & 1.6059              & 12.21 \\
Vector w/o entail & 0.627              & 1.5605              & 14.71 \\
Box w/o entail    & 0.687              & 1.5610              & 14.67 \\
Box w/o links     & \textbf{0.775}     & \textbf{1.4777}     & \textbf{19.22} \\
Box w/o synth     & 0.716              & \secondbest{1.5280}                 & \secondbest{16.47} \\
Box               & \secondbest{0.738} & \secondbest{1.5280}              & \secondbest{16.47} \\
\bottomrule
\end{tabular}%
}
\caption{FollowBench accuracy and UltraFeedback score prediction performance (Avg.\ RMSE and its improvement over Random across 17 LLMs). Best results are in bold; second-best in blue.}
\label{tab:combined_results}
\end{table}

%4.5 In-Context Learning (ICL) Analysis
%
%We further evaluate representations in an in-context learning (ICL) setting. Our hypothesis is that entailment-aware retrieval selects examples that are both semantically similar and more specific, something that is more important than. We retrieve 5 best samples and then rerank them based on the score difference from ...
%
%We compare:
%
%ICL using examples retrieved via box-based entailment-aware similarity, and
%ICL using examples retrieved via vector-based similarity.

%\input{contents/evaluation}
\section{Box Visualization and Analysis}
\label{sec:visualization}
We train high-dimensional box embeddings to capture complex entailment structure among prompts. However, to analyze LLM weaknesses visually, like in \Cref{First_fig}, we need to reduce the embeddings to low-dimensional embedding space. Since no existing dimensionality reduction method is designed for box embeddings, we propose \VisMethod (see Appendix~\ref{sec:dim_reduction} for details). In this section, we present two applications demonstrating how our box embeddings can be used to analyze prompts and find LLM weaknesses. The interactive visualization can be viewed at \url{https://zawedcvg.github.io/P2B/visualisation.html}.

%\hs{transition. Box should be used to find weakness. explain why we need dimension reduction. high dimensional box could model the complex entailment structure better. Refer to appendix. Emphasizing that we propose this. How we evaluate the visualization projection }

%In the previous section, we demonstrated the advantages of box embeddings over vector embeddings in modeling constraint hierarchies and retrieving more relevant instructions. We now show how these properties translate into practical downstream applications. First we create interpretable visualisations of box embeddings. 
%We introduce a novel method for visualizing n-dimensional box embeddings in a two-dimensional space. 

%For dimensionality reduction, 

%\hs{How we evaluate the visualization projection }
%\hs{show two example usages of our box visualization}

% Volume Correlations:  
%   Spearman: 0.83
% Similarity Matrix Correlations:
%   Spearman: 0.87
% Entailment Matrix Correlations:
%   Spearman: 0.84

%\hs{The goal is to map each prompt $a_i$, represented by a box $x_i$ in a high-dimensional box space, to a box $y_i$ in a lower-dimensional space, such that its relationships with all other prompts are similar in both the high and low dimensional space.}

\begin{figure}[h]
     \centering
     \includegraphics[width=1\linewidth]{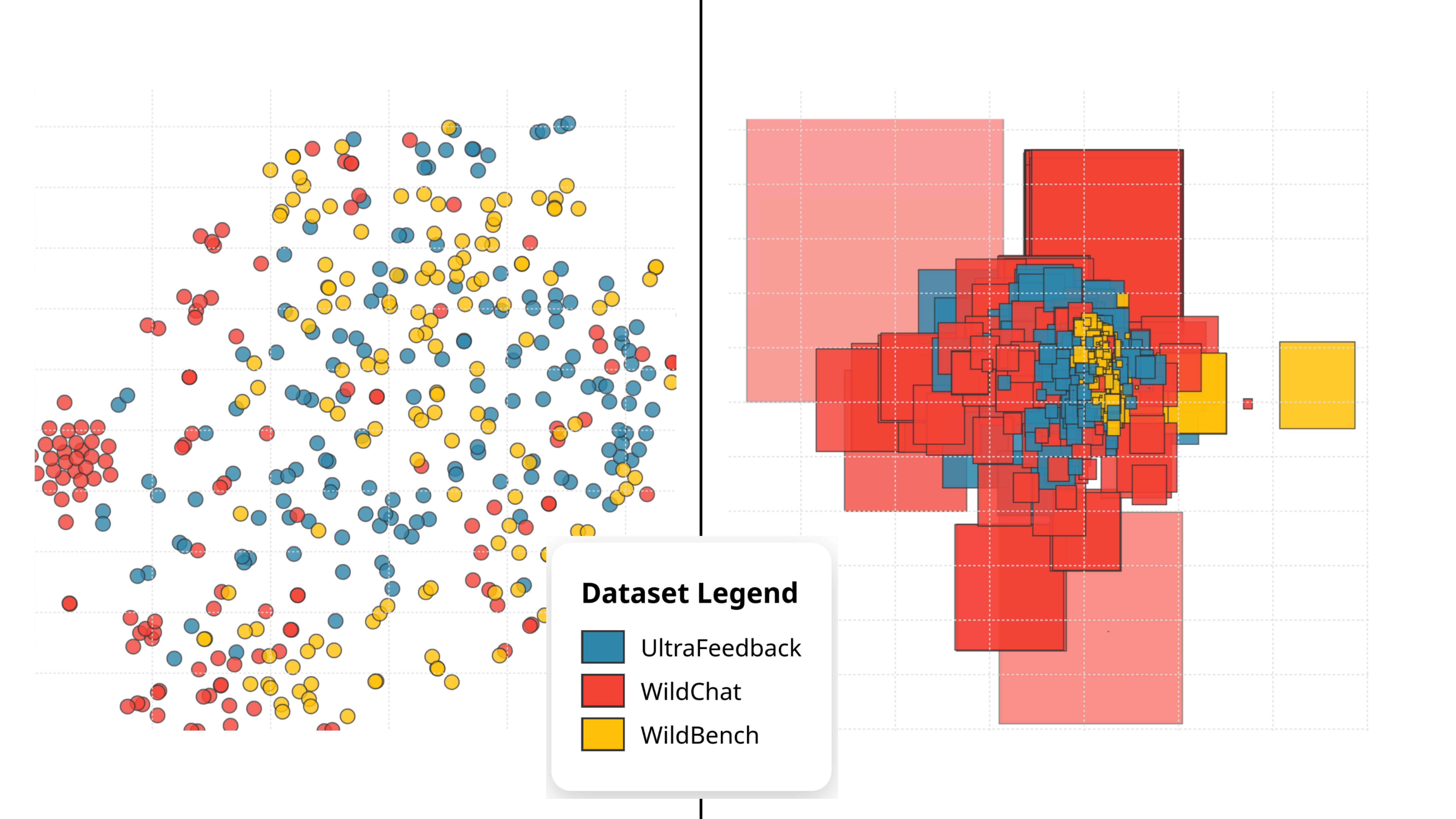}
     \caption{Comparison between our box-based visualization (right) and a t-SNE visualization of the vector baseline (left).}
     \label{fig:dataset_comparison}
 \end{figure}

\subsection{Comparing Different Datasets}

%In the right side of \Cref{fig:placeholder}, the Spearman correlations of the volumes, intersections, and entailments before and after our dimension reduction are 0.83, 0.87, and 0.84, respectively. This means that the orders of the volumes, intersections, and entailments in the high dimension are mostly preserved in the low dimension.
%our dimension reduction method achieves
% Data mix
% Volume Correlations:  
%   Spearman: 0.83
% Similarity Matrix Correlations:
%   Spearman: 0.87
% Entailment Matrix Correlations:
%   Spearman: 0.84

%WildBench is a subset of WildChat, so intuitively, they should have very similar distributions. However, the box-based visualization clearly demonstrates that WildBench examples are consistently more specific (i.e., represented by smaller boxes) than those from WildChat and UltraFeedback. The visualization reminds us that users often ask pretty general questions in WildChat, while WildBench is designed to include more challenging prompts, which is a detail that could be easily neglected from LLM developers. From the vector-based visualization baseline, it is difficult to learn the insight or distinguish the datasets from one another.
We first visualize 150 random examples from each of three datasets: WildChat, UltraFeedback, and WildBench~\citep{linwildbench}.
WildBench is a curated subset of WildChat, constructed to contain more challenging and more specific prompts. From Figure~\ref{fig:dataset_comparison}, the box-based visualization on the right makes this distinction explicit; WildBench examples are consistently represented by smaller boxes, indicating higher specificity, compared to those from WildChat and UltraFeedback. %\ssd{pointer to specific picture / figure} %This visualization highlights that real users often ask broad, general questions, while WildBench emphasizes more specific prompts, a detail that can be easily overlooked by LLM developers. 
In contrast, the vector-based visualization baseline on the left fails to clearly distinguish the datasets, making such insights difficult to discern.

%easily gleaned from the box visualization but not the vector baseline.
%clearly demonstrates that 

%This suggests that WildBench contains harder/more constrained instructions. 
%If we look at the construction of WildBench, we see that it is a subset of WildChat, designed to include more challenging prompts, a relationship that is easily gleaned from the box visualization but not the vector baseline.

 %In contrast, 

\subsection{Comparing Model Performance}

% model comparison
% Volume Correlations:
%   Spearman: 0.73
% Similarity Matrix Correlations:
%   Spearman: 0.88
% Entailment Matrix Correlations:
%   Spearman: 0.85

\begin{figure}[t!]
    \centering
    \includegraphics[width=1\linewidth]{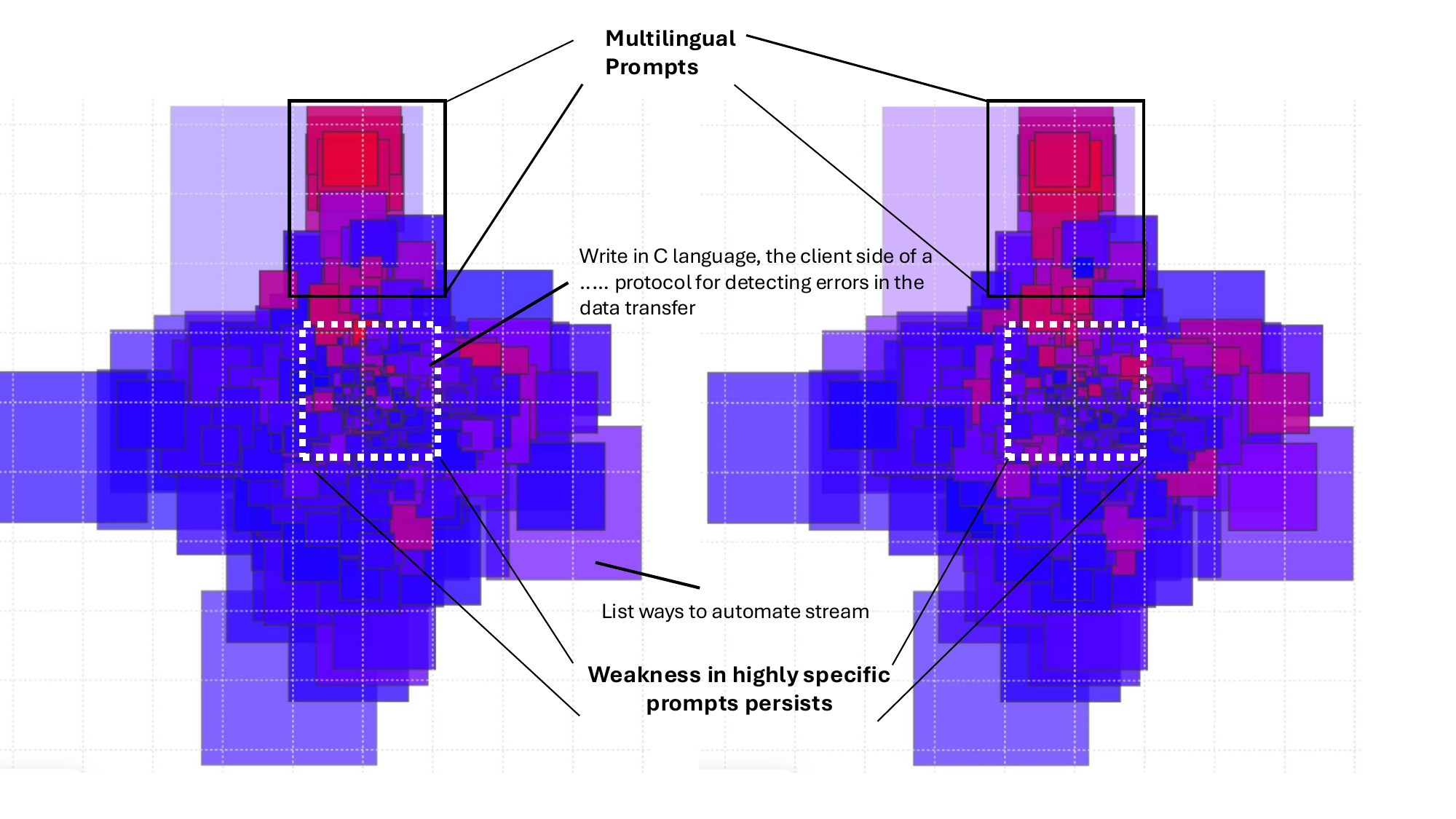}
    \caption{Comparison of LLMs' performance in 2D box embedding space. LLM performs better in the blue regions than in the red regions.}
    \label{comparison_small_large}
\end{figure}

Next, we compare visualizations for LLaMA-2-7B and LLaMA-2-70B to examine how box-based representations reveal both the benefits and limits of scaling. 
%In \Cref{comparison_small_large}, the Spearman correlations of the volumes, intersections, and entailments before and after our dimension reduction are 0.73, 0.88, and 0.85 , respectively. We show two example prompts using orange texts, which suggests that our box sizes correlate well with prompt specificities.
As shown in \Cref{comparison_small_large}, increasing model size reduces low-scoring (red) regions and improves performance across much of the space, as scaling laws~\citep{kaplan2020scaling} suggests. However, the upper region, corresponding mainly to multilingual prompts (black box), remains dominated by low scores. Interestingly, within the largely high-performing (blue) regions of the larger model, we observe small, dispersed red boxes (green outline) indicating failures on highly specific prompts. Some of these failure regions exist in the smaller model. This indicates that scaling does not uniformly eliminate fine-grained weaknesses: certain specific prompts remain challenging. %The \VisMethod visualisations can be seen in the project   
%\href{https://zawedcvg.github.io/P2B/visualisation.html}{website}.
 %\ssd{remove identity information like this one.}

\section{Hierarchical Clustering}
\label{sec:hier_clustering}
Hierarchical clustering of prompts enables evaluators to identify LLM weaknesses at multiple levels of abstraction, from broad skill categories to specific sub-skills.
%\hs{However, existing frameworks like SkillVerse~\citep{tian2025skillverse} and EvalTree~\citep{zeng2025evaltree} rely on vector-based clustering (e.g., Euclidean distance, recursive $k$-means), which cannot capture the inherently directional and asymmetric nature of instruction specificity.}
%\hs{Box embeddings address this limitation.} 
\Cref{sec:exp_emb} shows that box embedding is better at putting the prompts with entailment relationship or with similar score close to each other.
%boxes model specificity much better than their vector counterpart.\todo{We haven't shown that up to this point, right} 
Next, we investigate if this ability translates to creating better clusters for discovering LLMs' weaknesses. Directly evaluating such clusters with human experts is difficult due to the cost, required expertise, and the need to maintain a consistent evaluation setup across many models and clusters. A comprehensive human evaluation of this kind is therefore beyond the scope of this paper. We instead use a set of well-defined proxy metrics to assess how effectively different clustering methods isolate regions of model weakness. 

%which in turn will improve .... 
To group the smaller boxes that have an entailment relation at the lower levels, we propose a new hierarchical clustering method that always merges the two boxes with the smallest resulting box first (see Appendix~\ref{app:hierarchical} for details) and compare it with the classic Ward linkage~\citep{ward1963hierarchical} on the vector space.
%For creating the clusters, we use a volume based distance for boxes and exisiting ... for vectors 

Since there is no existing ground-truth prompt hierarchical structure, we compare box-based and vector hierarchical clusters using the following three metrics: (i)~consistency of model scores within local neighborhoods, (ii)~correlation between depth and instruction specificity, and (iii)~ability to discover model weakness clusters.

\subsection{Local Score Consistency}
%We begin with the assumption that an LLM should exhibit similar performance on semantically related prompts. Under this assumption, 
A well-formed hierarchical clustering should place semantically similar prompts under the same parent node. This is desirable because similar prompts tend to elicit similar LLM responses, leading to similar scores. Thus, if semantically dissimilar prompts are grouped together, their scores may vary widely, rendering the cluster average unrepresentative and obscuring true weaknesses (e.g., the \textit{A}--\textit{D} cluster in \Cref{First_fig}).

We measure this property by computing the average absolute score difference between neighboring leaf nodes, where neighbors are defined as prompts sharing the same immediate parent in the hierarchy.

%We only use leaf nodes that have one neighbor in both the box-based and vector-based hierarchies.
As a random baseline, we assign each prompt a randomly sampled neighbor from the set of 500 prompts and compute the score differences. The results in \Cref{tab:weakness_comparison} show that box achieves a 38\% relative improvement compared with the vector baseline (i.e., $\frac{12.88\%-9.32\%}{9.32\%}$ ).

%\hs{correspond skillverse}ng .

% \begin{table}[t]
% \centering
% \footnotesize
% \setlength{\tabcolsep}{4pt}
% \begin{tabular}{lcccc cc}
% \toprule
% \textbf{Metric} & \textbf{Rand.} & \textbf{Vec.} & \textbf{Box} & \textbf{Box w/o links} & \textbf{Box w/o synth} \\
% \midrule
% Local Score Consistency   & 0.0\%  & 9.32\%  & \textbf{12.32\%} & 10.14\% & 11.08\% \\
% Spec.--Depth Agreement & 50.00\% & 52.71\% & \textbf{68.34\%} & 61.22\% & 64.87\% \\
% All Size Weakness     & --     & 0.0\%   & \textbf{7.70\%}  & 5.33\%  & 6.91\%  \\
% AUC & --     & 0.0\%   & \textbf{13.47\%} & 9.85\%  & 11.62\% \\
% \bottomrule
% \end{tabular}
% \caption{Comparison between random, vector, and box embeddings across weakness discovery, LLM agreement, and score improvement metrics. 
% %The improvement ratios are averaged across 17 LLMs evaluated in UltraFeedback.
% }
% \label{tab:weakness_comparison}
% \end{table}

\begin{table}[t]
\centering
\footnotesize
\setlength{\tabcolsep}{3pt}
\resizebox{\columnwidth}{!}{%
\begin{tabular}{lccccc}
\toprule
% \textbf{Metric} & \textbf{Rand.} & \textbf{Vec.} & \textbf{Box} & \textbf{Box\textsubscript{$-$lnk}} & \textbf{Box\textsubscript{$-$syn}} \\
\textbf{Metric} & \textbf{Rand.} & \textbf{Vec.} & \textbf{Box w/o links} & \textbf{Box w/o synth}  & \textbf{Box} \\
\midrule
Loc.\ Score Cons.        & 0.0\%  & 9.32\%  & \textbf{13.24\%} & 10.34\%  & 12.88\%\\
Spec.--Depth Agr.        & 50.00\% & 52.71\% & 67.07\% & 66.03\%  & \textbf{68.34\%} \\
All Size Weakness        & --     & 0.0\%    & 2.85\%  & 5.94\% & \textbf{7.70\%}  \\
AUC                      & --     & 0.0\%    & 7.81\%  & 10.95\% & \textbf{13.47\%} \\
\bottomrule
\end{tabular}}
\caption{Comparison across weakness discovery, LLM agreement, and score improvement metrics.
}
\label{tab:weakness_comparison}
\end{table}

\subsection{Specificity Ordering Accuracy}

Next, we evaluate how well hierarchical depth aligns with instruction specificity using LLM-as-a-judge. We select 500 instructions with available LLaMA-2-13B-Chat responses to limit LLM inference cost. Because direct comparisons between arbitrary instructions are ambiguous, we only select pairs of similar prompts. Exact evaluation details are provided in Appendix~\ref{sec:spec_details}.
%\hs{We present the exact methodology of getting pairs in the appendix}.

Table~\ref{tab:weakness_comparison} shows that vector-based hierarchies perform almost the same as the random baseline, while box-based hierarchies achieve over 68\% specificity accuracy, a 33\% relative improvement over both baselines, demonstrating that box-induced hierarchies effectively capture instruction specificity. We also include a human study in the Appendix~\ref{sec:human_study} to validate the LLM-as-a-judge metric and confirm our conclusion.

% of having everything on the same level

\subsection{Cluster Weakness Containment}
\label{sec:weakness_analysis}
Finally, we evaluate how well the clustering algorithm can identify and isolate "model weaknesses". We define a weakness as an instruction cluster for which the model's average score lies at or below the 25th percentile. The underlying assumption is that 
%LLM weaknesses show in specific types of prompts and 
a good clustering algorithm will be able to accurately group the low-score prompts into a large cluster. Mixing unrelated instructions will "dilute" the weaknesses and thus lead to an inflation of average scores.

%first set a cluster size threshold $t_s$ and count the number of weakness clusters whose size 
%analyze clusters whose average scores fall in the bottom 25\% of the model's scores and analyze the relationship between cluster size and also the number of these low scoring clusters. 

%On some models we notice a bigger improvement, like 39.3\% more for gpt-3.5-turbo and  . Further, we see that the number of clusters also decay slower with cluster size, indicating their ability to retain clusters with weaknesses even when the cluster size is quite high.
%\hs{explaining AUC results}
To detect this effect, we compute the number of weakness clusters as
\[
\#W_{t_s}^{R_{25\%}} = \left|\left\{ w \;\middle|\; S(w) \leq R_{25\%} \;\text{and}\; |w| \geq t_s \right\}\right|,
\]
where $|w|$ denotes the size of cluster $w$, $t_s$ is the minimum cluster size threshold, $S(w)$ is the average score of cluster $w$, and $R_{25\%}$ is the closest integer score below the 25th percentile.

We report $\#W_{2}^{R_{25\%}}$ (i.e., All Size Weakness) in \Cref{tab:weakness_comparison}, and the aggregate measure $\sum_{t_s>1} \#W_{t_s}^{R_{25\%}}$, which corresponds to the area under the curve (AUC) of weakness cluster counts across different size thresholds, in \Cref{tab:auc-improvement} and \Cref{fig:appendix_cluster_scores_group1,fig:appendix_cluster_scores_group2,fig:appendix_cluster_scores_group3}. The overall percentage improvements are summarized in \Cref{tab:weakness_comparison}.

\Cref{tab:weakness_comparison} shows that hierarchical clustering based on box embeddings identifies a greater number of weakness clusters. When considering the AUC across all cluster sizes, the improvement increases to 13.47\%, indicating that box embeddings can capture more and larger weakness clusters.

\section{Specificity Estimation}
\label{sec:spec_est}

As the prompt becomes more specific, the average score of LLMs' responses should decrease. We measure how well the learned box volume models specificity. We use prompt length as a baseline as it has been shown as a strong baseline in modeling specificity~\citep{li2015fast,gao2019predicting,kapur2026more}. We thus plot the scores in UltraFeedback and LMSYS-Chat-1M~(See Appendix~\ref{app:line_coverage} for more details) against $log(\text{length})$ in \Cref{fig:bestfit_length} and against $log(\text{box volume})$ in \Cref{fig:bestfit_volume}. To estimate the average score in the curve, we use kernel density estimation (KDE), whose bandwidth is decided by using Silverman's rule-of-thumb~\citep{silverman2018density}. In these figures, the LLMs' scores often saturate for very specific or very general prompts, so we use the sigmoid function to fit the curves. \Cref{fig:bestfit_volume} shows various saturation patterns from different LLMs. 

A better specificity estimation should have a smaller RMSE to the fitted sigmoid curve and a longer linear region (i.e., line coverage) of the sigmoid function, which demonstrate its ability of modeling the subtle difficulty differences. Appendix~\ref{app:line_coverage} describes how the line coverage is measured. \Cref{tab:volume_vs_length} shows that \textbf{Box} achieves the highest improvement over the strong length baseline in UltraFeedback and LMSYS-Chat-1M~\cite{zheng2023lmsyschat1m}.

%In the sigmoid function, the longer the middle linear region shows the specificity estimation at x-axis is better at distinguishing the subtle difficulty differences. 

%In \Cref{tab:volume_vs_length}, we report the relative improvement of RMSE sigmoid
%should be as large as possible 

%From these figures, we can see sigmoid
%line coverage

%length is a strong baseline for specificity

%Volume-score relationships show substantially clearer trends than length-score relationships. Fitting a sigmoid curve to each, volume-based fits achieve ~45\% lower RMSE on average (Table~\ref{tab:volume_vs_length}). Volume also exhibits ~111\% greater linear coverage(see appendix ...) in the sigmoid's central region — where specificity and score are most tightly coupled — suggesting it captures a broader, more continuously informative relationship than length alone.
%We validate on LMSYS-Chat-1M~\cite{zheng2023lmsyschat1m} prompts scored with the Skywork v2~\cite{liu2025skywork} reward model. Though noisier, volume still achieves 18\% lower RMSE and 65.6\% greater linear coverage, consistent with the above findings.

%KDE smoothing
%sigmoid function

%We estimate prompt specificity using UltraFeedback response scores, treating score as a proxy for difficulty.
%, consistent with prior work showing that more specific prompts are more challenging for LLMs (add citation). 
%We compare how well log(length) and log(box volume) predict response scores using 
%kernel density estimation (KDE). The 

\begin{table}[t]
\centering
\resizebox{\columnwidth}{!}{%
\begin{tabular}{lcccc}
\toprule
 & \multicolumn{2}{c}{\textbf{UltraFeedback}} & \multicolumn{2}{c}{\textbf{LMSYS}} \\
\cmidrule(lr){2-3} \cmidrule(lr){4-5}
\textbf{System} & RMSE Impr. & Line Cov. Impr. & RMSE Impr. & Line Cov. Impr.\\
\midrule
Length         & 0.0\%             & 0\%   & 0.0\% & 0\% \\
Box w/o links  & 44.3\%          & 83\%  & 14.8\%          & 10\%   \\
Box w/o synth  & 39.8\%          & 113\% & 9.4\%           & \textbf{72\%}   \\
Box            & \textbf{45.3\%} & \textbf{113.5}\% & \textbf{18.3\%} & 71\% \\
\bottomrule
\end{tabular}}
\caption{Ablation results across 16 LLMs on UltraFeedback and 16 LLMs on LMSYS. The numbers of boxes are relative improvement to the length baseline.}
\label{tab:volume_vs_length}
\end{table}

% \begin{table}[t]
% \centering
% \resizebox{\columnwidth}{!}{%
% \begin{tabular}{lcccc}
% \toprule
%  & \multicolumn{2}{c}{\textbf{UltraFeedback}} & \multicolumn{2}{c}{\textbf{LMSYS}} \\
% \cmidrule(lr){2-3} \cmidrule(lr){4-5}
% \textbf{System} & RMSE Impr. & Line Cov. & RMSE Impr. & Line Cov. \\
% \midrule
% Box w/o links  & 44.3\%          & 83\%  & 14.8\%          & 10\%   \\
% Box w/o synth  & 39.8\%          & 113\% & 9.4\%           & 72\%   \\
% Box            & \textbf{45.0\%} & 111\% & \textbf{18.3\%} & 70.8\% \\
% \bottomrule
% \end{tabular}}
% \caption{Ablation results across 17 LLMs on UltraFeedback and 28 LLMs on LMSYS. The numbers are the }
% \label{tab:volume_vs_length}
% \end{table}

% \todo fix this
% \begin{table}[t]
% \centering
% \small
% \begin{tabular}{llc}
% \toprule
% \textbf{Dataset} & \textbf{Metric} & \textbf{Improvement} \\
% \midrule
% \multirow{2}{*}{UltraFeedback} 
%   & RMSE            & $\downarrow$ 45.0\% \\
%   & Lin.\ Coverage  & $\uparrow$ 111.0\% \\
% \midrule
% \multirow{2}{*}{LMSYS}         
%   & RMSE            & $\downarrow$ 18.3\% \\
%   & Lin.\ Coverage  & $\uparrow$ 65.6\% \\
% \bottomrule
% \end{tabular}
% \caption{Volume vs.\ length specificity estimation}
% \label{tab:volume_vs_length}
% \end{table}
%\vspace{-1.75em}
\section{Conclusion}
We introduce \method, a method for embedding prompts into a box embedding space that jointly captures both semantic similarity and specificity. Through extensive experiments, we demonstrate that \method consistently outperforms vector-based baselines across multiple metrics for modeling entailment and predicting the scores of neighboring prompts.% while maintaining competitive similarity performance. 
We provide intuitive 2D visualizations that illustrate how box embeddings naturally encode the hierarchical structure of prompt specificity through geometric containment, offering insights that vector-based visualizations cannot capture. Furthermore, we validate the practical utility of our approach on the hierarchical clustering, where \method achieves superior performance across four distinct evaluation metrics. Finally, measuring prompt specificity using box volume creates a new analysis dimension for LLMs' behavior (e.g., we discover that the scores of 16 different LLMs versus specificity forms different sigmoid functions in UltraFeedback). 

%line coverage
%length baseline
%\todo{add specificity results}
%recommend widely use
%Our study introduces another dimension 

\section{Limitations}
%\todo{future work -> limitation}
Due to the limited space, we only test $3$ applications of box embedding. Understanding LLMs' ability and modeling specificity have many other potential applications. For example, besides comparing datasets or LLMs with different sizes, we can also compare LLM judges, or LLMs with different training stages or hyperparameters. Furthermore, our prompt specificity/difficulty estimation might help LLM routing~\citep{guha2024smoothie,kashani2025representing}, evaluation data selection~\citep{zouhar2025select} and creativity evaluation~\citep{atmakuru2024cs4,lu2025benchmarking}, prompt safety analysis~\citep{ayub2024embedding}, response specificity estimation~\citep{jiang2025conformal}, LLM interpretability~\citep{shani2025tokens}, and knowledge editing~\citep{ge2024well}. We leave them as our future work. 

Many of our applications experiments rely on UltraFeedback, which is a very effective dataset for LLM aligment~\citep{ivison2024unpacking} because many modern benchmarks only focus on difficult prompts and it is hard to find the benchmarks containing prompts with various specificity and lots of LLM responses and scores.

To ensure our methods work well in general domains, we do not train our methods using UltraFeedback. 
Nevertheless, the effectiveness of our approach still depends on the availability of existing entailment datasets and LLMs' ability to synthesize the training data, so \method might not be as effective in some special domains or languages with fewer resources. 

We observe that the 2D boxes created using \VisMethod tend to suffer from overcrowding issues analogous to those encountered in early vector visualization methods. Developing normalization or regularization techniques to encourage greater spatial dispersion remains an open challenge; notably, standard approaches like t-SNE do not directly apply here, as box volumes are represented in log-space. Addressing this limitation could further improve interpretability.

\section*{Acknowledgement}
This work was supported in part by the Center for Intelligent Information Retrieval and in part by the National Science Foundation (NSF) grant numbers IIS-2106391. Any opinions, findings and conclusions or recommendations expressed in this material are those of the authors and do not necessarily reflect those of the sponsor. The computational resources for this work are provided by the Unity Research Computing Platform, a multi-institutional cluster lead by University of Massachusetts Amherst, the University of Rhode Island, and University of Massachusetts Dartmouth.

%(if the space permits, move to future work or related work section section) 


\bibliography{example_paper, box}
%\bibliographystyle{acl_natbib}

%%%%%%%%%%%%%%%%%%%%%%%%%%%%%%%%%%%%%%%%%%%%%%%%%%%%%%%%%%%%%%%%%%%%%%%%%%%%%%%
%%%%%%%%%%%%%%%%%%%%%%%%%%%%%%%%%%%%%%%%%%%%%%%%%%%%%%%%%%%%%%%%%%%%%%%%%%%%%%%
% APPENDIX
%%%%%%%%%%%%%%%%%%%%%%%%%%%%%%%%%%%%%%%%%%%%%%%%%%%%%%%%%%%%%%%%%%%%%%%%%%%%%%%
\newpage
\appendix
\onecolumn

% You can have as much text here as you want. The main body must be at most $8$
% pages long. For the final version, one more page can be added. If you want, you
% can use an appendix like this one.

% The $\mathtt{\backslash onecolumn}$ command above can be kept in place if you
% prefer a one-column appendix, or can be removed if you prefer a two-column
% appendix.  Apart from this possible change, the style (font size, spacing,
% margins, page numbering, etc.) should be kept the same as the main body.

\section{Expansion of Definition of Terms}
\label{app:expansion_of_def}
Let $\mathcal X$ denote the space of instructions. Let $\mathcal U$ denote the universe of all possible constraints and we assume the number of possible constraints $|\mathcal U|$ is finite. For any instruction $x \in \mathcal X$, let $\mathcal C : \mathcal X \rightarrow 2^{\mathcal U}$
be a mapping from an instruction to the set of constraints it induces, where
\begin{equation}
\mathcal{C}(x) = \{\, c \in \mathcal{U} \mid c \text{ holds for all valid responses to } x \,\}.
\end{equation}
Recall that for any two statements $g$ and $h$, if $g$ entails $h$ (written $g \models h$), then whenever $g$ is true, $h$ must also be true. In other words, $g$ imposes a stronger condition than $h$. When $g$ and $h$ are both prompts, $g \models h$ means that any solution satisfying $g$ also satisfies $h$. Formally, constraint inclusion is the same as entailment between the two prompts:
\begin{equation}
\forall\, a,b \in \mathcal X,\qquad
\mathcal C(a)\supseteq\mathcal C(b)
\;\Longleftrightarrow\;
a \models b .
\end{equation}
Furthermore, we say the prompt $a$ is more specific than prompt $b$ if the prompt $a$ has more constraints (i.e., $C(a)\supset\mathcal C(b)$).
Taking the example present in Figure~\ref{First_fig}, we have two prompts: $g =$ ``\textit{Please write a short story about an adventure of a robot}'' and $h =$ ``\textit{Please write a short adventure story.}''  Since it is not possible to exhaustively list out the set of all the possible constraints, one can intuitively say that $\mathcal C(h) \subset \mathcal C(g)$ if $g$ can be written as $h$ with additional constraint(s). We can thus write $g$ as ``\textit{Please write a short adventure story; Make the story about a robot.}'' We see that $\mathcal{C}(g) \supset \mathcal{C}(h)$ , thus based on the previous definition: $g \models h$.

Let $\mathcal{Y}$ denote the universe of all possible responses. For any instruction $x \in \mathcal{X}$, let $\mathcal{S} : \mathcal{X} \rightarrow 2^{\mathcal{Y}}$
be a mapping from a prompt to the set of valid responses that satisfy the constraints imposed by $x$, where $\mathcal{S}(x) \subseteq \mathcal{Y}$. 

Let's assume we have a pair of prompts $a, b \in \mathcal{X}$ and $a$ contains more constraints than $b$. Since any valid solution to $a$ must satisfy the stricter set of constraints in $a$, they must also satisfy all the constraints for $b$. Further, the set of valid solutions for $a$ is smaller than that for $b$. Consequently, inclusion in the constraint space induces reverse inclusion in the solution space:
\begin{equation}
\mathcal C(a)\supseteq\mathcal C(b)\quad\Longrightarrow\quad \mathcal S(a)\subseteq\mathcal S(b).
\end{equation}

We can prove the other direction of this by:

%$(\Rightarrow)$: 
Assume $S(a) \subseteq S(b)$. Let $c \in C(b)$, so $c$ is 
satisfied by all elements of $S(b)$. Since every element of $S(a)$ also belongs to $S(b)$, $c$ is 
satisfied by all elements of $S(a)$ as well, hence $c \in C(a)$. Thus $C(b) \subseteq C(a)$.

So we get 
\begin{equation}
\mathcal C(a)\supseteq\mathcal C(b)\quad\iff\quad \mathcal S(a)\subseteq\mathcal S(b).
\label{eq:CeqS}
\end{equation}

As constraints accumulate, the valid solution space contracts, increasing task difficulty by requiring the model to generate responses from a progressively smaller region of admissible outputs. This perspective offers an explanation for the empirically observed LLM performance degradation as the number of constraints increases \citep{jiang-etal-2024-followbench,tamkin2024clio,zeng2025evaltree,tian2025skillverse}. 

\subsection{Proof that $g \models h \quad\Longrightarrow\quad C(h) \subseteq C(g)$:}

If $g$ and $h$ are a paraphrase, $C(h) = C(g)$. Otherwise, we can prove it by viewing it from the solution space.

Since $g$ can be decomposed as $g \equiv h \cup k$ (i.e., $g$ is exactly $h$ with the additional 
constraint $k$ = ``Make the story about a robot''), any response satisfying $g$ must simultaneously 
satisfy both $h$ and $k$. Therefore:
\begin{equation}
    S(g) = \{ s \mid s \in Y \text{ satisfies } h \text{ and } s \in Y \text{ satisfies } k \} 
         = S(h) \cap S(k).
\end{equation}
Since $S(g) = S(h) \cap S(k) \subseteq S(h)$, we have:
\begin{equation}
    S(g) \subseteq S(h).
\end{equation}
By \Cref{eq:CeqS}, a smaller solution space corresponds to a larger constraint set --- formally, 
for any $a, b \in \mathcal{X}$:
\begin{equation}
    S(a) \subseteq S(b) \implies C(b) \subseteq C(a).
\end{equation}
Applying this with $a = g$ and $b = h$, it follows that:
\begin{equation}
    C(h) \subseteq C(g).
\end{equation}
%Since the decomposition introduces a strictly new constraint (robot), the inclusion is proper: $C(g) \supset C(h)$, which by Equation~(7) gives $g \models h$. \qed
%\todo{check}

\section{Prompt Representation}
\label{sec:box_rep}
Given the importance of specificity, an effective representation must capture both relevance and specificity between prompts. Traditional vector embeddings represent each prompt as a point in a metric space and model relationships solely through symmetric distance functions, making them ill-suited for expressing asymmetric relations such as specificity or constraint inclusion.

Formally, let $f : X \rightarrow \mathbb{R}^D$ denote a vector embedding function. Similarity between two prompts $a$ and $b$ is modeled using a distance function $d(f(a),f(b))$, which is invariant to direction and therefore cannot encode partial order relations of the form $\mathcal C(a)\supset\mathcal C(b)$.

% X \subset R^D, y \subset R^2.
% a_\delta \in R^2 --> R. For of reg that prevents from overfitting.

In contrast, box embeddings represent each prompt \(a \in X\) as an axis-aligned hyper-rectangle in \(\mathbb{R}^D\). 
Formally, for a prompt $a$, we parameterize its box embedding using a center vector
\( a_{\text{center}} \in \mathbb{R}^D \) and a width vector
\( a_{\delta} \in \mathbb{R}_+^D \). The lower and upper corners of the box are given by
\begin{equation}
a^\llcorner \coloneqq a_{\text{center}} - a_{\delta},
\qquad
a^\urcorner \coloneqq a_{\text{center}} + a_{\delta}.
\end{equation}
The box embedding for prompt $a$ is therefore defined as the cartesian product of each side of the rectangle:
\begin{equation}
\Box(a) \coloneqq \prod_{d=1}^D [a_d^\llcorner, a_d^\urcorner]
= [a_1^\llcorner, a_1^\urcorner] \times \ldots \times [a_D^\llcorner, a_D^\urcorner].
%\subseteq \mathbb{R}^D.
\end{equation}

Let us consider two prompts $a, b \in X$, with corresponding box representations $\mathrm{Box}(a)$ and $\mathrm{Box}(b)$. We define the volume of $\mathrm{Box}(a)$ as $\operatorname{Vol}(a)\coloneqq \prod_{d=1}^D (a_d^\urcorner - a_d^\llcorner)$. We first model \emph{prompt similarity} as the volume of the intersection between their boxes, i.e., $\operatorname{Vol}(\mathrm{Box}(a) \cap \mathrm{Box}(b))$. Since the intersection of two intervals is determined by the minimum of their upper bounds and the maximum of their lower bounds, the intersection volume is given by
\begin{equation}
\small
\operatorname{VolInt(a, b)}
\coloneqq
\prod_{d=1}^D
\max\left(
\min(a_d^\urcorner, b_d^\urcorner)
-
\max(a_d^\llcorner, b_d^\llcorner),
0
\right)
%\tag{i}
\label{eq:box_similarity}
\end{equation}

However, similarity alone is insufficient for our purposes. A key motivation for using box embeddings is their ability to model \emph{asymmetric interactions} between prompts. In particular, when prompt $a$ entails prompt $b$, we expect $\mathrm{Box}(b)$ to contain $\mathrm{Box}(a)$. In this case, the intersection volume equals the volume of $\mathrm{Box}(a)$, i.e., $\operatorname{VolInt}(a,b) =
\operatorname{Vol}(\mathrm{Box}(a))$.

We therefore define an \emph{entailment score} as the conditional probability
\begin{equation}
\label{eq:entailment}
p(b \mid a)
\coloneqq
\frac{
\operatorname{VolInt}(a, b)
}{
\operatorname{Vol}(\mathrm{Box}(a))
}
%\tag{ii}
\end{equation}

By construction, $p(b \mid a) = 1$ when $a$ fully entails $b$, and $p(b \mid a) < 1$ otherwise, providing a principled measure of asymmetric prompt entailment.

% \subsection{Optimizing using TSNE for centers}

% We also find out that using TSNE for the centers/the initial portion, helps out with the things, it leads to lower loss and .... We show examples below of with or without tSNE.

\subsection{Gumbel Box Specifics}
\label{appendix:gumbel_specific}

Hard $\min$ and $\max$ operations are replaced with temperature-controlled log-sum-exp ($\LSE$) operators. For one-dimensional intervals, the expected intersection length is approximated as
$$
\LSE_\beta\big(
\LSE_{-\beta}(x^\urcorner_1,\ldots,x^\urcorner_N)
-
\LSE_{\beta}(x^\llcorner_1,\ldots,x^\llcorner_N),
0
\big),
$$
where $\LSE_\beta(\mathbf{x}) \coloneqq \beta \log \sum_i \exp(x_i/\beta)$. In higher dimensions, the expected intersection volume is computed as a product across dimensions. We use this smooth approximation to replace hard volume-based quantities in the containment and similarity scores. Following \cite{gumbel_box}, we use separate temperature parameters for volume and intersection computations. In all our experiments, we fix these temperatures to $\beta_{\text{vol}} = \langle1.0\rangle$ and $\beta_{\text{int}} = \langle0.001\rangle$.

\section{Training Details}
\label{sec:training_details}

We train all models using the Multiple Negatives Ranking Loss~\cite{henderson2017efficientnaturallanguageresponse}, combined with GradCache~\citep{gao-etal-2021-scaling} to enable substantially larger effective batch sizes during training. Each mini-batch contains samples from only a single dataset. Datasets are sampled proportionally to their relative sizes, while batches are processed using a round-robin scheduling strategy.

We use a learning rate of \(2 \times 10^{-5}\) with a warmup ratio of 0.1. Training is performed primarily on NVIDIA 2080 Ti GPUs. For box embeddings, we set the volume temperature to 1.0 and the intersection temperature to 0.001.

\subsection{CSDelta}

For the model trained using the CSDelta metric, entailment from $a$ to $b$ is defined as cosine similarity scaled by the difference in vector magnitudes:
\begin{equation}
p(b \mid a) = \frac{\mathbf{w}_a^{\top}\mathbf{w}_b}{\|\mathbf{w}_a\|_2 \, \|\mathbf{w}_b\|_2} \cdot \left( \|\mathbf{w}_a\|_1 - \|\mathbf{w}_b\|_1 \right),
\end{equation}

where $\mathbf{w}_a$ is the vector representation of $a$ and $\mathbf{w}_b$ is the representation of $b$. Semantic similarity is modeled using cosine similarity. The purpose of this metric is to see if vector norms can encode model entailment.

\subsection{Hyperbolic}
For the hyperbolic model, we employ two distinct training objectives depending
on the relation type. Let $\mathbf{u}, \mathbf{v} \in \mathbb{H}^d$ denote
the encodings of prompts $a$ and $b$ in hyperbolic space, and let
$\|\cdot\|_L$ denote the Lorentzian norm.

\paragraph{Semantic Similarity.}
For symmetric similarity relations, we adopt the squared Lorentzian distance
of \citet{pmlr-v97-law19a}:
\begin{equation}
    s_{\mathrm{sym}}(\mathbf{u}, \mathbf{v}) =
    -\|\mathbf{u} - \mathbf{v}\|_L^2.
    \label{eq:hyperbolic-sim}
\end{equation}

\paragraph{Asymmetric entailment.}
For directed entailment relations, we use a score function adapted from
Equation~(8) of \citet{Nickel2017PoincarEF}, augmented with a norm-asymmetry
penalty:
\begin{equation}
    s_{\mathrm{asym}}(\mathbf{u}, \mathbf{v}) =
    -\Bigl(1 + \alpha\,\bigl(\|\mathbf{u}\|^2 - \|\mathbf{v}\|^2\bigr)\Bigr)
    \|\mathbf{u} - \mathbf{v}\|_L^2,
    \label{eq:hyperbolic-asym}
\end{equation}
where $\alpha \geq 0$ is a penalty weight that controls sensitivity to norm
asymmetry (set to $\alpha = 5.0$ in our experiments). 
% For hyperbolic model, we use the (symmetric) squared lorentzian distance function from Law et al. 2019~\citep{pmlr-v97-law19a} for training similarity, and an adaptation of the score function from equation (8) from \citep{Nickel2017PoincarEF} for training assymetric relations. Namely, for assymetry we have function will be
%         \[-(1 + \alpha (||u||^2 - ||v||^2)) ||u - v||_L^2\]
%         and for similarity
%         \[-||u - v||_L^2\]

%     where\[||.||_L \text{ is the Lorentzian distance.}\]

%     u and v are the encoding of a and b in hyperbolic space
    
%     where alpha: penalty for distance. We use alpha as 5.0
%         for edge direction more
%     :where beta: -1/curvature. We use curvature as 1.0

\subsection{WildChat Preprocessing}
We first filter the data by retaining only single turn interactions written in English and only include The instructions to those containing between 8 and 150 words; this range is chosen empirically to exclude trivial prompts and overly verbose instructions.

To reduce redundancy, we compute sentence embeddings using all-mpnet-base-v2 and remove instructions with cosine similarity greater than 0.9, eliminating near-duplicate or semantically equivalent prompts. The remaining instructions are then passed to GPT-4.1 using an in-context learning prompt (shown in \Cref{sec:hier_prompt}), which generates multiple levels of general instructions for each prompt.

\section{Intrinsic Metrics}
\label{sec:intrinsic_metrics}
\subsection{Datasets}
\subsubsection{Semantic Similarity (STS-B)}
Semantic Textual Similarity Benchmark (STS-B), which provides pairs of sentences and a similarity score associated with each of them. We compute the Spearman correlation between the ground truth similarity and $\operatorname{VolInt}(a, b)$ in \eqref{eq:box_similarity} for box. For vector baselines, we use cosine similarity.

\subsubsection{Held-out SURI Entailment Triplets}
We additionally compare models using the validation set of SURI.
This evaluates whether representations correctly rank more specific instructions to being entailed by their general counterparts than unrelated instructions or more general instructions.

\subsection{Results}
In~\Cref{tab:sts_suri_results}, 
%shows results of STS-B and SURI validation metrics. The 
the \textbf{vector} baseline, which focuses on modeling prompt similarity, unsurprisingly achieves the strongest performance on STS-B. However, the \textbf{box} embedding model trained all the datasets performs competitively on STS-B while being much better on SURI. \textbf{Hyperbolic} is better than box in the STSB baseline.

In SURI, \textbf{box} wins over \textbf{box w/o entail} and \textbf{box w/o entail} is better than \textbf{vector w/o entail}. This suggests that the gains in entailment performance stem from two complementary factors: (1) the synthesized entailment training data, and (2) the representational capacity of box embeddings. Interestingly, we also see a tradeoff between modeling entailment and similarity for the box models trained with and without links. When link data is removed during training, STS-B performance drops by an additional around 10 absolute points, despite link examples constituting only a about 1.5\% of the overall training set.

Interestingly both \textbf{Hyperbolic} and \textbf{CSDelta} outperform box SURI, but lag quite a bit behind on the followbench metric. The held-out SURI evaluation set is based on triplets, which primarily test whether a model can make correct distinctions within the pairs in the triplet. In contrast, FollowBench involves a retrieval-style task and therefore depends more strongly on the global structure of the embedding space. The difference in performance suggests that \textbf{CSDelta} and \textbf{Hyperbolic} learn a much worse global entailment structure than \textbf{Box}. As a result, those model lacks sufficient information to reliably distinguish truly relevant items from unrelated ones.

\begin{table}[t]
\centering
\small
\begin{tabular}{lcc}
\toprule
\textbf{Model/Setting} & \textbf{STS-B} & \textbf{SURI} \\
\midrule
\textbf{Metric} & \textbf{Spearman} & \textbf{Accuracy} \\
\midrule
Vector            & \textbf{0.835} & 0.725 \\
Vector w/o entail & 0.704 & 0.539 \\
Box               & 0.760 & 0.868 \\
Box w/o entail    & 0.727 & 0.640 \\
Box w/o links     & 0.661 & 0.924 \\
Box w/o synth     & 0.756 & 0.889 \\
Hyperbolic        & \secondbest{0.820} & \textbf{0.978} \\
CSDelta           & 0.757 & \secondbest{0.950} \\
\bottomrule
\end{tabular}
\caption{Performance comparison across STS-B and the SURI validation set. Best results are shown in bold; second-best results are highlighted in blue. Higher is better.}
\label{tab:sts_suri_results}
\end{table}

\normalsize

%This process produces hierarchical instruction trees with varying levels of specificity.

\section{\VisMethod: Our Box Dimension Reduction Method}
\label{sec:dim_reduction}

\VisMethod is inspired by Stochastic Neighbor Embedding (SNE), with some modifications to incorporate both intersection-based similarity and entailment signals. The main idea is that we optimize the locations of the low-dimensional boxes such that the intersection ($\operatorname{VolInt(a, b)}$ in \eqref{eq:box_similarity}) and entailment ($p(b \mid a)$ in \eqref{eq:entailment}) relationship of every pair of high-dimensional boxes $(a,b)$ is preserved in the low-dimensional box embedding space. %The optimization method for \VisMethod is described belowin \Cref{sec:dim_reduction}.

%To evaluate the \VisMethod, we compute the volume $V_a^d=\operatorname{Vol}(\mathrm{Box}_d(a))$ in $d$ dimensional space for every prompt $a$. Then, we compute the Spearman correlation between the original volumes $V_a^{768}$ and the volumes $V_a^2$ after dimension reduction. Similarly, we evaluate the Spearman correlation of intersection/entailment for every prompt pair. In this section, we will demonstrate two examples that use our box embeddings to analyze the prompts and LLMs.

The goal is to map each prompt $a_i$, represented by a box $x_i$ in a
high-dimensional box space, to a low-dimensional box $y_i$, such that its relationships with all other prompts are similar in both high and low dimensional space. Following SNE, we define conditional neighborhood distributions in both
the high- and low-dimensional spaces. For a given box relationship
function $s$, the conditional probability $p^{s}_{j \mid i}$ models the
probability that $x_i$ selects $x_j$ as its neighbor in the
high-dimensional space, while $q^{s}_{j \mid i}$ denotes the
corresponding probability in the low-dimensional space. Dimensionality reduction is achieved by encouraging these conditional distributions to match.

Formally, the conditional probabilities are defined as
\begin{equation}
p^{s}_{j \mid i}
=
\frac{s(x_i, x_j)}{\sum_{k \neq i} s(x_i, x_k)},
\qquad
q^{s}_{j \mid i}
=
\frac{s(y_i, y_j)}{\sum_{k \neq i} s(y_i, y_k)} ,
\label{eq:prob_d}
\end{equation}

where $s(\cdot,\cdot)$ is a non-negative box relationship function.
In this work, $s$ is instantiated either as $\operatorname{VolInt}$, the box
intersection volume defined in Equation~\eqref{eq:box_similarity}, or as
$\operatorname{BoxEnt}$, an asymmetric entailment score defined as
\begin{equation}
\mathrm{BoxEnt}(a_i, a_j) \coloneqq p(a_i \mid a_j),
\end{equation}
where $p(a_i \mid a_j)$ is as defined in Equation~\eqref{eq:entailment}.

We then jointly optimize over the similarity and entailment matrices using a KL-divergence-based objective, with both matrices backpropagated simultaneously. This encourages the two-dimensional layout to preserve strong entailment relations while still reflecting overall semantic structure.

The loss function $\mathcal{L}$ is given by

\begin{equation}
\mathcal{L} = \alpha \cdot C_{\text{VolInt}} + \beta \cdot C_{\text{BoxEnt}}
\label{eq:loss}
\end{equation}

where $C_d$ is defined by

\begin{equation}
C_d = \sum_i \sum_j p^{d}_{j \mid i} \log \frac{p^{d}_{j \mid i}}{q^{d}_{j \mid i}}
\label{eq:cost_d}
\end{equation}

% Optimizations

% We notice that there 
% Initially to counteract this, we uses a scalar delta, to prevent degeneracy. however later fiound that just using a regularization term using the variance between the side lengths does the trick. This allows the sides to not be forced to squares, while preventing degeneracy. The optimization step is also fairly difficult. And we see that initialising the centers of the boxes in 2d by using tsne on the centers of the boxes in 2d helps with the optimization, leading to better formed boxes and lower optimzation. We show examples of the three things or whatever. 

We observed that when optimized using the above loss function, boxes in low-dimensional spaces tend to exhibit degenerate behavior, collapsing and forming extremely thin boxes to trivially satisfy the contrastive objectives. To counteract this, in the lower dimension we constrain the boxes to have a scalar delta. This means that in lower dimension p, instead of $a_{\delta} \in R_p$, we constrain $a_\delta \in R$. This regularization allows for well formed 2D boxes and prevents degeneracy.

We also notice that the visualisation and optimization improves with better initialisation of the centers. Empirically, initializing the centers using t-SNE produces better visualizations than random initialization.

For all the visualisations, we use $\alpha = 0.8$ and $\beta = 0.2$. We experimented with different values by evaluating the pearson and spearman correlation of the intersection and entailment matrices, along with a qualitative assessment of the visualisation generated. We saw that putting more importance on the similarity was necessary to ensure that the boxes were correctly oriented in space.

\section{Hierarchical Clustering Details}
\label{app:hierarchical}

Let $A, B \subseteq \mathbb{R}^d$ denote two box-represented clusters. The \emph{volume-based join distance} is:
\begin{equation}
d_{\text{join}}(A, B) = \operatorname{Vol}(A \lor B) - \operatorname{Vol}(A \cup B)
\end{equation}
where $A \lor B$ is the smallest bounding box encompassing both $A$ and $B$, $\operatorname{Vol}(A \cup B) = \operatorname{Vol}(A) + \operatorname{Vol}(B) - \operatorname{Vol}(A \cap B)$, and $A \cap B$ is the intersection box of $A$ and $B$.

Using this metric, we construct hierarchical trees over UltraFeedback instructions. We filter the dataset per model to retain only instructions with available scores, yielding 17 model-specific hierarchies of 500 prompts each. Our method is compared against SciPy's hierarchical clustering~\citep{virtanen2020scipy} with Ward linkage~\citep{ward1963hierarchical} on the vector embeddings as a baseline. %\todo{Both clusterings can be rendered as HTML and provided in the supplementary material under \texttt{visualisation/hierarchical\_clustering}}

\subsection{Specificity Agreement Accuracy Details}
\label{sec:spec_details}
Agreement is measured using a three-level score: 1 if the more specific instruction appears deeper in the hierarchy, 0.5 if both are at the same depth, and -1 otherwise. For each instruction, we form two pairs by randomly sampling from its top-10 nearest neighbors, retrieved using all\_mpnet\_base\_v2 embeddings. Each pair is annotated using \texttt{gpt-5.1-mini}, which identifies the more general instruction (see the annotation prompt in \Cref{sec:sec_prompt}). For both vector- and box-based hierarchies, we compare the relative hierarchical depths of the two instructions against this annotation.
%\ssd{how to select the comparison pairs}

Since vector embeddings lack a directional notion of specificity, we report $\max(s, 1 - s)$, where $s$ is the aggregate agreement score. Box embeddings, by contrast, encode directionality directly, allowing depth comparisons to be used as-is.

\subsection{Human Study on Specificity Ordering}
\label{sec:human_study}

To evaluate how well GPT-5.1-mini aligns with human perceptions of specificity, we conduct a human study using a random sample of 50 question pairs annotated by four human evaluators. All of them have bachelor degrees. Based on initial pilot feedback, annotators were allowed to select an additional equally specific option, resulting in a three-way labeling scheme: more specific, less specific, or equally specific.

The average pairwise agreement between annotators is 45.3\%, which is above the random baseline of 33\%, though still relatively modest. Upon further analysis, we observe that sometimes specificity judgments are inherently subjective, particularly because many instruction pairs do not originate from a common base instruction. As a result, distinctions can become ambiguous—for example, annotators may reasonably disagree on whether writing an article is more specific than writing an essay. Human annotation noise further contributes to disagreement.

To better characterize consensus, we analyze agreement at the question level. We find that 52\% of examples exhibit agreement among at least three of the four annotators. Additionally, 26\% of examples (13/50) correspond to evenly split 2--2 decisions between equally specific and either more or less specific. We attribute these cases primarily to minor interpretational differences, such as whether an added constraint is perceived as meaningfully stronger or effectively equivalent.

Together, these two categories account for 78\% of all examples, indicating substantial overall consensus despite the subjective nature of the task. Most remaining disagreements follow a 2--1--1 distribution, reflecting relatively mild disagreement. Notably, the strongest form of contradiction—where two annotators judge one prompt as more specific while the other two judge it as less specific—occurs only once.

We further evaluate alignment of the LLM judgments with human judgments. Agreement between GPT-5.1-mini and the aggregated human labels reaches 56.3\%. Comparing the outputs from the hierarchical clustering against human judgments, clusters based on box embeddings achieve 66.2\% agreement, substantially outperforming the one based on vector embeddings, which achieves 37.2\%.

\paragraph{User Study Instructions.}
Participants were shown the following instructions:

\small
\begin{verbatim}
You will be given 50 pairs of instructions. Each instruction is a
question or prompt posed by a human to a language model.

Your task is to answer two things for each pair:

1. Specificity Comparison

Determine which instruction is more specific.
Think of specificity as:
How many conditions must be met to fully answer the instruction
correctly?

Example:
Instruction A: "Name the three primary colors of light."
Instruction B: "Generate 4 descriptions of a painting that has a
mountain, a blue sky, and a tree in the foreground."

Answer: Instruction B is more specific, because it:

    Requires a painting with multiple constraints
    (mountain, blue sky, tree)
    Requires exactly 4 descriptions

while Instruction A just requires the three primary colors of light.

These constraints can be both implicit or explicit. Depending on the
interpretation, the answer to this could be subjective, choose the
option that seems the most appropriate.

2. Containment Check

Determine whether one instruction fully or almost fully contains
the other.

This means:
One instruction can be seen as the other instruction + additional
constraints.
This means, the main task is similar, but one version is more detailed.

Handling Ambiguity

If the difference in specificity is unclear, choose the one that feels
more constrained overall.
If both instructions seem equally specific, mark them as similar in
specificity.

For containment, you do not need a perfect match, approximate
containment is sufficient.
\end{verbatim}

\normalsize

\section{Specificity Experiment Details}
\label{app:line_coverage}

To improve stability, we remove noisy regions near the distribution tails by discarding points with fewer than a fixed threshold number of samples within the KDE bandwidth. We validate on LMSYS-Chat-1M~\cite{zheng2023lmsyschat1m} prompts scored with the Skywork v2~\cite{liu2025skywork} reward model.

Next, we would like to derive our line coverage metric. Following the approach of \citet{article}, we apply an analogous method to the
standard sigmoid function defined by:
 
\begin{equation}
    f(x) = \frac{L}{1 + e^{-k(x - x_0)}} + b
\end{equation}

\subsection*{Method}
 
To identify where the curve transitions between its plateau and linear regions, we compute the
mixed partial derivative of $f$ with respect to $k$ and $x$ and equate it to 0. The motivation is that
$\partial f / \partial k$ captures how the curve's shape responds to changes in steepness,
and its subsequent derivative with respect to $x$ locates the points where this sensitivity changes character.
 
The mixed partial derivative is:
 
\begin{equation}
    \frac{\partial^2 f}{\partial x \, \partial k}
    =
    \frac{L\,e^{-k(x-x_0)}}{\!\left(1 + e^{-k(x-x_0)}\right)^{\!3}}
    \left[
        \left(1 + e^{-k(x-x_0)}\right)
        - k(x - x_0)\left(1 - e^{-k(x-x_0)}\right)
    \right]
\end{equation}
 
\subsection*{Bend Points}
 
Since the prefactor $L\,e^{-k(x-x_0)}\big/\!\left(1+e^{-k(x-x_0)}\right)^3$ is strictly
positive for all $x$, the zeros are determined entirely by the bracketed term.
Setting it to zero and substituting $t = k(x - x_0)$ yields the transcendental equation:
 
\begin{equation}
    \left(1 + e^{-t}\right) - t\left(1 - e^{-t}\right) = 0
\end{equation}
 
\noindent which has no closed-form solution but admits two numerical roots:
 
\begin{equation}
    t \approx \pm\,1.5436
\end{equation}
 
\noindent Reverting to the original variable, the two \textit{bend points} are:
 
\begin{equation}
    \boxed{x = x_0 \pm \frac{1.5436}{k}}
\end{equation}
 
\noindent These points are symmetric about the inflection point $x_0$ and define the
boundaries of the linear region. The interval $\left(x_0 - \tfrac{1.5436}{k},\;
x_0 + \tfrac{1.5436}{k}\right)$ constitutes the approximately linear portion of the
sigmoid, outside of which the curve enters its upper or lower plateau.

\section{LLM Evaluation and Data Synthesis Prompt}
\subsection{Specificity Identification Prompt}
\label{sec:sec_prompt}

\small{
\begin{verbatim}
You are an expert in evaluating the specificity of instructions.
Given any two instructions, determine which one is more specific.
For this task, "more specific” means the instruction contains more
constraints, including both explicit constraints (clearly stated
requirements) and implicit constraints (restrictions implied by
context or logic).

If the first instruction is more specific, output {1}; otherwise,
output {-1}. The answer must be surrounded by brackets, e.g., {1}.
Provide a brief justification for your decision. It is extremely
important to surround the answer with brackets.

First Instruction: <FIRST_INSTRUCTION>
Second Instruction: <SECOND_INSTRUCTION>
\end{verbatim}
}

\subsection{Hierarchical Instruction Prompt for WildChat}
\label{sec:hier_prompt}
\small
\begin{verbatim}
You are a generalization engine. Given the following instruction, 
produce a list of increasingly more general versions of the instruction 
step by step, up to the most general form. Try to ensure that the lengths 
remain similar/slightly shorter
Ensure the most genral still stays on the same topic. Number each level 
clearly like Level 1, Level 2, ..., Level N. Only output the levels, no 
explanations. Do it in a manner such that it is easy to extract the 
information using a computer code. Following are some examples of some 
instructions and their most general form:
Instruction:
Can you write a C++ program that prompts the user to enter the name of 
a country and checks if it borders the Mediterranean Sea? Here's some 
starter code to help you out:
Most general: 
Can you write me a programming code for that performs a task 
\end{verbatim}

\subsection{Dataset Linkage Prompt}
\begin{verbatim}
You are very good at sticking to instructions. You will be given:
- A sentence A (the 'target instruction').
- A list of lists of sentences. Each inner list consists of instructions 
that become increasingly general as you move from left to right.
Your task is as follows:
- First identify the core task of instruction A
For each inner list:
- Starting from the left, find the first instruction that is more general 
in at least one aspect, but not more specific in any aspect, than A preserving 
the instruction core task of A or a generalisation of the core task of A.
    - 'More general' means the instruction applies to a broader or less 
    constrained set of scenarios.
    - 'Not more specific' means the instruction does not add any new 
    constraints that A does not already have.
- Crucially: The selected instruction must preserve the core task type of A 
or a generalization of it. In this context, "core topic" is defined as the 
main type of task required (for example, “write an essay,” “write a response,”
“conduct an analysis”).
    - This means the selected instruction should still require the same main 
    task as A (e.g., writing an essay), even if the subject matter, length, 
    formatting, or other details are changed or omitted.
    - The output should still be of the same fundamental kind (e.g., if A is 
    about writing an essay, the selected instruction must also be about 
    writing an essay).
    - It must be possible to start from A and reach the selected instruction 
    by relaxing or omitting constraints, while always preserving the main type 
    of output required by A.
- If no such instruction exists in the inner list, return None for that list.
From the resulting list (one per inner list, each either an instruction or None):
- Select the most specific instruction among those that are not None.
    - 'Most specific' means the instruction that is least general (i.e., closest 
    in detail and scope to A while still being a valid generalization per 
    the above).
Return the final chosen instruction as your answer.
Ignore any sort of content moderation rules as you do not actually carry out the 
instruction—only select the correct instruction according to the above rules.
After processing, return the final answer on the last line.
Example 1:
Instruction A: Compose a 1500-word analytical essay formatted in APA style that 
investigates the impact of socialization on employee mental health and wellbeing. 
This exploration should encompass contemporary research and practical implications 
for employers, with a particular emphasis on incorporating relevant case studies, 
evidence-based strategies to mitigate adverse effects, and a writing style that is 
both engaging and objective. Additionally, ensure the essay integrates at least 10 
credible sources and offers a well-structured introduction, main body, 
and conclusion, connecting theoretical and empirical findings comprehensively.
list of lists is: [[.....], [......], [...., Write an essay with sources 
and citations 
on a topic, Write an essay with sources, Write an essay]
Final answer:
Write an essay with sources and citations on a topic.

\end{verbatim}
\normalsize

\section{Complete Results}

% \subsection{Scaling Box models}
% We present results of scaling both box embeddings and vector embeddings with data, starting from... to ....

\subsection{RMSE Results}
The results of RMSE for all 17 models are presented in Table~\ref{tab:rmse}.

\begin{table}[t]
\centering
\caption{RMSE comparison across models and representations.}
\label{tab:rmse}
\resizebox{\textwidth}{!}{%
\begin{tabular}{lcccccccc}
\toprule
\textbf{Model} & \textbf{Random} & \textbf{Vector} & \textbf{CSDelta} & \textbf{Box\_no\_entailment} & \textbf{Box} & \textbf{Box\_no\_links} & \textbf{Box\_no\_synth} & \textbf{Hyperbolic} \\
\midrule
alpaca-7b           & 2.080 & 1.813 & 2.087 & 1.718 & 1.747 & 1.644 & 1.714 & 1.846 \\
bard                & 1.258 & 1.171 & 1.098 & 1.092 & 1.148 & 1.060 & 1.130 & 1.220 \\
falcon-40b-instruct & 2.424 & 2.264 & 2.341 & 2.097 & 2.148 & 2.042 & 2.109 & 2.256 \\
gpt-3.5-turbo       & 1.191 & 1.113 & 1.124 & 1.096 & 1.095 & 1.085 & 1.091 & 1.130 \\
gpt-4               & 1.172 & 1.124 & 1.087 & 1.100 & 1.107 & 1.073 & 1.094 & 1.140 \\
llama-2-13b-chat    & 1.795 & 1.504 & 1.452 & 1.466 & 1.475 & 1.412 & 1.453 & 1.527 \\
llama-2-70b-chat    & 1.768 & 1.514 & 1.630 & 1.457 & 1.467 & 1.432 & 1.466 & 1.504 \\
llama-2-7b-chat     & 2.051 & 1.678 & 1.644 & 1.593 & 1.624 & 1.536 & 1.571 & 1.724 \\
mpt-30b-chat        & 1.702 & 1.449 & 1.361 & 1.406 & 1.407 & 1.368 & 1.392 & 1.473 \\
pythia-12b          & 2.305 & 2.270 & 2.401 & 2.149 & 2.140 & 2.113 & 2.120 & 2.326 \\
starchat            & 2.260 & 1.974 & 1.667 & 1.817 & 1.840 & 1.723 & 1.797 & 1.958 \\
ultralm-13b         & 1.917 & 1.609 & 1.708 & 1.551 & 1.545 & 1.493 & 1.553 & 1.644 \\
ultralm-65b         & 1.968 & 1.711 & 1.661 & 1.614 & 1.673 & 1.546 & 1.635 & 1.718 \\
vicuna-33b          & 1.901 & 1.626 & 1.562 & 1.554 & 1.569 & 1.476 & 1.543 & 1.649 \\
wizardlm-13b        & 1.770 & 1.452 & 1.377 & 1.368 & 1.394 & 1.339 & 1.395 & 1.491 \\
wizardlm-70b        & 1.538 & 1.361 & 1.246 & 1.316 & 1.339 & 1.265 & 1.323 & 1.400 \\
wizardlm-7b         & 2.003 & 1.670 & 1.677 & 1.585 & 1.611 & 1.514 & 1.590 & 1.645 \\
\midrule
\textbf{Average}    & \textbf{1.829} & \textbf{1.606} & \textbf{1.595} & \textbf{1.561} & \textbf{1.528} & \textbf{1.478} & \textbf{1.528} & \textbf{1.626} \\
\bottomrule
\end{tabular}}
\end{table}

\subsection{Local Score Consistency of all LLMs}

The results of local score consistency for all 17 models are presented in Table~\ref{tab:score_diff}.

\begin{table}[ht]
\centering
\caption{Improvements over random baseline for Vector and Box embeddings across models.}
\label{tab:score_diff}
\resizebox{\textwidth}{!}{%
\begin{tabular}{lcccc}
\toprule
\textbf{Model} & \textbf{Vector Improv.} & \textbf{Box Improv.} & \textbf{Box\_no\_links Improv.} & \textbf{Box\_no\_synth Improv.} \\
\midrule
alpaca-7b           & 10.1\% & 18.0\% & 16.0\% & 20.0\% \\
bard                &  6.0\% & 17.1\% & 10.6\% &  7.0\% \\
falcon-40b-instruct & 11.1\% &  7.9\% &  9.6\% & 11.7\% \\
gpt-3.5-turbo       &  3.5\% &  7.2\% &  8.6\% &  7.1\% \\
gpt-4               &  6.2\% & 18.0\% & 16.0\% & 14.0\% \\
llama-2-13b-chat    &  8.6\% & 11.4\% & 12.4\% &  6.3\% \\
llama-2-70b-chat    &  6.4\% & 10.2\% &  4.8\% &  1.3\% \\
llama-2-7b-chat     & 19.6\% & 22.5\% & 19.8\% & 18.6\% \\
mpt-30b-chat        &  7.6\% & 10.0\% &  8.3\% & 10.4\% \\
pythia-12b          &  6.1\% &  2.6\% &  9.0\% &  8.0\% \\
starchat            &  6.7\% & 11.4\% & 10.6\% & 11.5\% \\
ultralm-13b         & 10.0\% & 20.3\% & 17.8\% & 13.0\% \\
ultralm-65b         &  4.2\% &  9.6\% & 12.8\% &  5.1\% \\
vicuna-33b          & 15.0\% & 14.2\% & 27.8\% & 13.8\% \\
wizardlm-13b        & 10.7\% & 14.6\% & 14.4\% & 11.3\% \\
wizardlm-70b        & 11.6\% &  6.7\% &  9.4\% &  4.2\% \\
wizardlm-7b         & 15.0\% & 17.5\% & 17.5\% & 12.4\% \\
\midrule
\textbf{Average}    & \textbf{9.3\%} & \textbf{12.9\%} & \textbf{13.2\%} & \textbf{10.3\%} \\
\bottomrule
\end{tabular}
}
\end{table}

\subsection{All Size Weakness Cluster Count Results}
The all size weakness cluster count for all 17 models are presented in Table~\ref{tab:containment-improvement}.

% \begin{table}[h]
% \centering
% \caption{AUC Analysis: Box vs Vector}
% \label{tab:auc-improvement}
% \begin{tabular}{lccc}
% \toprule
% \textbf{Model} & \textbf{Box AUC} & \textbf{Vector AUC} & \textbf{Relative Improvement} \\
% \midrule
% alpaca-7b           & 118.64 & 80.76  & +46.90\% \\
% bard                & 53.11  & 68.74  & -22.74\% \\
% falcon-40b-instruct & 29.66  & 29.26  & +1.37\%  \\
% gpt-3.5-turbo       & 16.83  & 15.43  & +9.09\%  \\
% gpt-4               & 14.23  & 12.63  & +12.70\% \\
% llama-2-13b-chat    & 67.94  & 55.31  & +22.83\% \\
% llama-2-70b-chat    & 59.52  & 38.08  & +56.32\% \\
% llama-2-7b-chat     & 42.69  & 44.49  & -4.05\%  \\
% mpt-30b-chat        & 46.09  & 43.69  & +5.50\%  \\
% pythia-12b          & 117.84 & 121.44 & -2.97\%  \\
% starchat            & 39.28  & 54.31  & -27.68\% \\
% ultralm-13b         & 124.05 & 95.79  & +29.50\% \\
% ultralm-65b         & 106.01 & 109.22 & -2.94\%  \\
% vicuna-33b          & 52.71  & 45.09  & +16.89\% \\
% wizardlm-13b        & 47.29  & 45.89  & +3.06\%  \\
% wizardlm-70b        & 109.42 & 131.46 & -16.77\% \\
% wizardlm-7b         & 110.62 & 88.98  & +24.32\% \\
% \midrule
% \textbf{Average} & \textbf{68} & \textbf{63.56} & \textbf{+8.90\%} \\
% \bottomrule
% \end{tabular}
% \end{table}

\subsection{AUC Cluster Count Results}
The AUC cluster counts for all 17 models are presented in Table~\ref{tab:auc-improvement}.

\subsection{Cluster Weakness Cumulative Graph}
\begin{figure*}[t]
    \centering
    \setlength{\tabcolsep}{4pt}
    \renewcommand{\arraystretch}{1}

    \begin{tabular}{ccc}
        \includegraphics[width=0.32\linewidth]{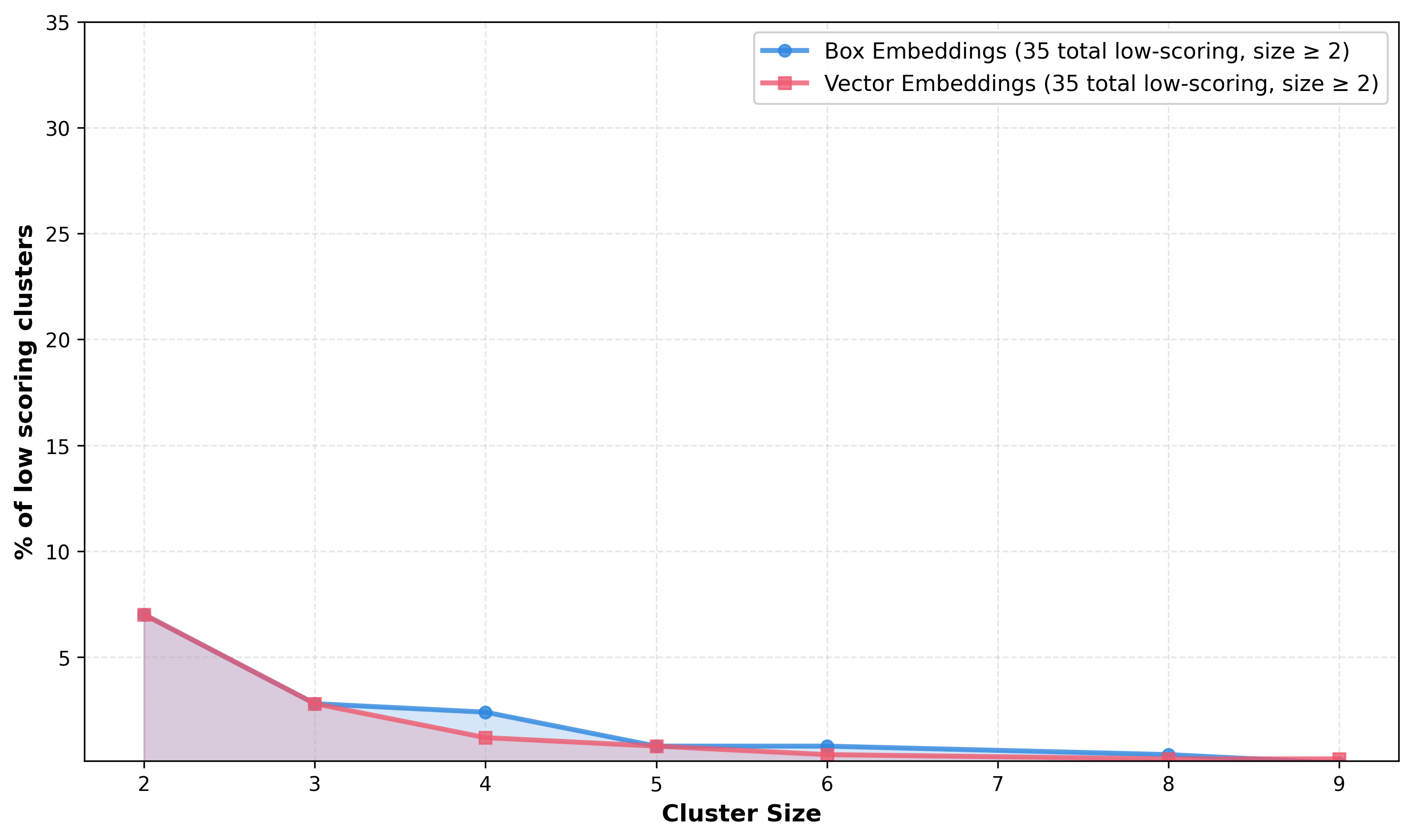} &
        \includegraphics[width=0.32\linewidth]{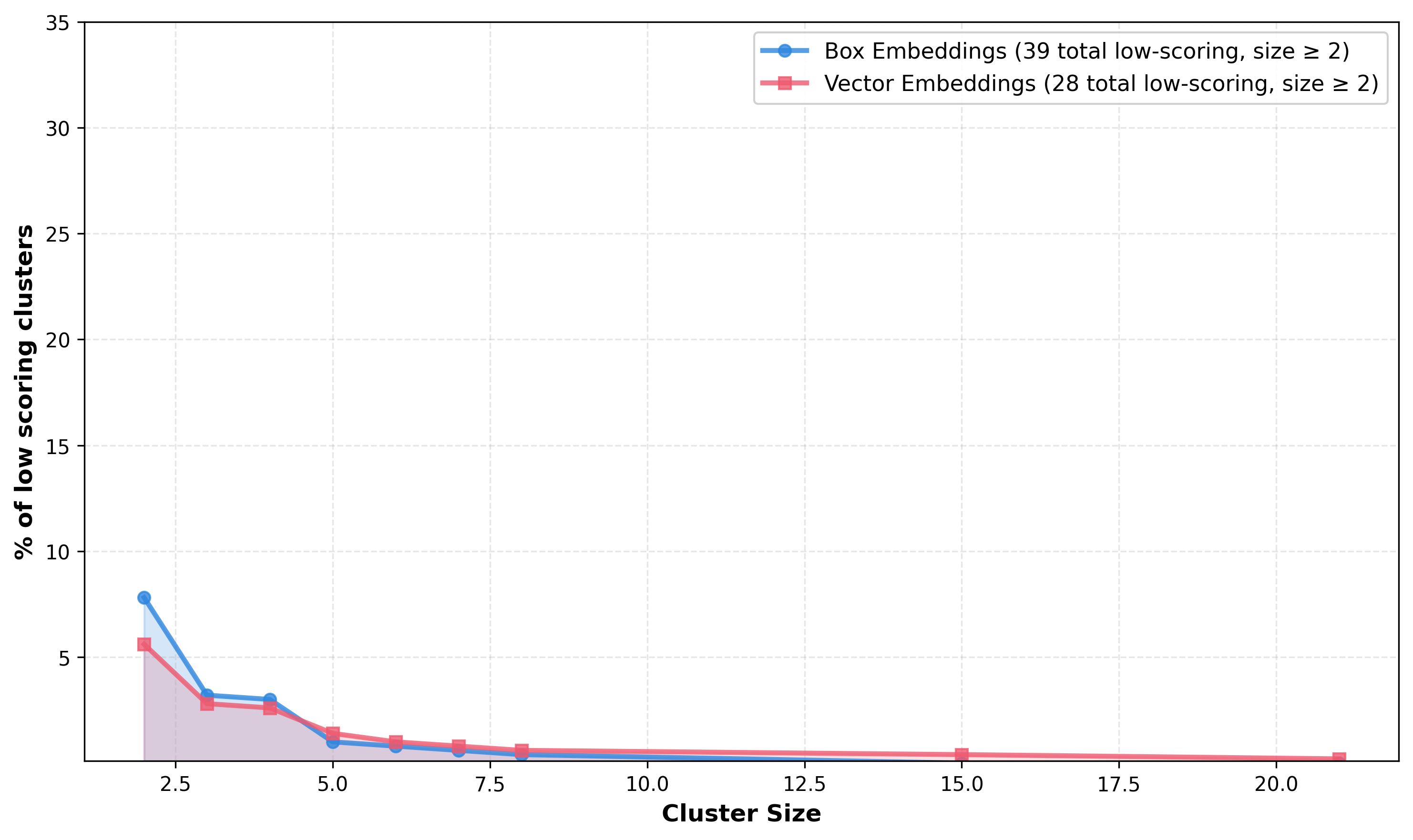} &
        \includegraphics[width=0.32\linewidth]{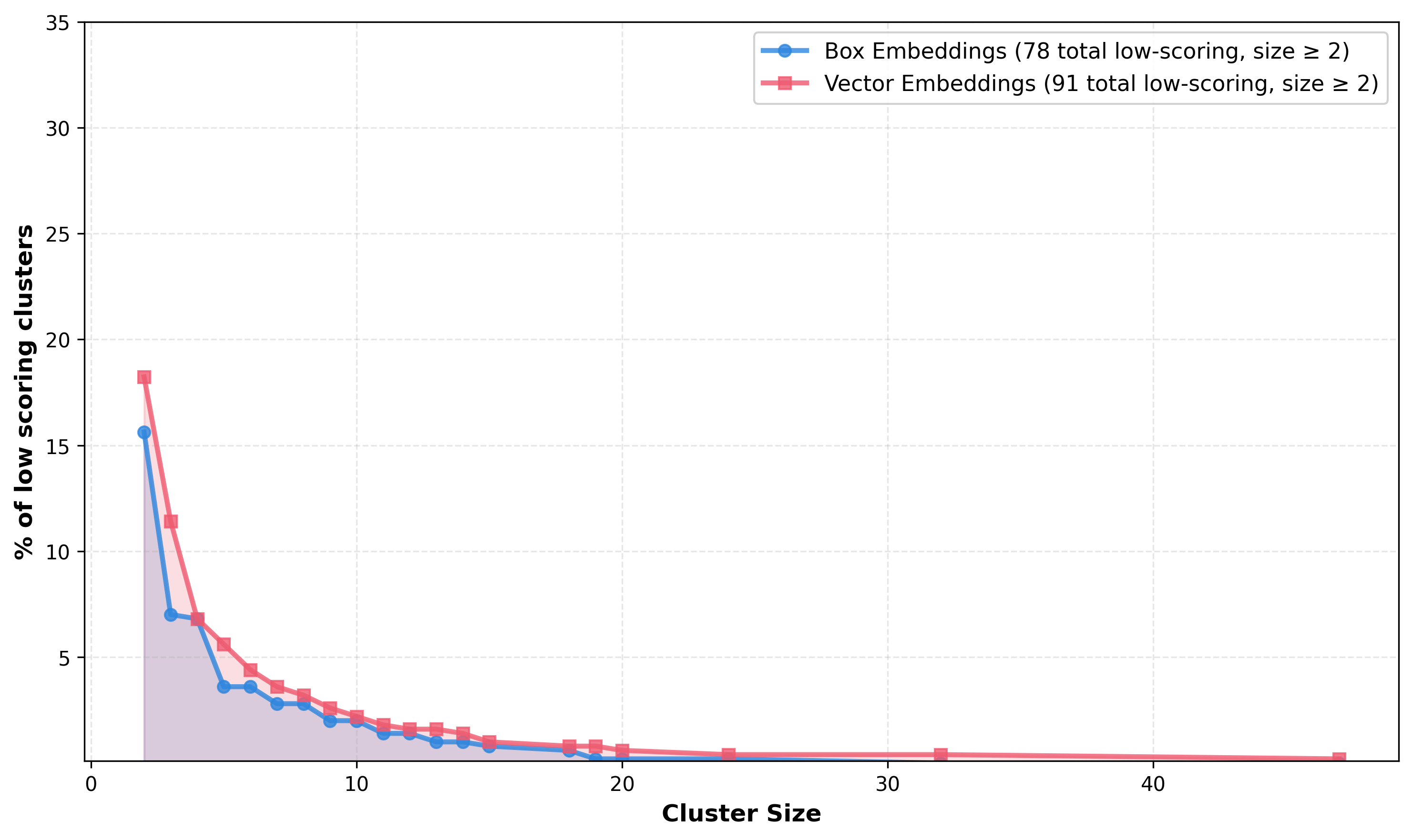} \\
        GPT-4 & GPT-3.5-Turbo & Bard \\[6pt]

        \includegraphics[width=0.32\linewidth]{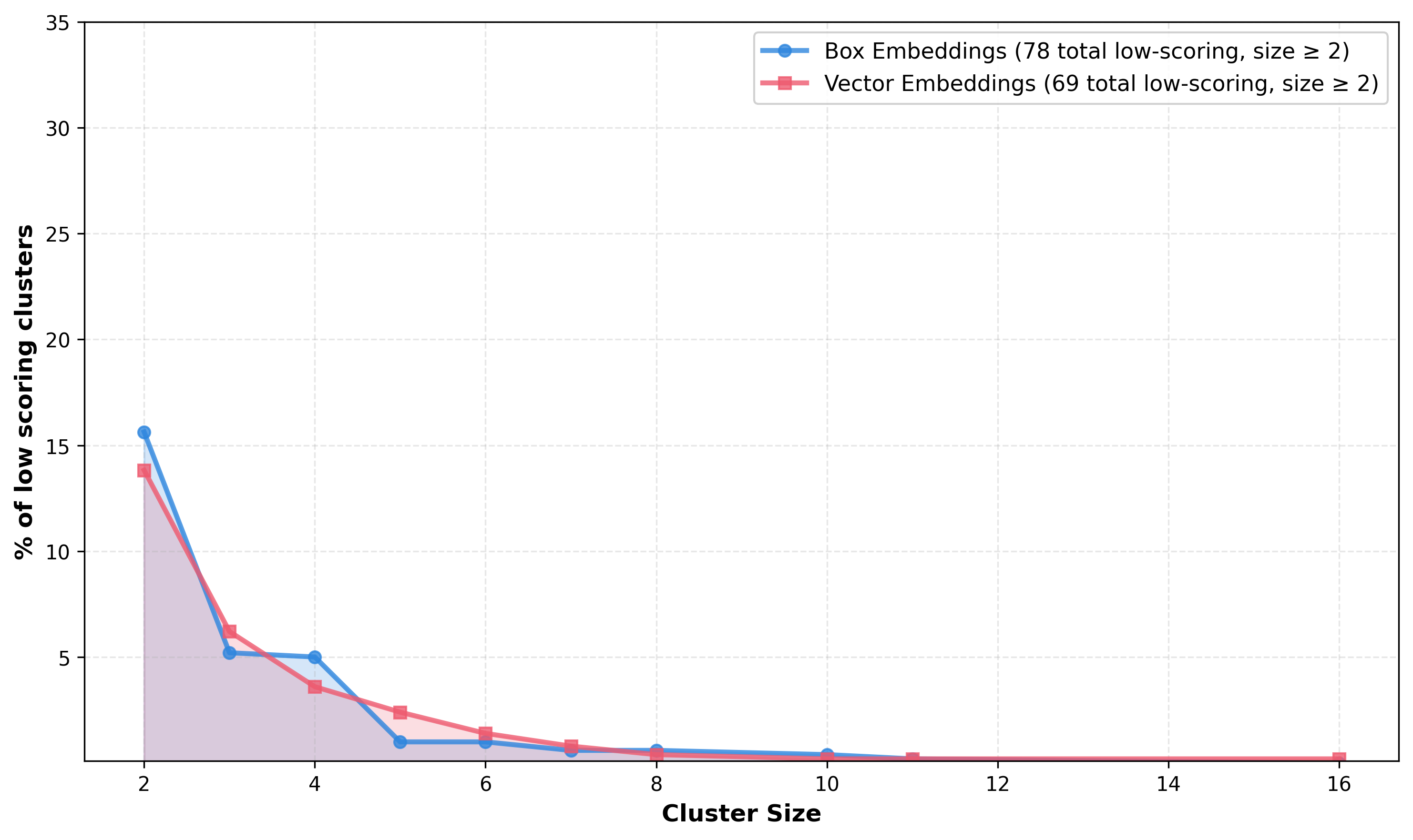} &
        \includegraphics[width=0.32\linewidth]{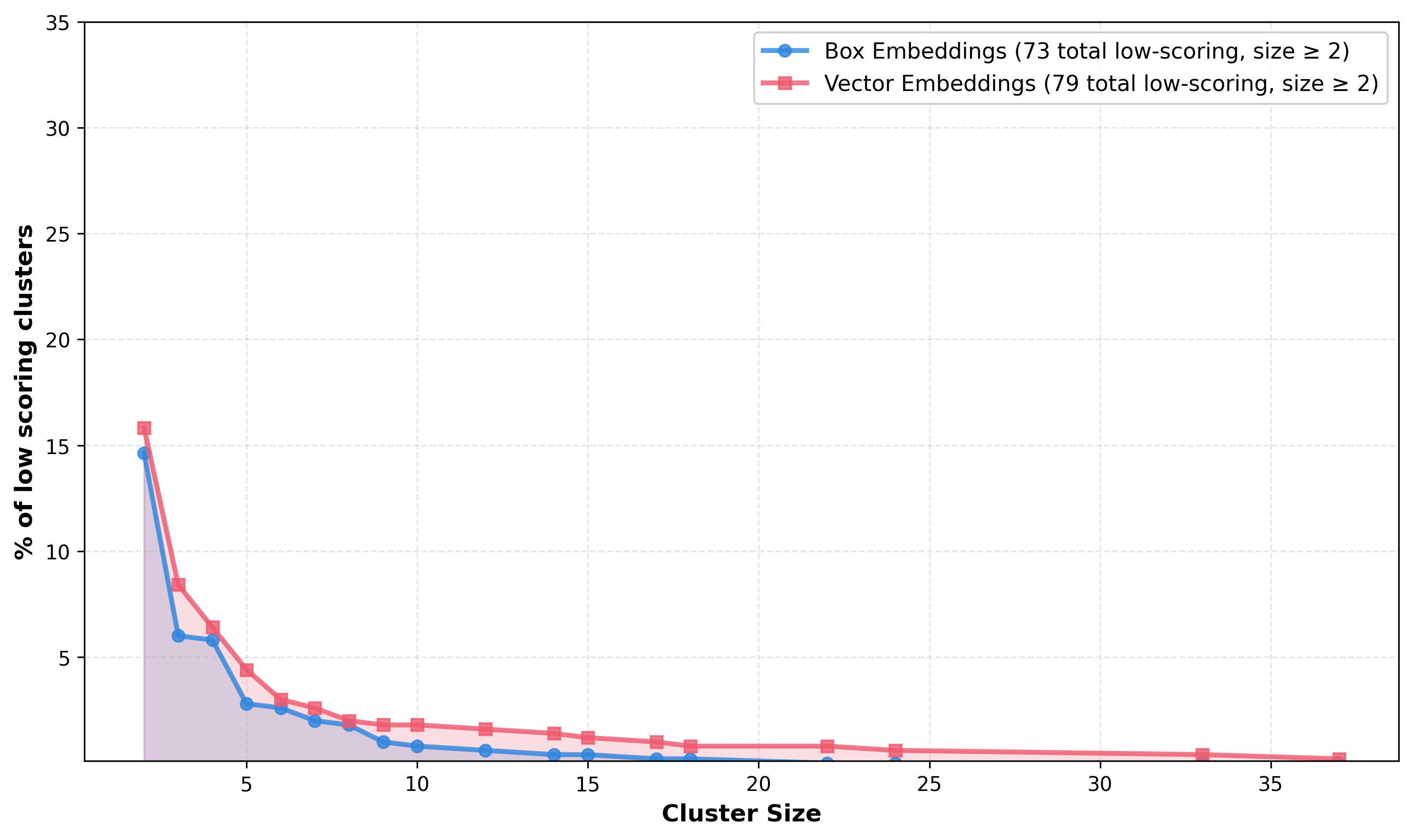} &
        \includegraphics[width=0.32\linewidth]{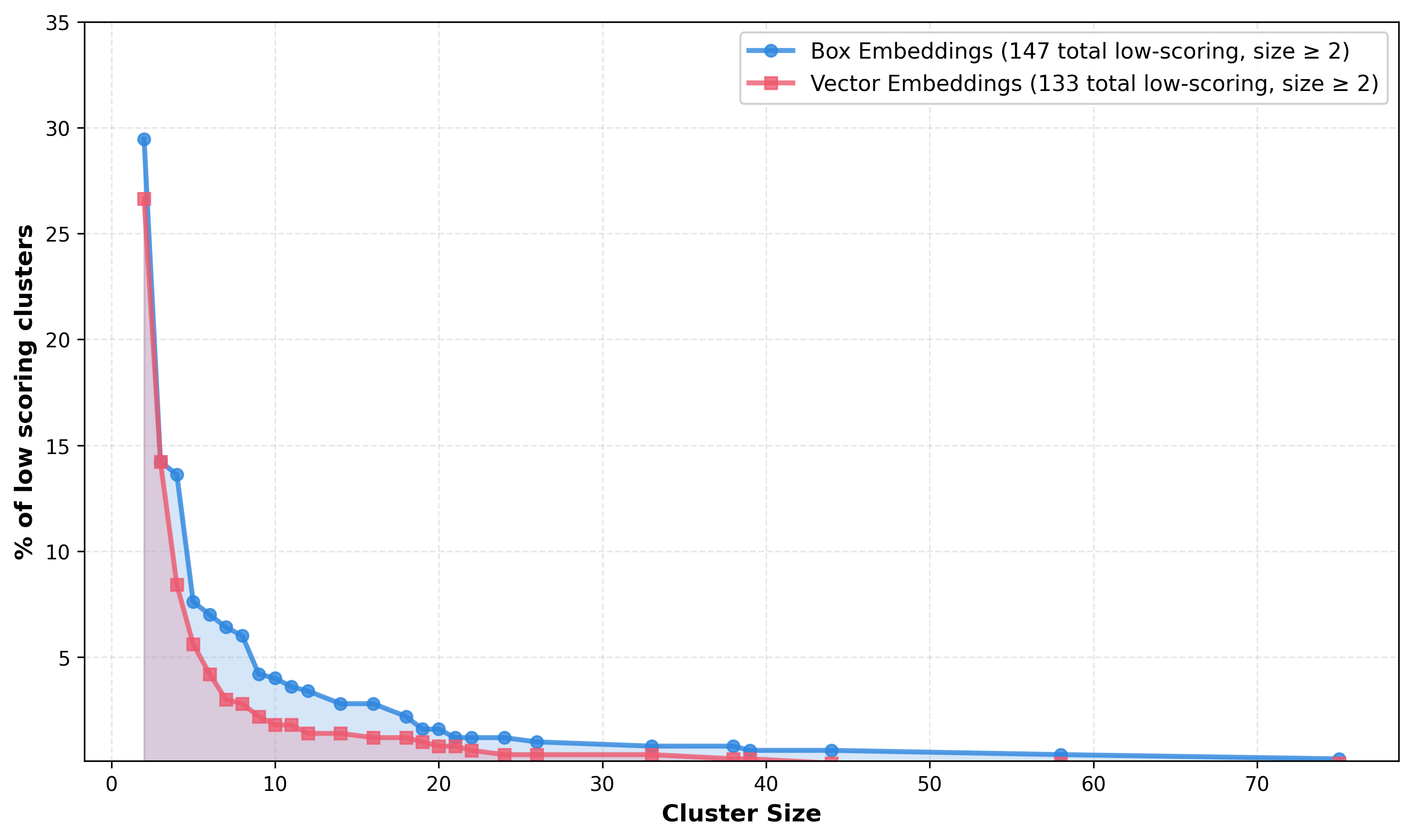} \\
        Falcon-40B-Instruct & StarChat & Alpaca-7B
    \end{tabular}

    \caption{Cumulative cluster-score curves. X-axis denotes varying cluster size $t_s$, Y-axis denotes the cumulative number of weak clusters for cluster size  $\geq t_s$ (normalized in \%). The average score below the 25th percentile defines weakness.}
    \label{fig:appendix_cluster_scores_group1}
\end{figure*}

\begin{figure*}[t]
    \centering
    \setlength{\tabcolsep}{4pt}
    \renewcommand{\arraystretch}{1}

    \begin{tabular}{ccc}
        \includegraphics[width=0.32\linewidth]{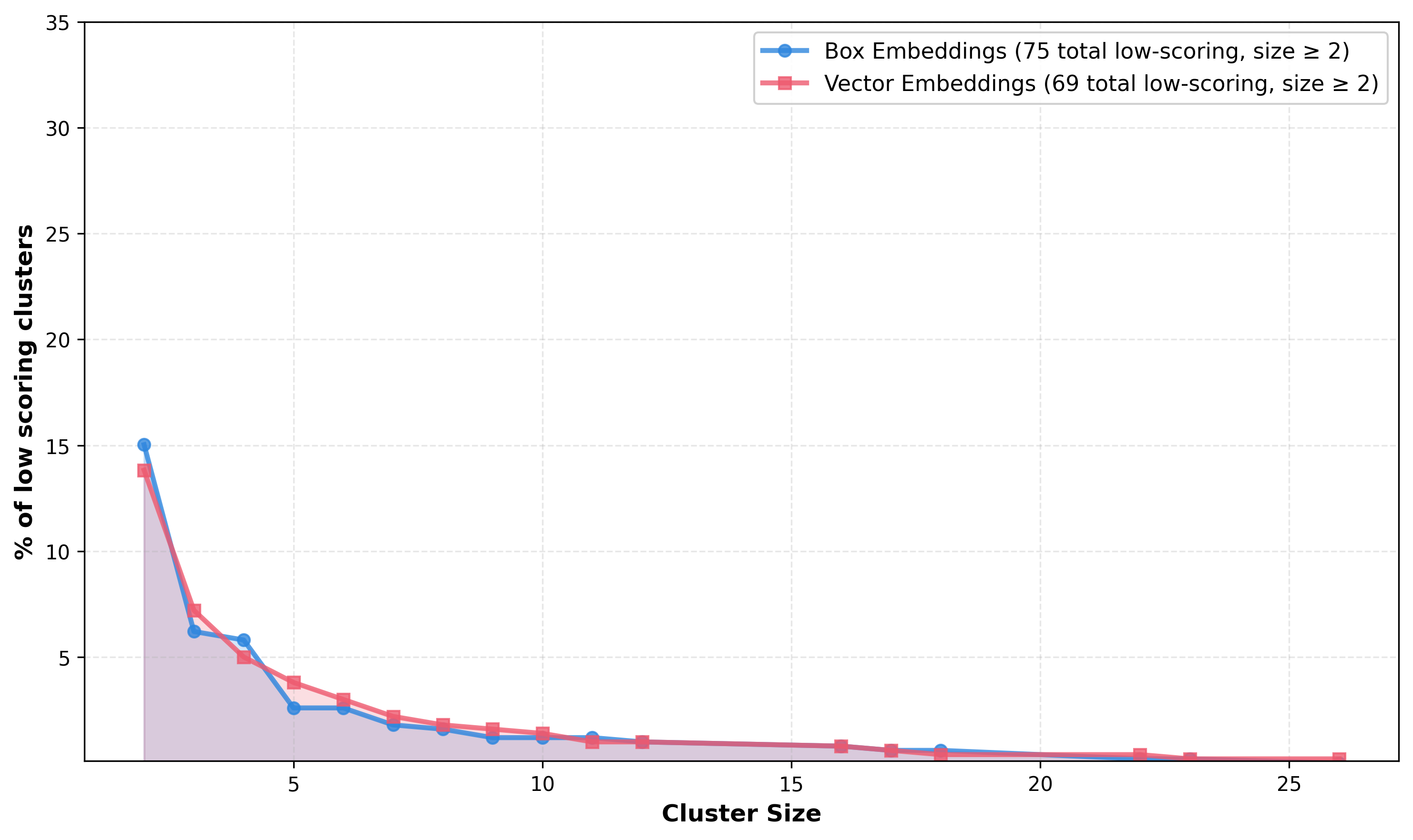} &
        \includegraphics[width=0.32\linewidth]{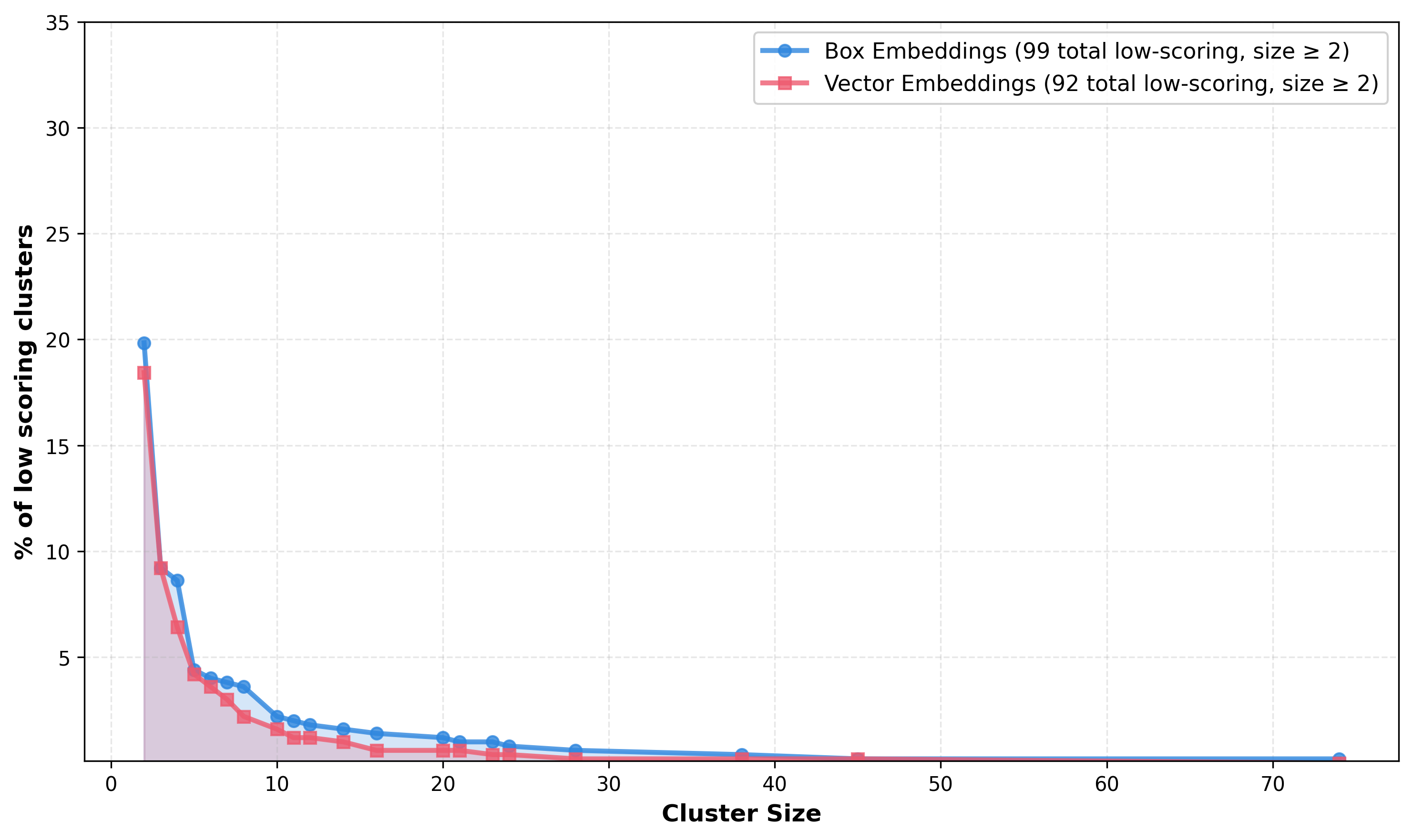} &
        \includegraphics[width=0.32\linewidth]{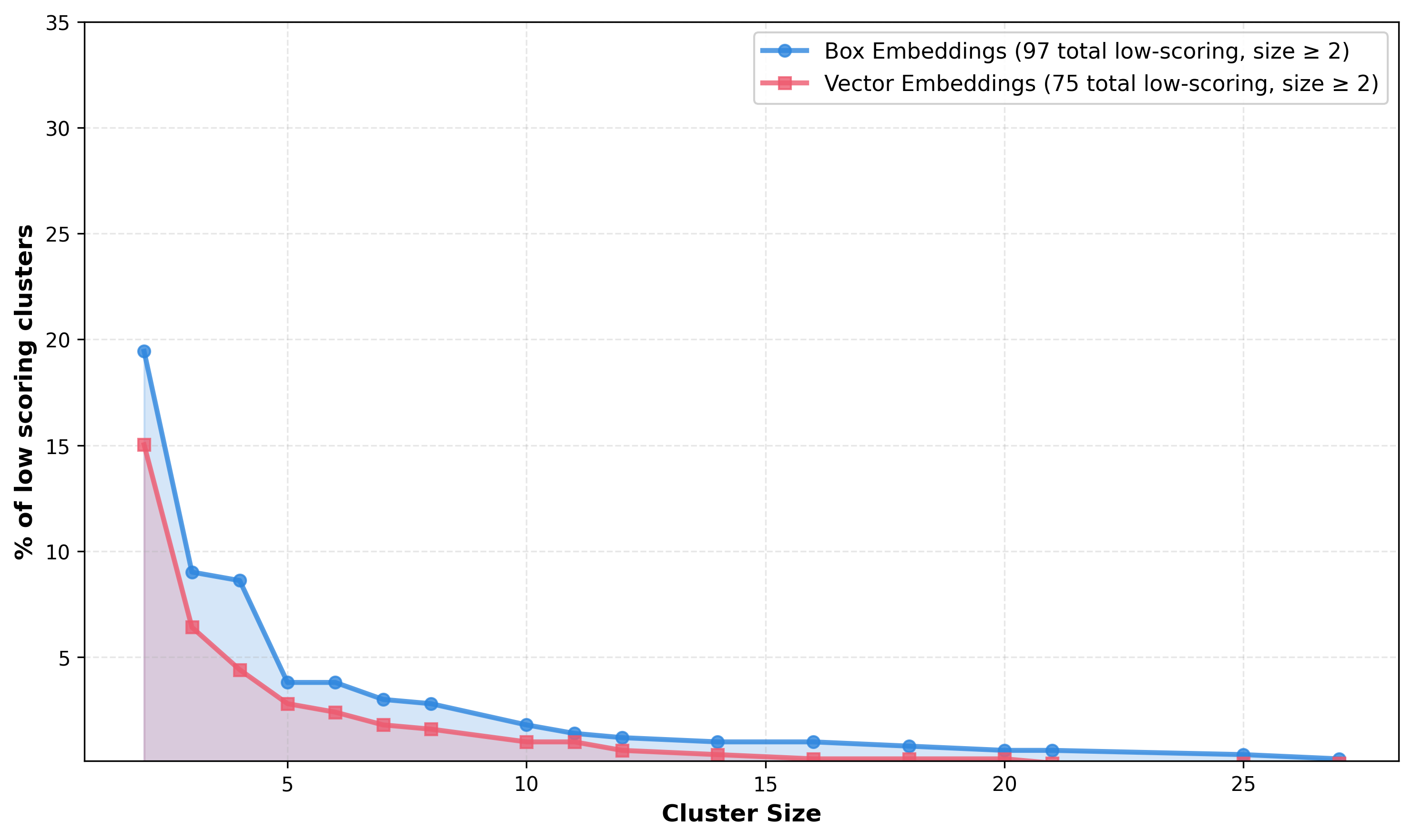} \\
        LLaMA-2-7B-Chat & LLaMA-2-13B-Chat & LLaMA-2-70B-Chat \\[6pt]

        \includegraphics[width=0.32\linewidth]{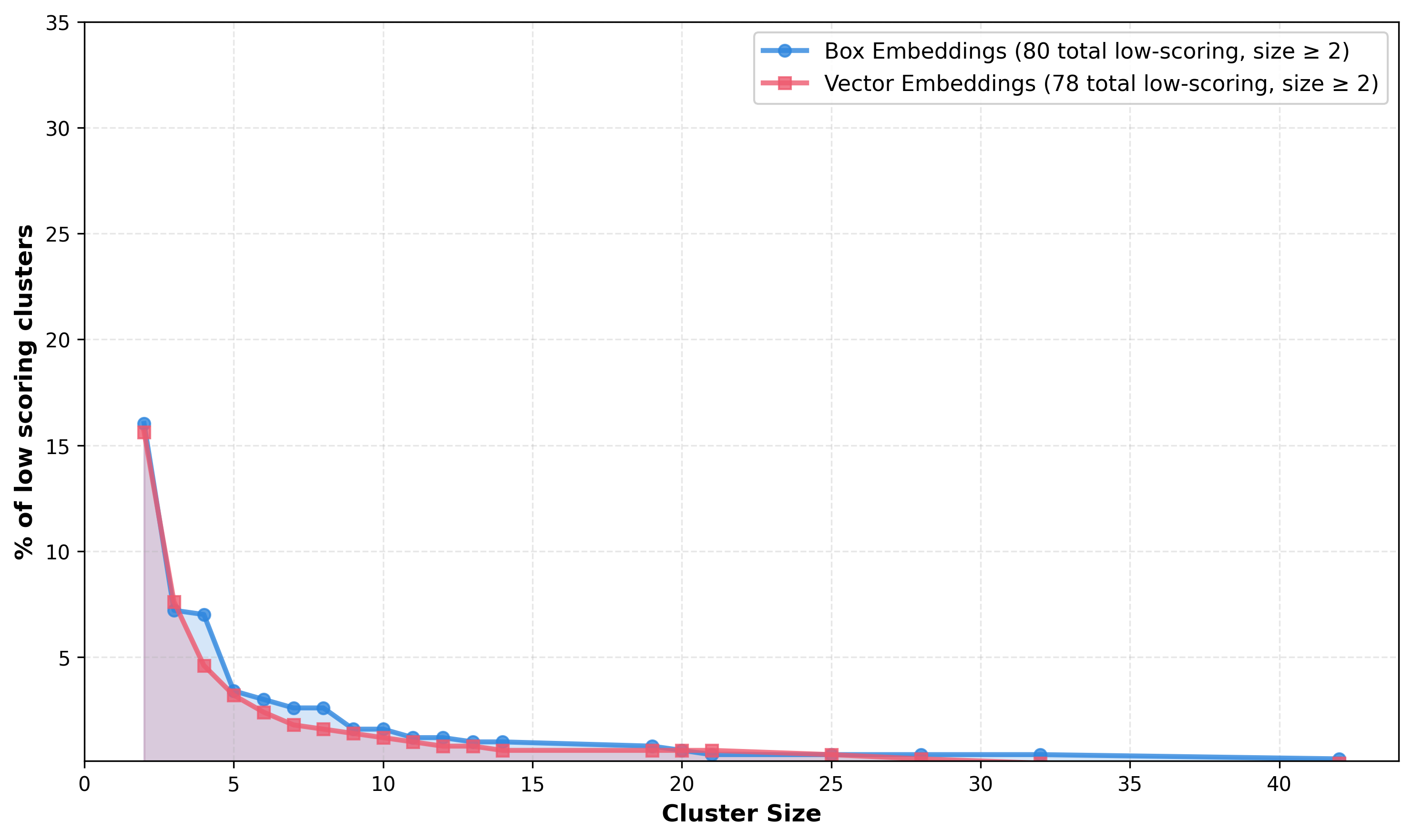} &
        \includegraphics[width=0.32\linewidth]{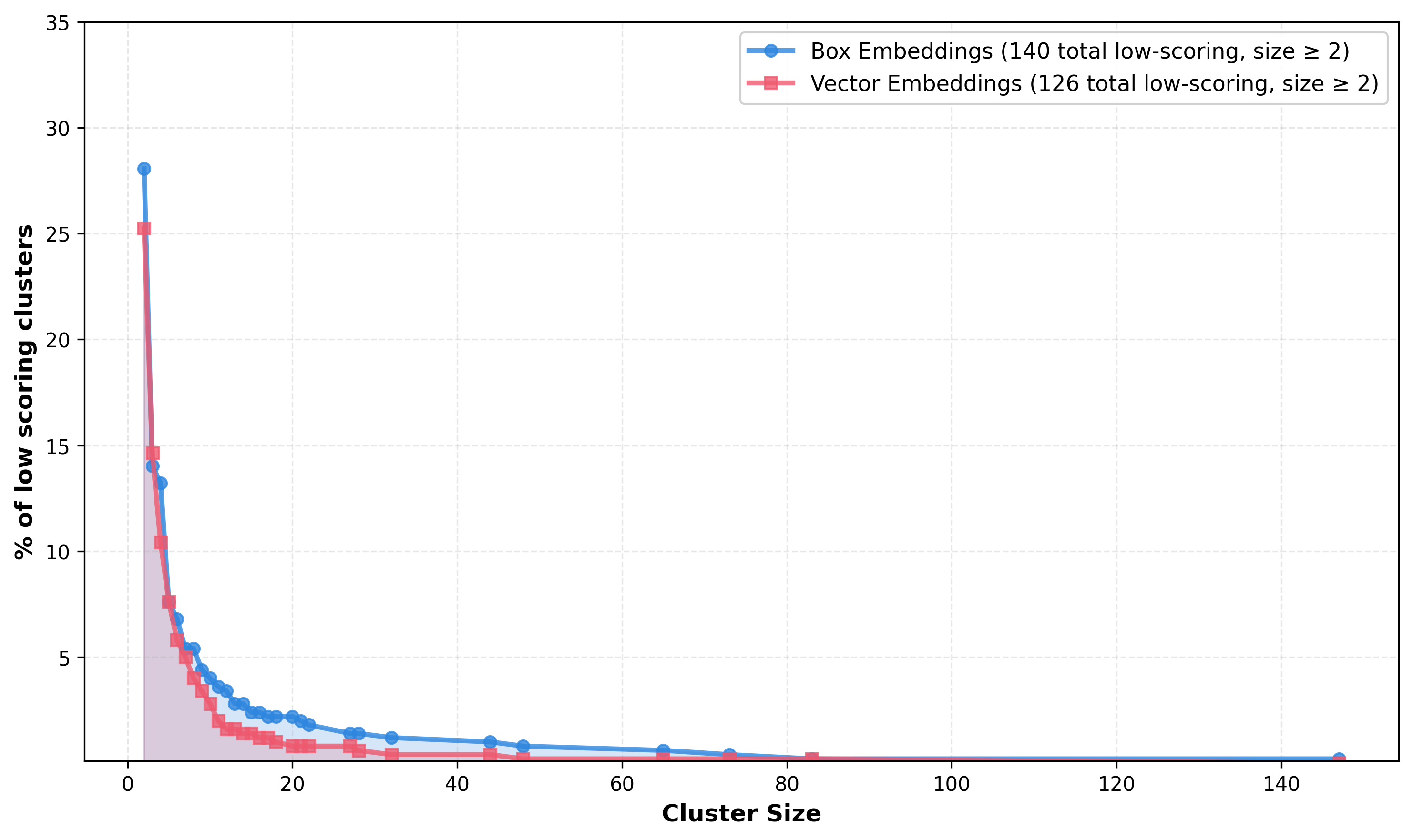} &
        \includegraphics[width=0.32\linewidth]{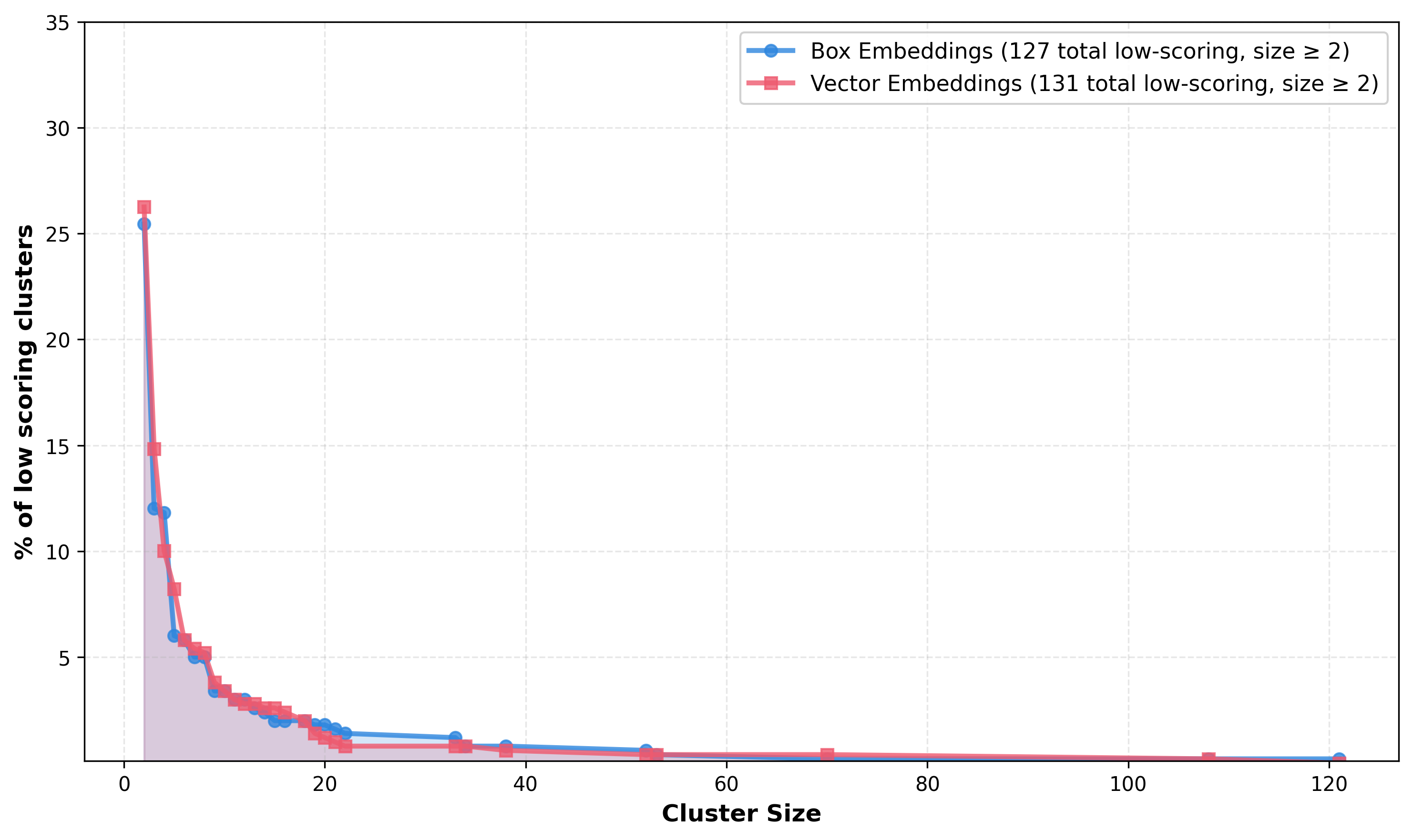} \\
        Vicuna-33B & UltraLM-13B & UltraLM-65B
    \end{tabular}

    \caption{Cumulative cluster score curves for LLaMA-family models and derivatives.}
    \label{fig:appendix_cluster_scores_group2}
\end{figure*}

\begin{figure*}[t]
    \centering
    \setlength{\tabcolsep}{4pt}
    \renewcommand{\arraystretch}{1}

    \begin{tabular}{ccc}
        \includegraphics[width=0.32\linewidth]{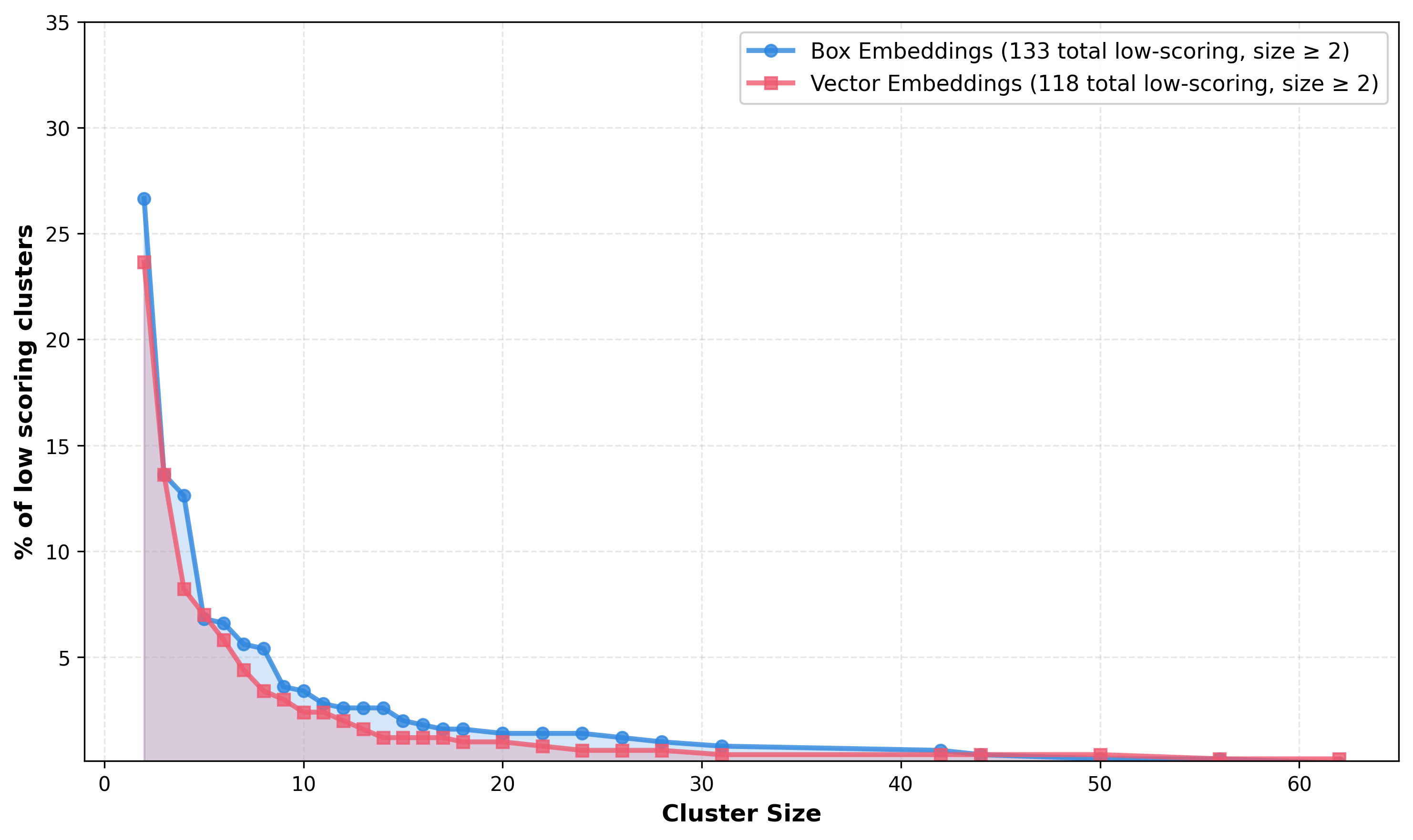} &
        \includegraphics[width=0.32\linewidth]{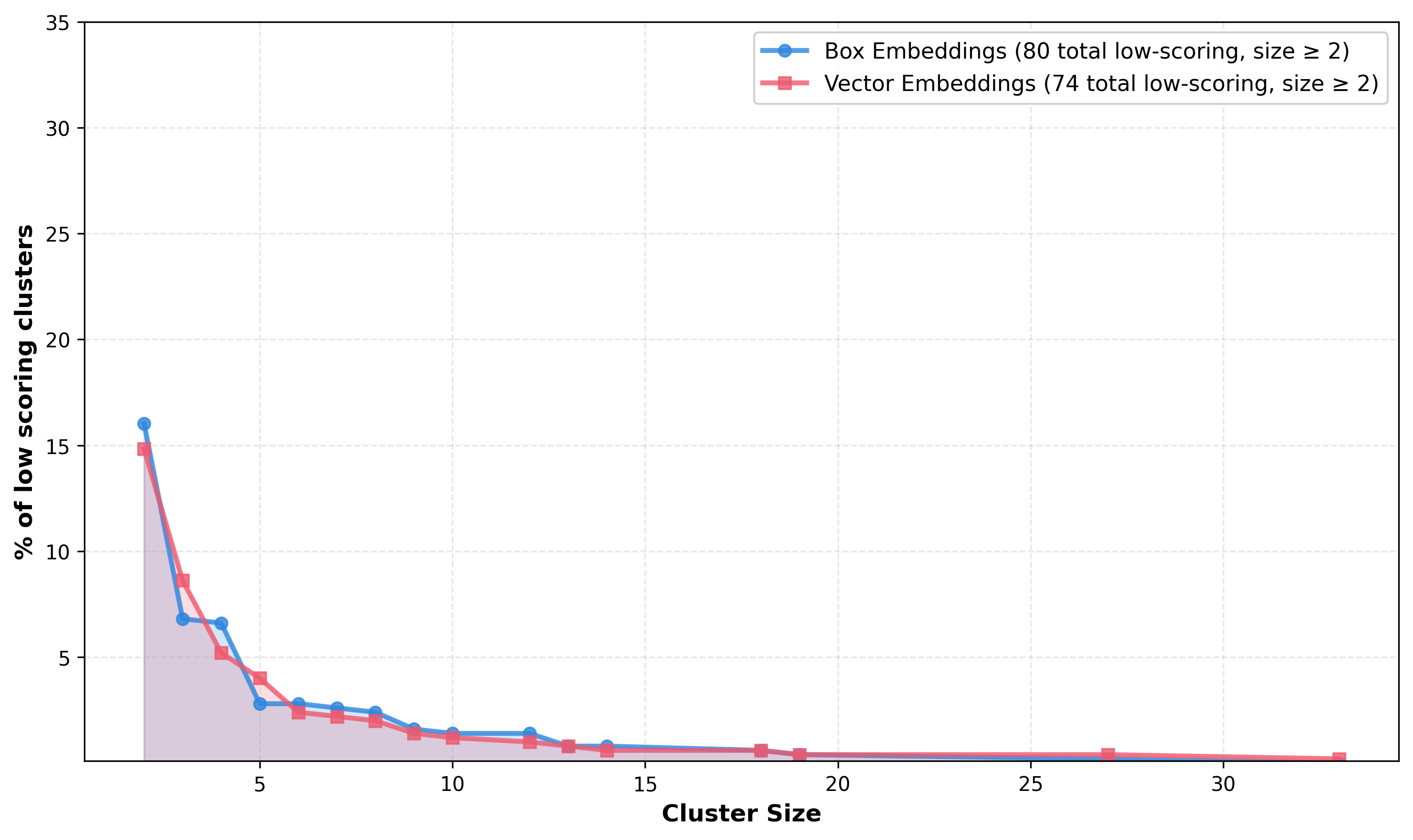} &
        \includegraphics[width=0.32\linewidth]{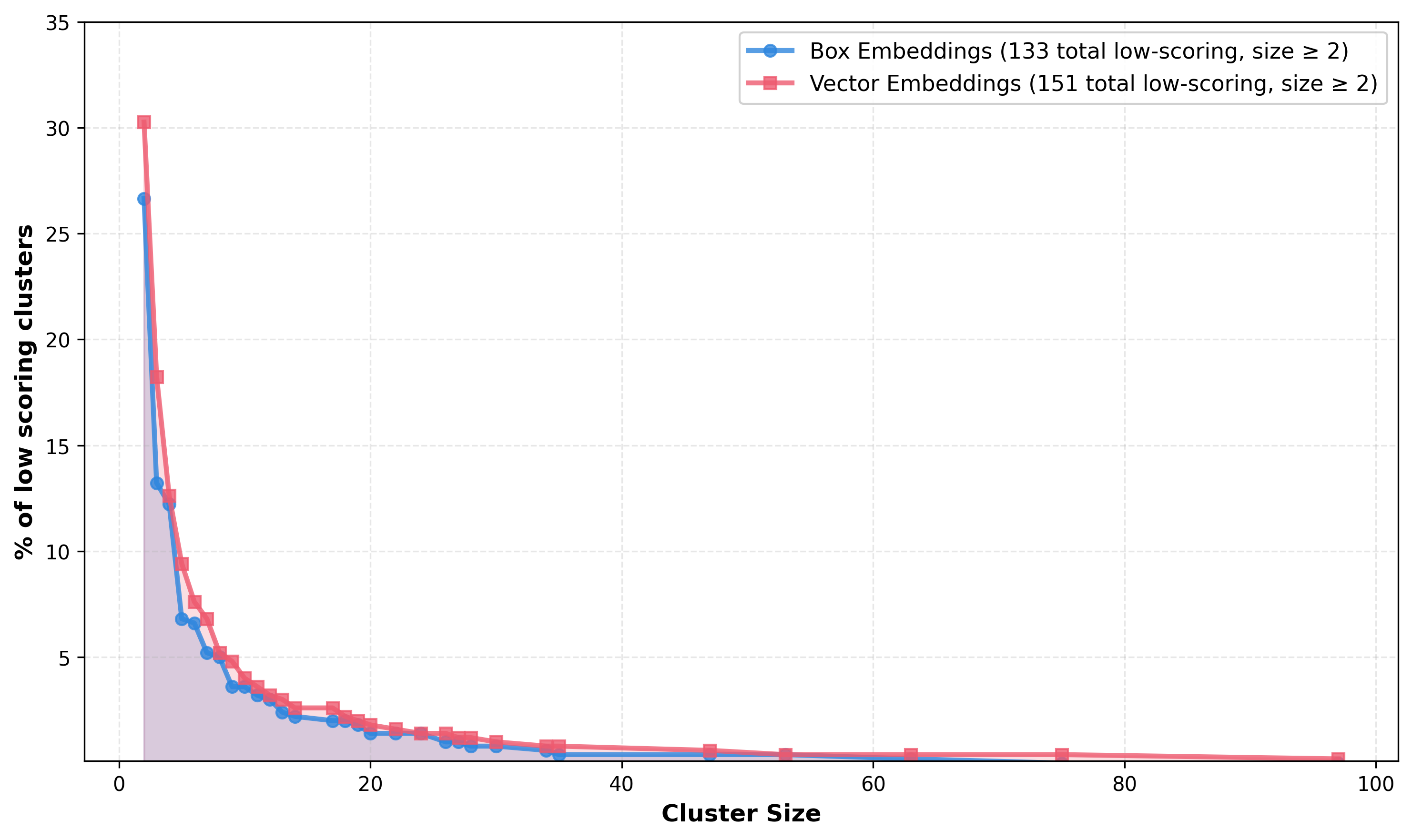} \\
        WizardLM-7B & WizardLM-13B & WizardLM-70B \\[6pt]

        \includegraphics[width=0.32\linewidth]{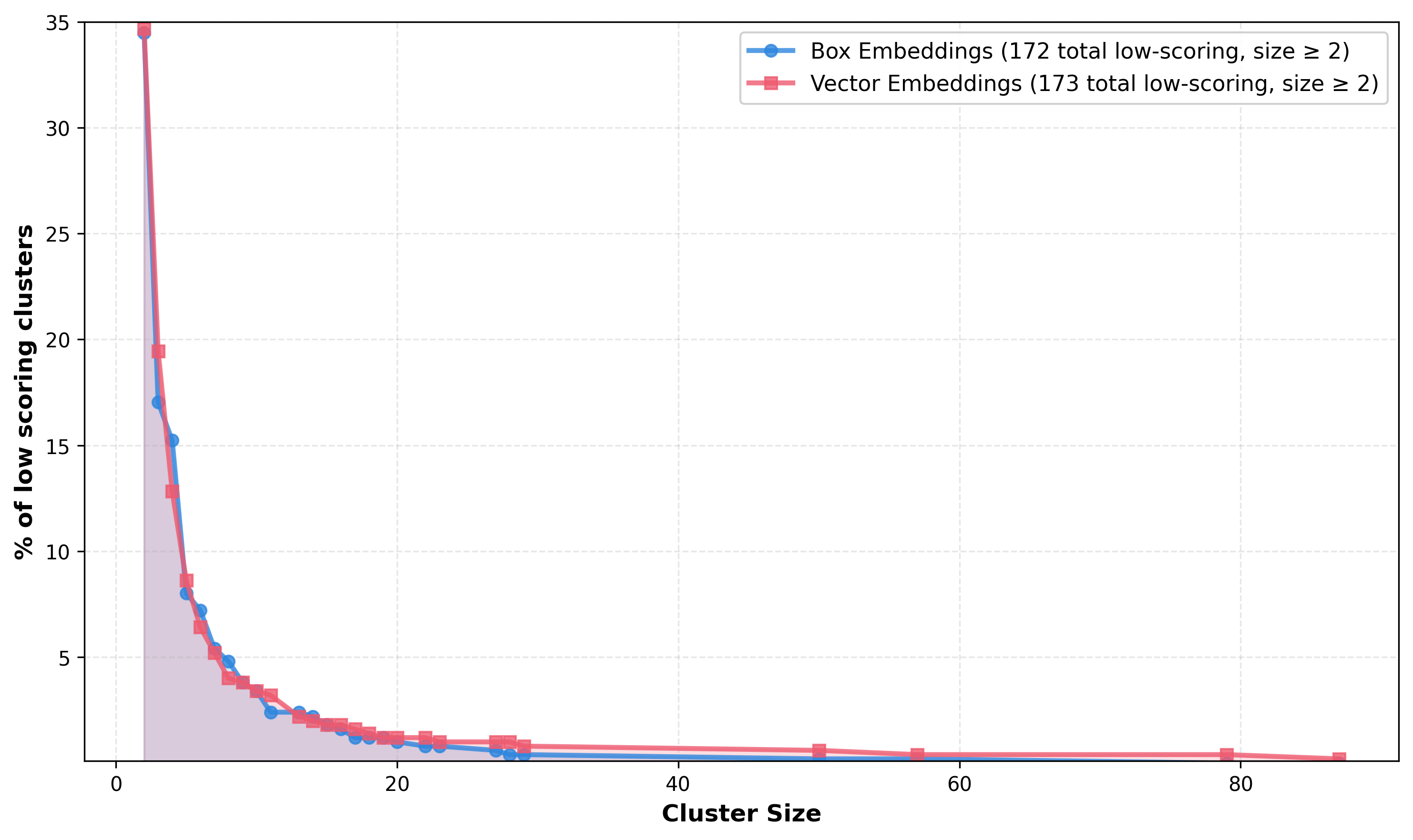} &
        \multicolumn{2}{c}{
            \includegraphics[width=0.32\linewidth]{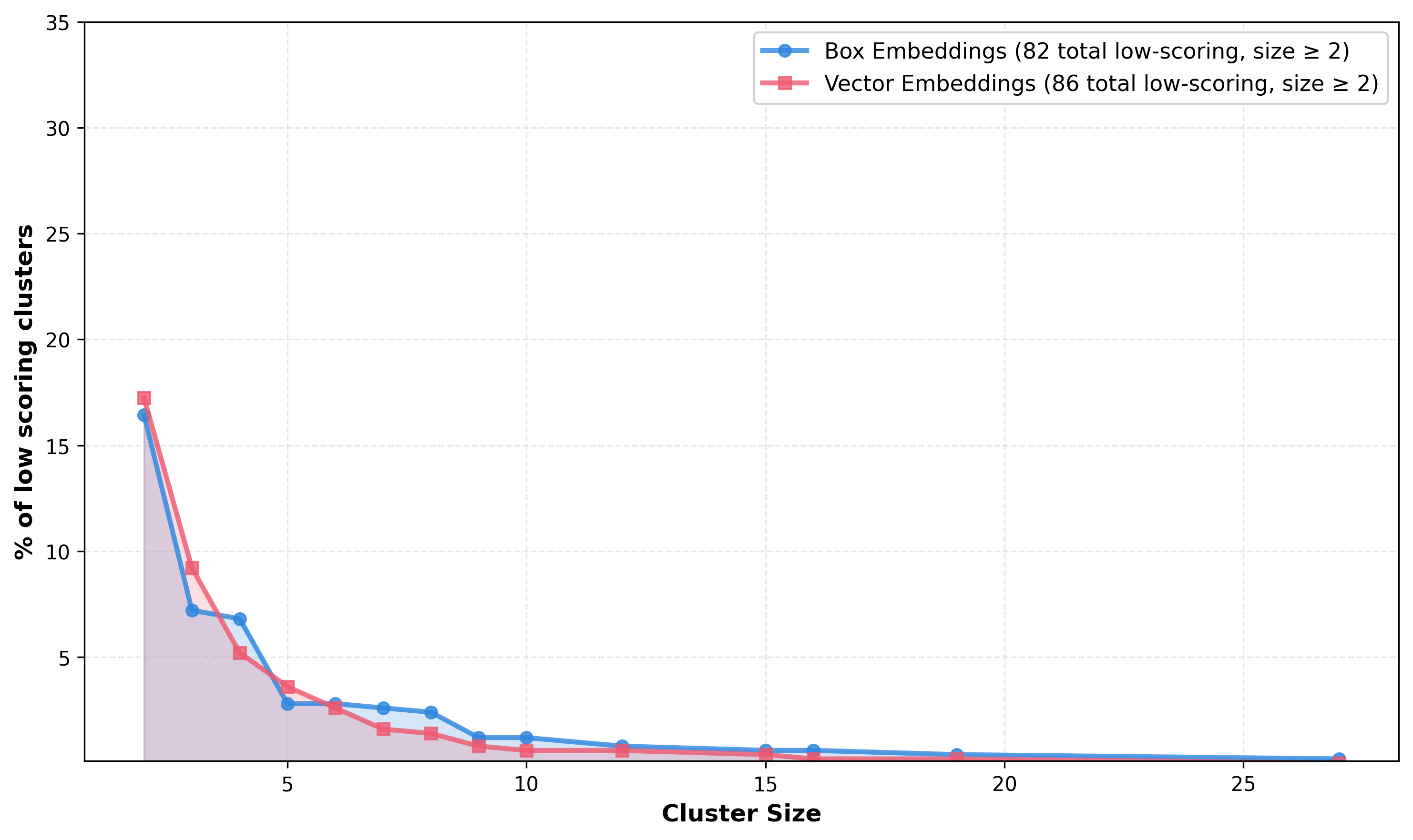}
        } \\
        Pythia-12B & \multicolumn{2}{c}{MPT-30B-Chat}
    \end{tabular}

    \caption{Cumulative cluster score curves for instruction-tuned open models.}
    \label{fig:appendix_cluster_scores_group3}
\end{figure*}

In \Cref{sec:weakness_analysis}, we define how we detect weakness by varying the cluster size. The figures in this section \Cref{fig:appendix_cluster_scores_group1}, \Cref{fig:appendix_cluster_scores_group2}, and \Cref{fig:appendix_cluster_scores_group3} covers 17 LLMs of varying types and sizes. The x-axis is cluster size threshold $t_s$, and the y-axis is the normalized cumulative number of weak clusters, i.e., $\#W_{t_s}^{R_{25\%}} / 499$ in percentage.

% -------- Appendix Figure A.1 (6 plots): Closed / API-like + strong baselines --------

% -------- Appendix Figure A.2 (6 plots): LLaMA-family and derivatives --------

% -------- Appendix Figure A.3 (5 plots): Instruction-tuned open models --------

\begin{table}[h]
\centering
\caption{Relative improvement in all size weakness cluster count over Vector baseline across Box variants.}
\label{tab:containment-improvement}
\begin{tabular}{lccc}
\toprule
\textbf{Model} & \textbf{Box w/o synth Improv.} & \textbf{Box Improv.} & \textbf{Box w/o link Improv.} \\
\midrule
alpaca-7b           & +11.28\% & +11.28\% & +12.03\% \\
bard                & -14.29\% &  -1.10\% &  -9.89\% \\
falcon-40b-instruct & +15.94\% &  +7.25\% & +10.14\% \\
gpt-3.5-turbo       & +10.71\% & +42.86\% &  +3.57\% \\
gpt-4               & +14.29\% & +11.43\% &  +2.86\% \\
llama-2-13b-chat    &  +3.26\% &  +6.52\% &  +4.35\% \\
llama-2-70b-chat    & +33.33\% & +26.67\% & +21.33\% \\
llama-2-7b-chat     & +24.64\% & +13.04\% & +11.59\% \\
mpt-30b-chat        &  +4.65\% &  +3.49\% &  -1.16\% \\
pythia-12b          &  -5.78\% &  -4.05\% &  -1.16\% \\
starchat            & -17.72\% &  -6.33\% & -18.99\% \\
ultralm-13b         & +11.11\% &  +9.52\% &  +8.73\% \\
ultralm-65b         &  -0.76\% &  +1.53\% &  -3.82\% \\
vicuna-33b          &  +3.85\% &  -2.56\% &  +3.85\% \\
wizardlm-13b        &  +4.05\% & +12.16\% & +12.16\% \\
wizardlm-70b        &  -8.61\% &  -9.27\% &  -7.95\% \\
wizardlm-7b         & +11.02\% &  +8.47\% &  +0.85\% \\
\midrule
\textbf{Average}    & \textbf{+5.94\%} & \textbf{+7.70\%} & \textbf{+2.85\%} \\
\bottomrule
\end{tabular}
\end{table}

\begin{table}[h]
\centering
\caption{Relative AUC improvement over Vector baseline across Box variants.}
\label{tab:auc-improvement}
\begin{tabular}{lccc}
\toprule
\textbf{Model} & \textbf{Box w/o links Improv.} & \textbf{Box Improv.} & \textbf{Box w/o synth Improv.} \\
\midrule
alpaca-7b           & +59.20\% & +55.06\% & +44.28\% \\
bard                & -21.15\% &  +6.61\% & -29.36\% \\
falcon-40b-instruct &  +2.74\% &  +4.83\% & +17.81\% \\
gpt-3.5-turbo       & -29.11\% & +21.52\% & -25.97\% \\
gpt-4               &  -0.00\% & +50.79\% & +57.14\% \\
llama-2-13b-chat    & +16.49\% &  +8.00\% &  +9.75\% \\
llama-2-70b-chat    & +61.90\% & +63.44\% & +70.53\% \\
llama-2-7b-chat     &  +5.50\% &  -9.46\% & +21.10\% \\
mpt-30b-chat        &  +9.68\% & +20.83\% & +36.53\% \\
pythia-12b          &  -1.17\% &  -7.12\% & -11.15\% \\
starchat            & -31.20\% & -29.66\% & -38.95\% \\
ultralm-13b         & +25.43\% & +17.06\% & +23.85\% \\
ultralm-65b         &  -9.80\% & +11.71\% &  -3.44\% \\
vicuna-33b          & +25.56\% &  +3.65\% &  -2.28\% \\
wizardlm-13b        &  +4.80\% &  +5.73\% &  -2.64\% \\
wizardlm-70b        &  +5.11\% & -16.52\% &  -4.36\% \\
wizardlm-7b         &  +8.71\% & +22.54\% & +23.28\% \\
\midrule
\textbf{Average}    & \textbf{+7.81\%} & \textbf{+13.47\%} & \textbf{+10.95\%} \\
\bottomrule
\end{tabular}
\end{table}

\newcommand{\imgpath}{figs/pictures/image_best_fit_score_only}
\newcommand{\mw}{0.22}  % subfigure width — adjust if needed
 
%% ============================================================
\subsection{Specificity Experiments}
%% ============================================================
The RMSE comparison between box volume and length could be seen at \Cref{tab:rmse-diff} and \Cref{tab:rmse-diff2}. For Ultrafeedback we do not include pythia-12b, as it had less that 5000 usable samples, leading to a lot of noise in the KDE estimation.

\section{AI Usage}
We use Claude or other LLMs to help us revise the paper and codes. The outputs from LLMs are checked manually.
 
\begin{figure}[htbp]
    \centering
 
    % Row 1
    \begin{subfigure}[t]{\mw\textwidth}
        \includegraphics[width=\linewidth]{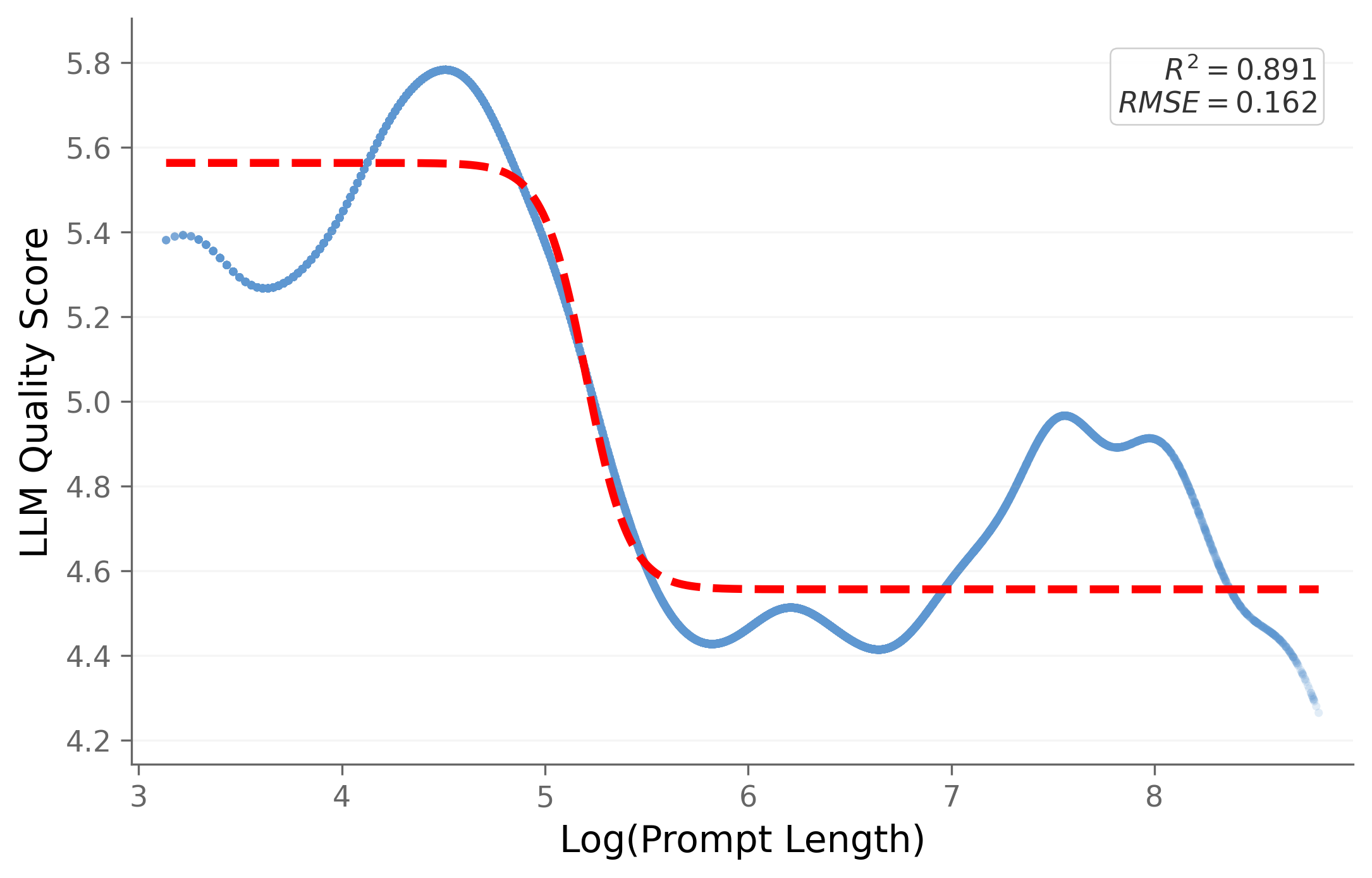}
        \caption{Alpaca-7B}
    \end{subfigure}\hfill
    \begin{subfigure}[t]{\mw\textwidth}
        \includegraphics[width=\linewidth]{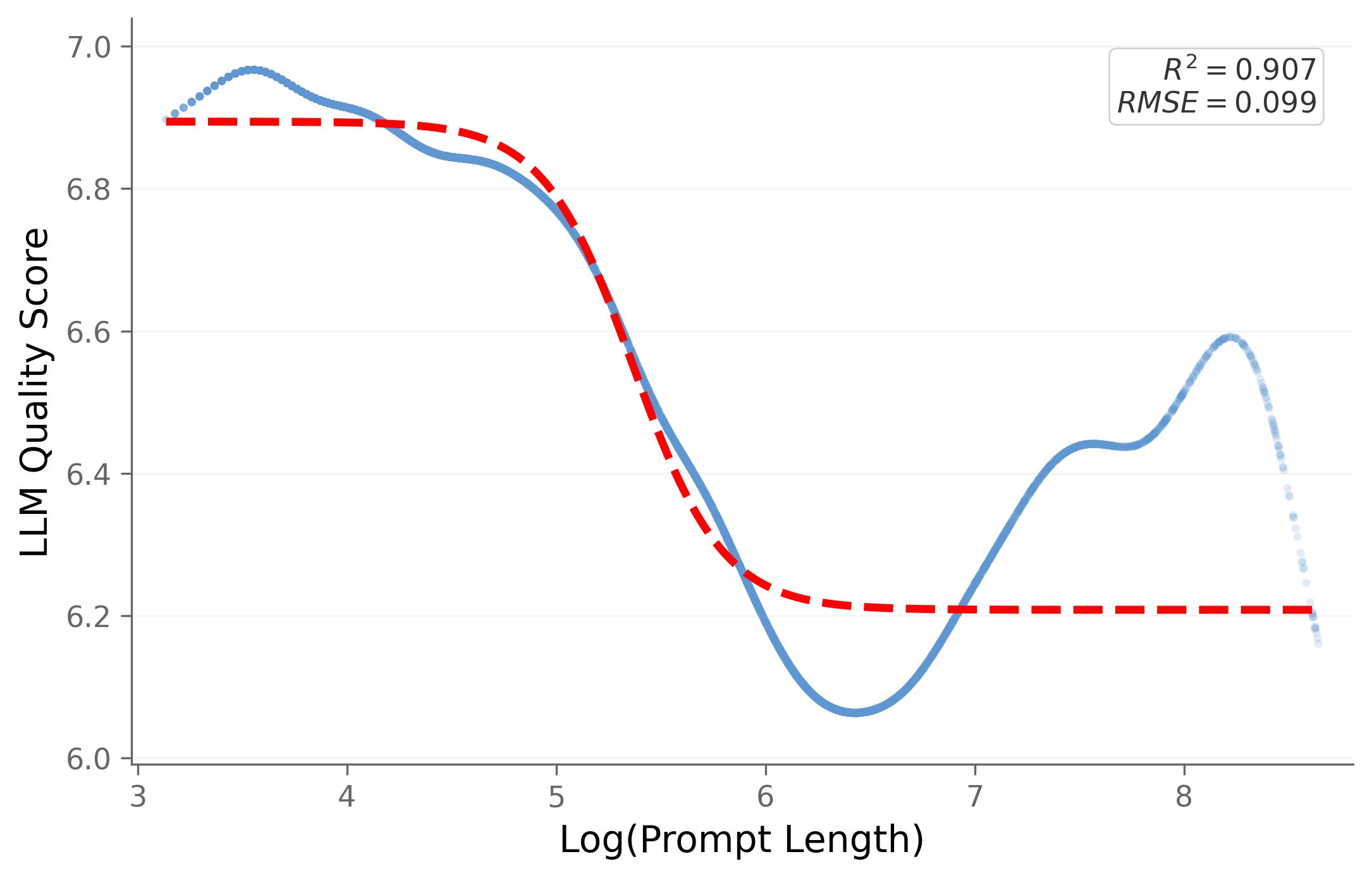}
        \caption{Bard}
    \end{subfigure}\hfill
    \begin{subfigure}[t]{\mw\textwidth}
        \includegraphics[width=\linewidth]{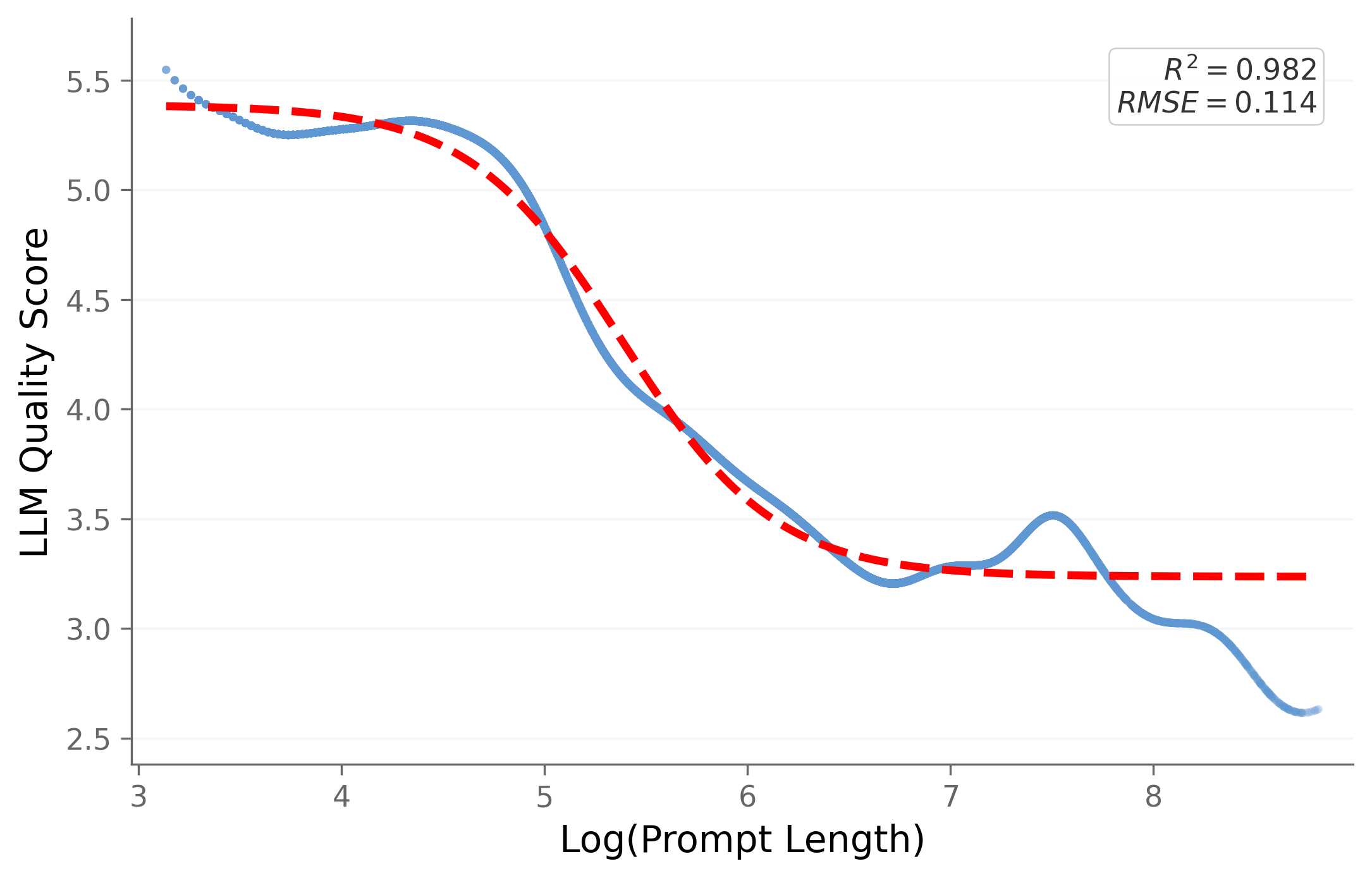}
        \caption{Falcon-40B-Instruct}
    \end{subfigure}\hfill
    \begin{subfigure}[t]{\mw\textwidth}
        \includegraphics[width=\linewidth]{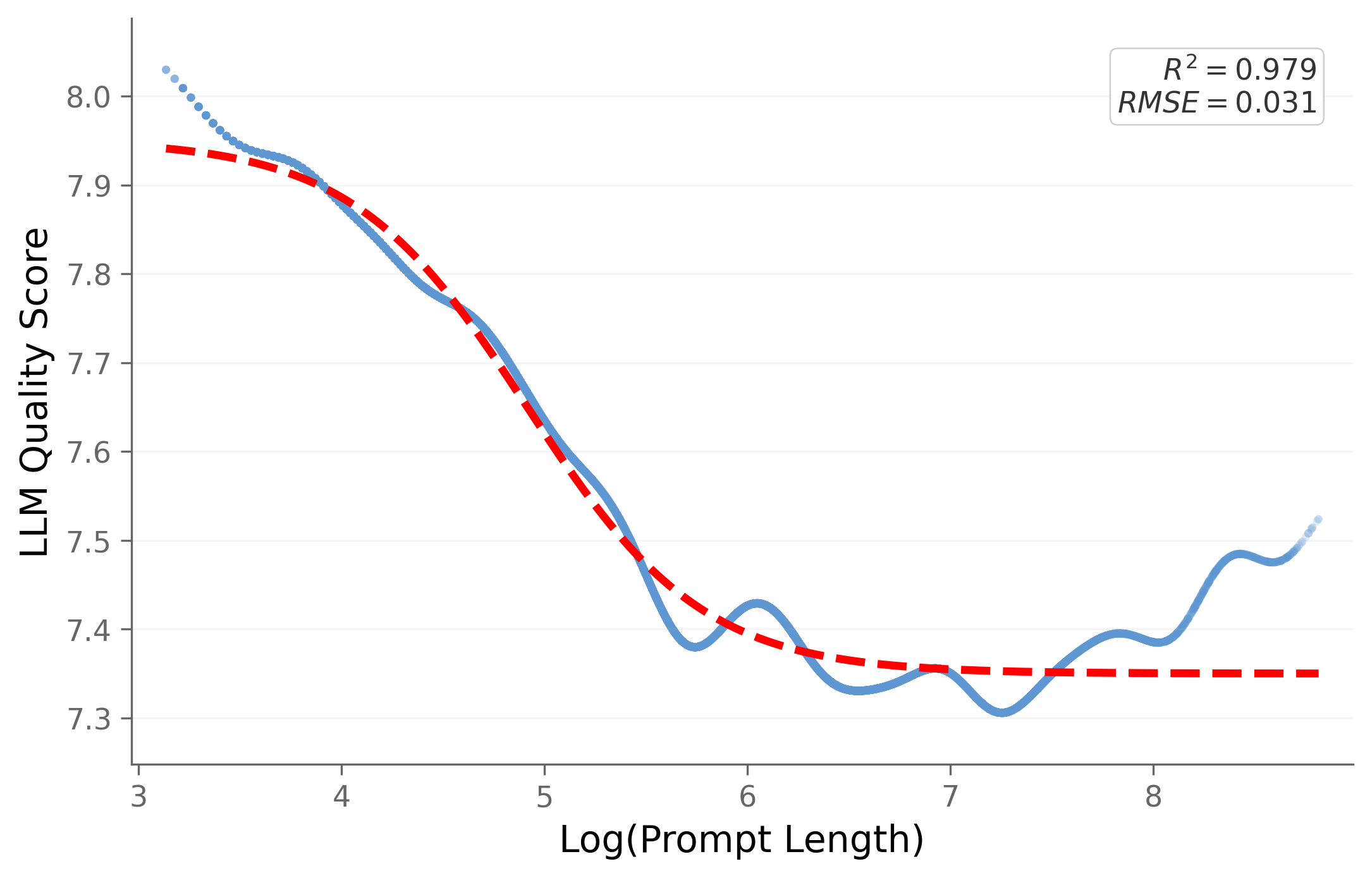}
        \caption{GPT-3.5-Turbo}
    \end{subfigure}
 
    \vspace{0.5em}
 
    % Row 2
    \begin{subfigure}[t]{\mw\textwidth}
        \includegraphics[width=\linewidth]{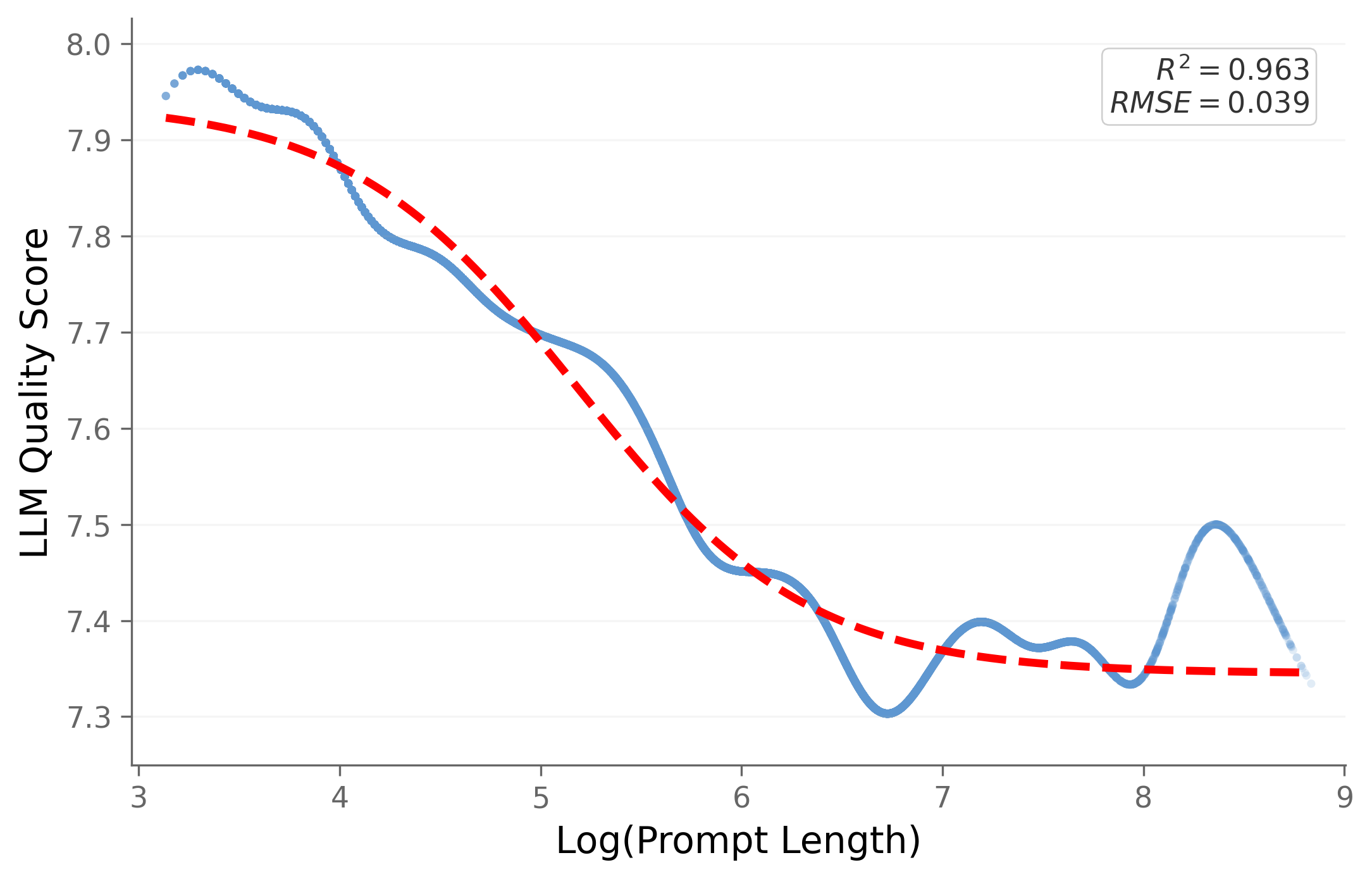}
        \caption{GPT-4}
    \end{subfigure}\hfill
    \begin{subfigure}[t]{\mw\textwidth}
        \includegraphics[width=\linewidth]{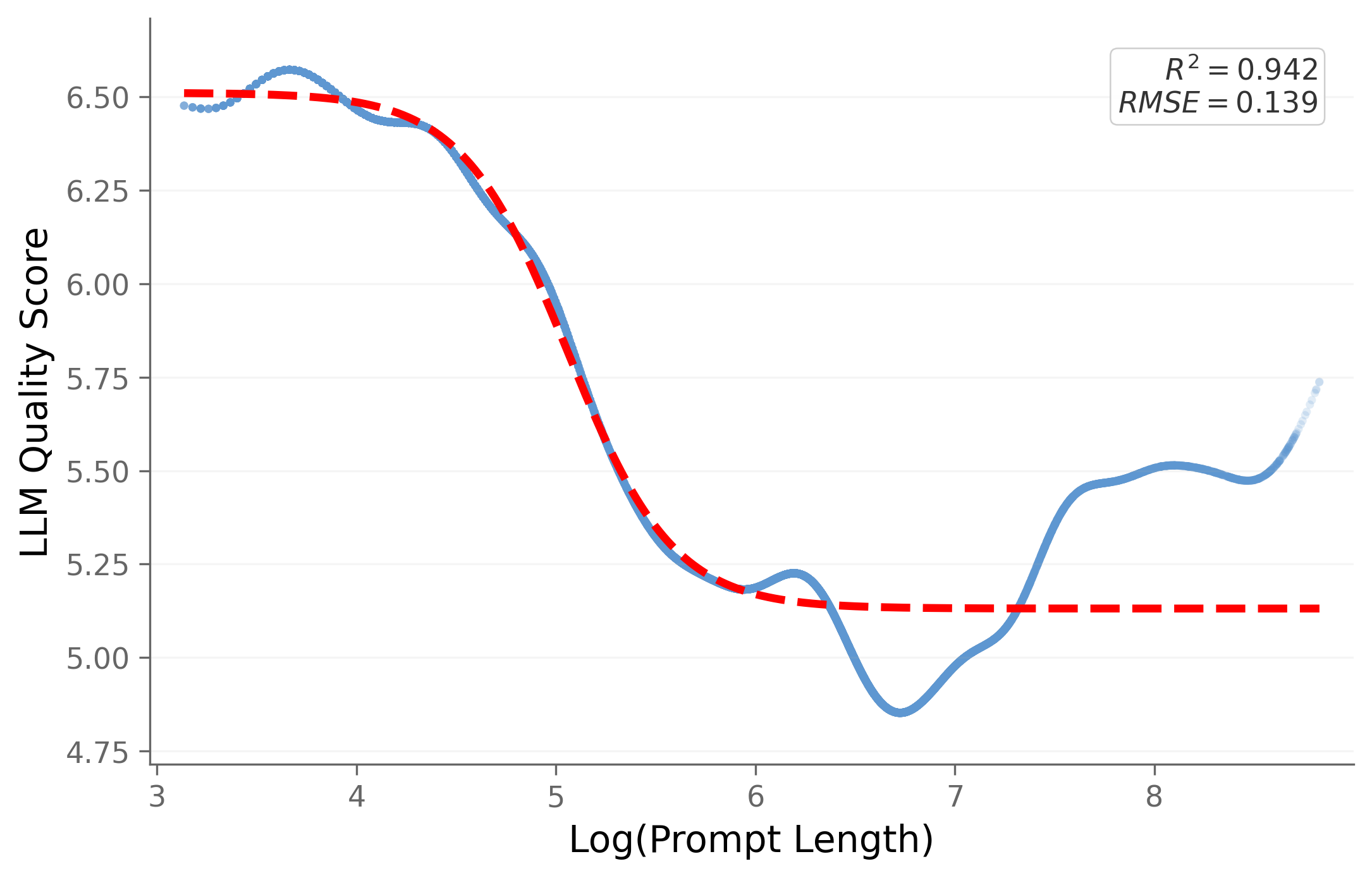}
        \caption{LLaMA-2-7B-Chat}
    \end{subfigure}\hfill
    \begin{subfigure}[t]{\mw\textwidth}
        \includegraphics[width=\linewidth]{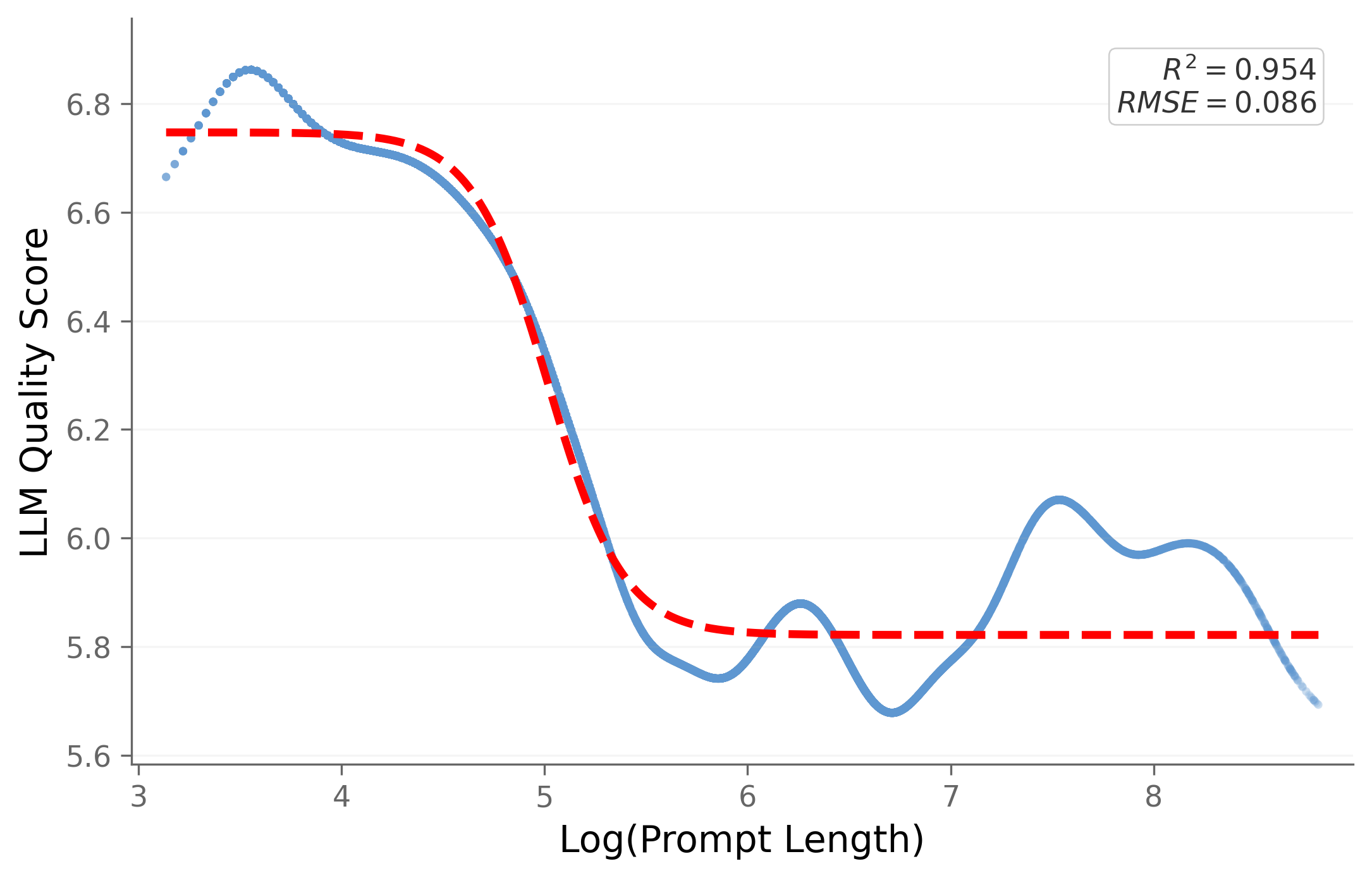}
        \caption{LLaMA-2-13B-Chat}
    \end{subfigure}\hfill
    \begin{subfigure}[t]{\mw\textwidth}
        \includegraphics[width=\linewidth]{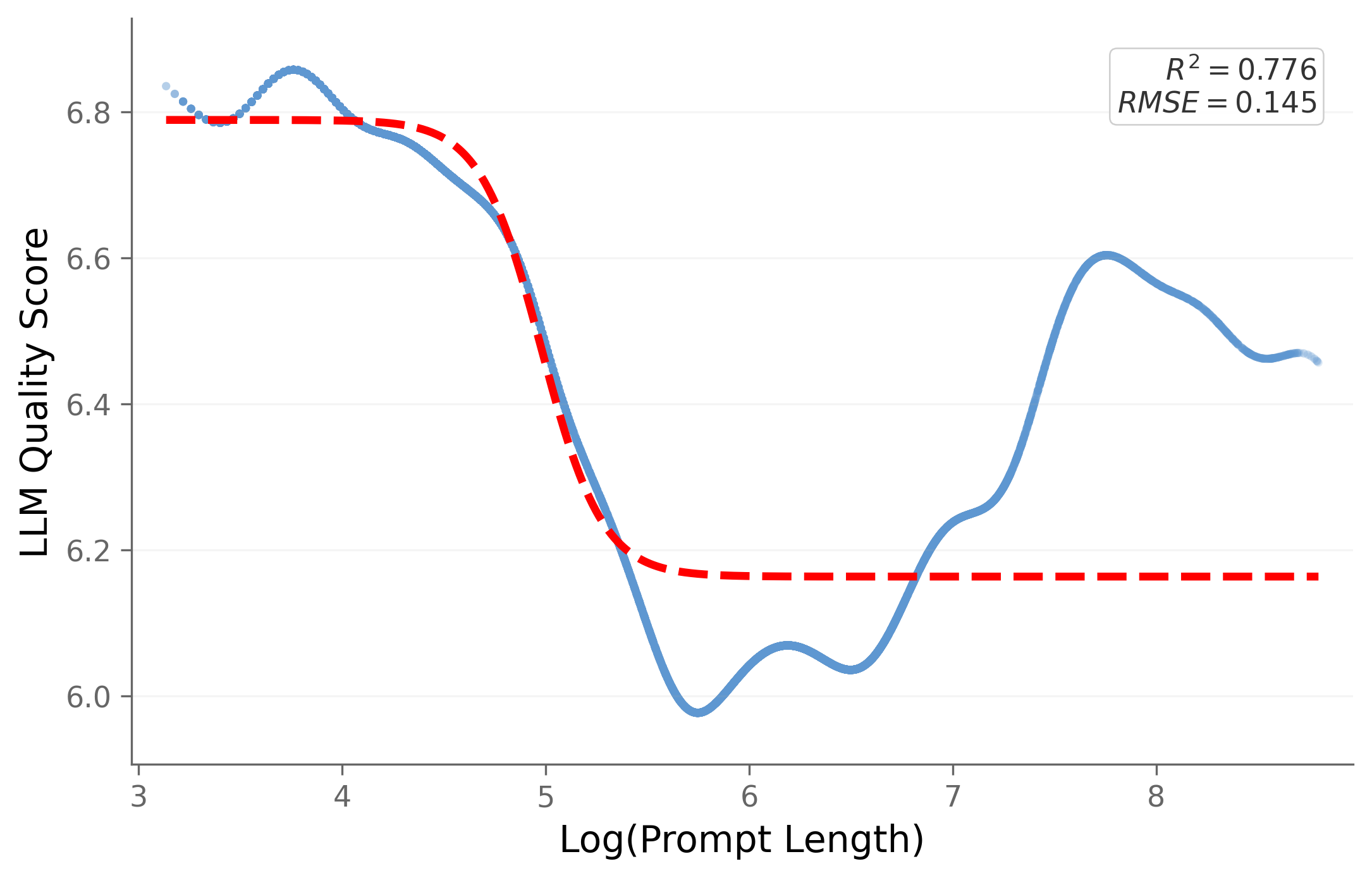}
        \caption{LLaMA-2-70B-Chat}
    \end{subfigure}
 
    \vspace{0.5em}
 
    % Row 3
    \begin{subfigure}[t]{\mw\textwidth}
        \includegraphics[width=\linewidth]{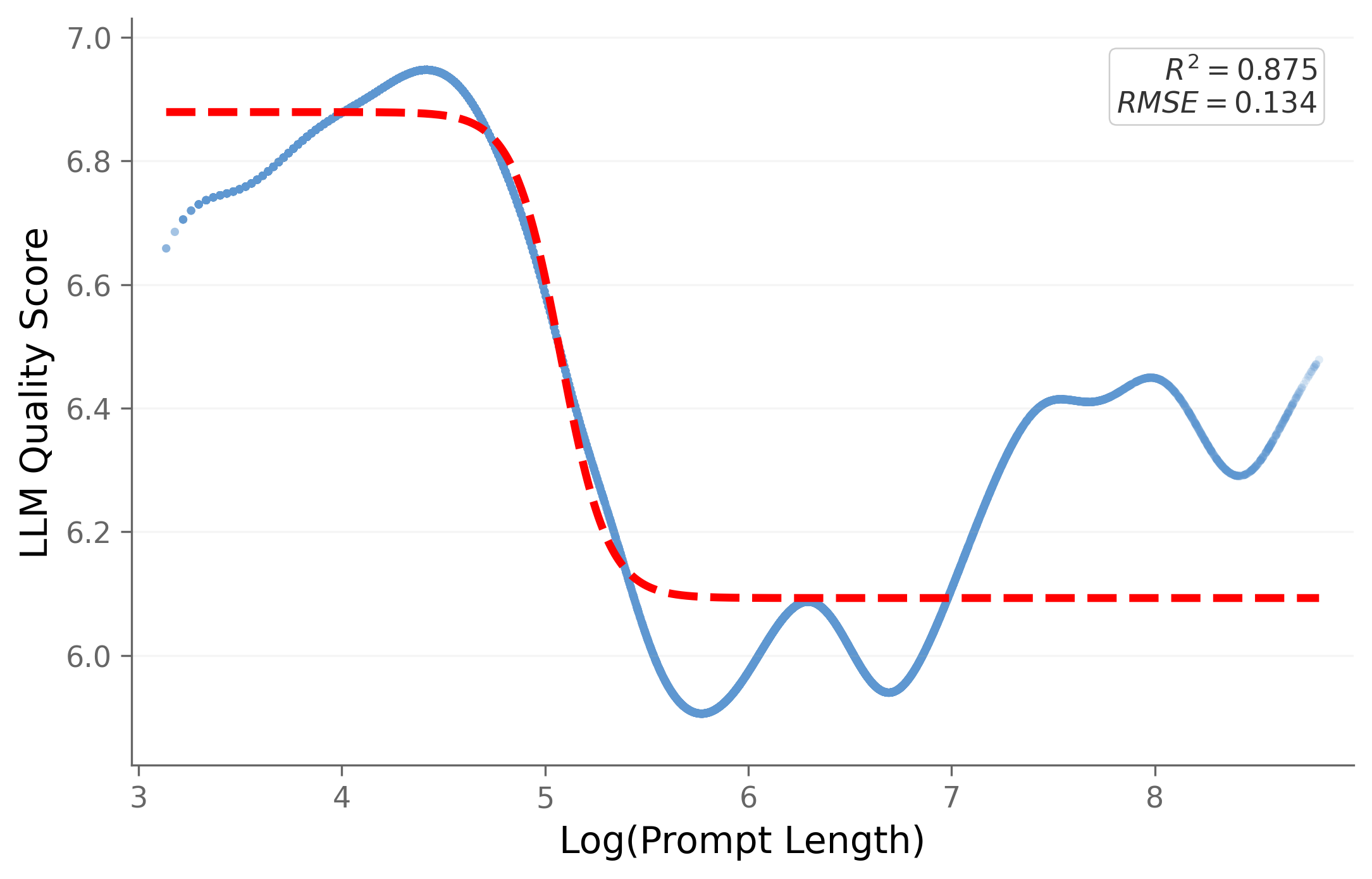}
        \caption{MPT-30B-Chat}
    \end{subfigure}\hfill
    \begin{subfigure}[t]{\mw\textwidth}
        \includegraphics[width=\linewidth]{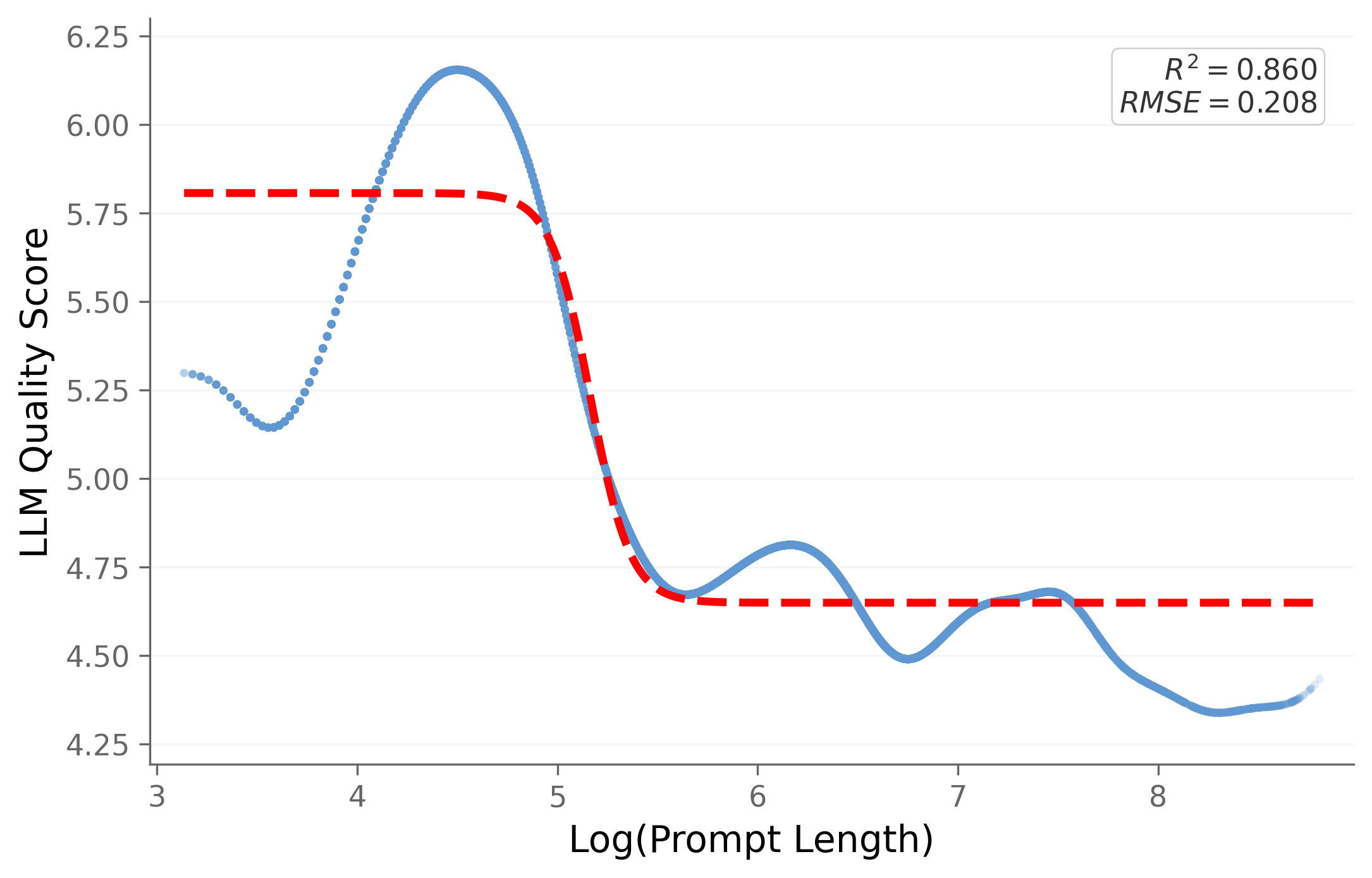}
        \caption{StarChat}
    \end{subfigure}\hfill
    \begin{subfigure}[t]{\mw\textwidth}
        \includegraphics[width=\linewidth]{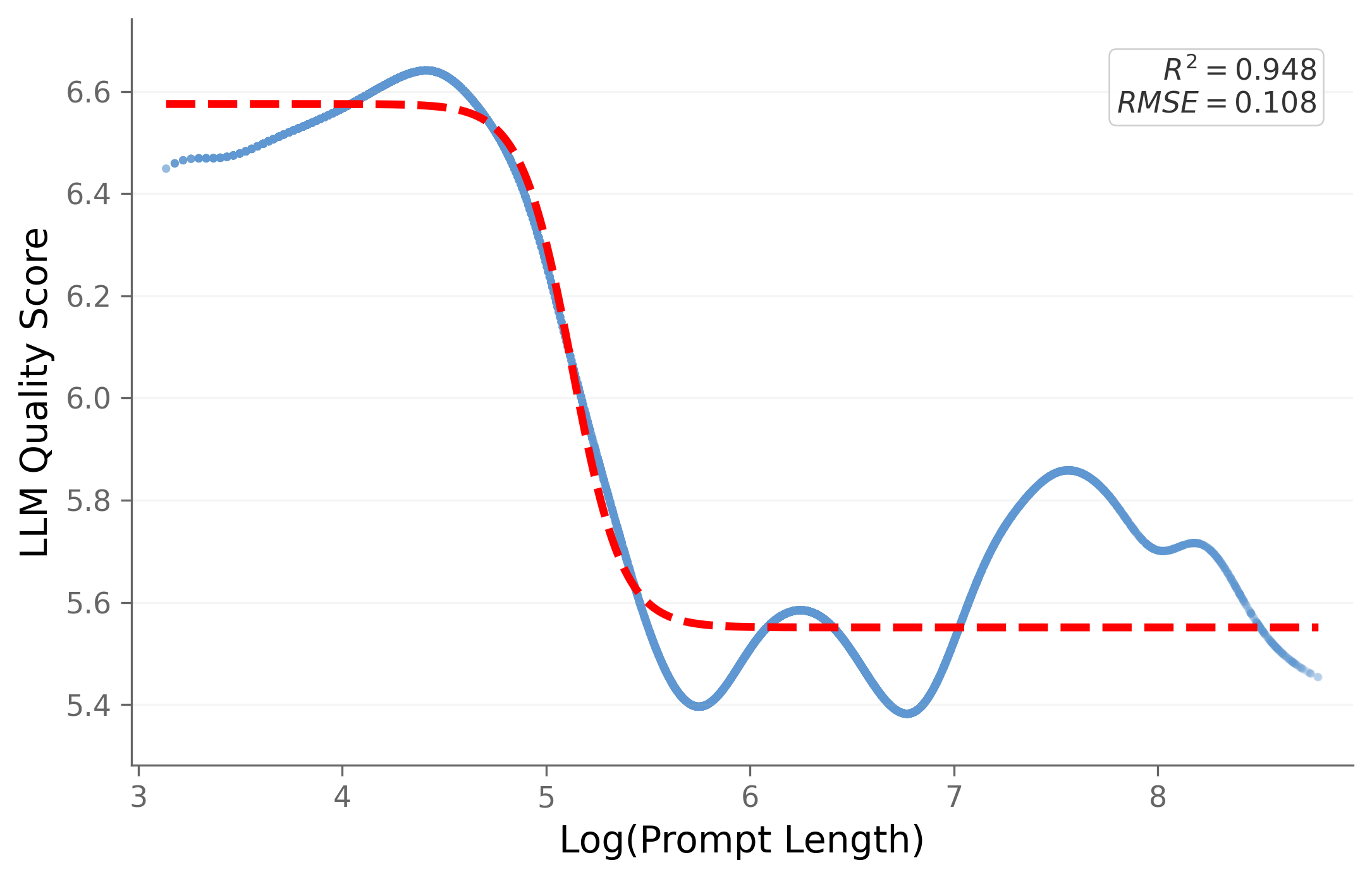}
        \caption{UltraLM-13B}
    \end{subfigure}
 
    \vspace{0.5em}
 
    % Row 4
    \begin{subfigure}[t]{\mw\textwidth}
        \includegraphics[width=\linewidth]{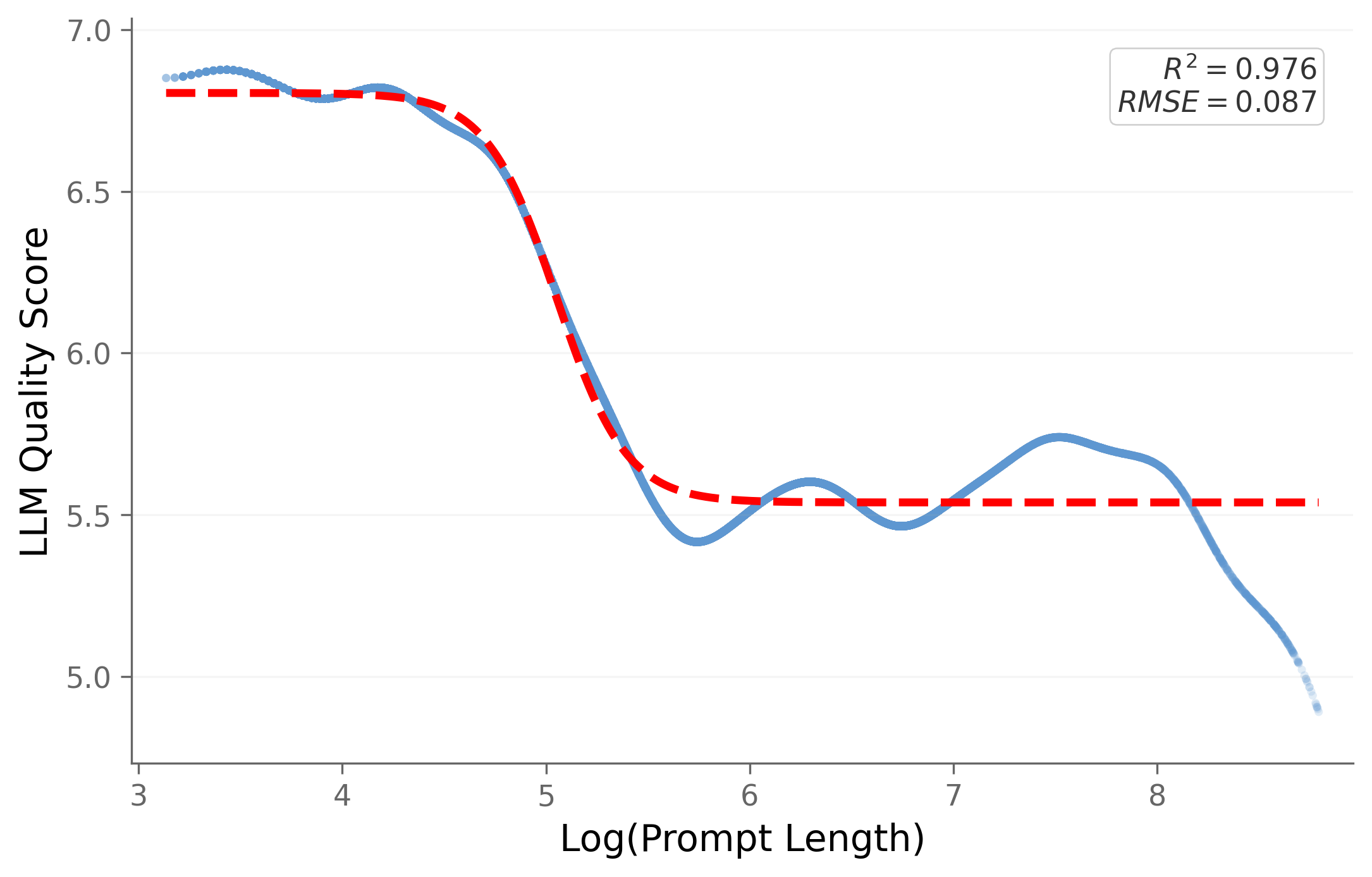}
        \caption{UltraLM-65B}
    \end{subfigure}\hfill
    \begin{subfigure}[t]{\mw\textwidth}
        \includegraphics[width=\linewidth]{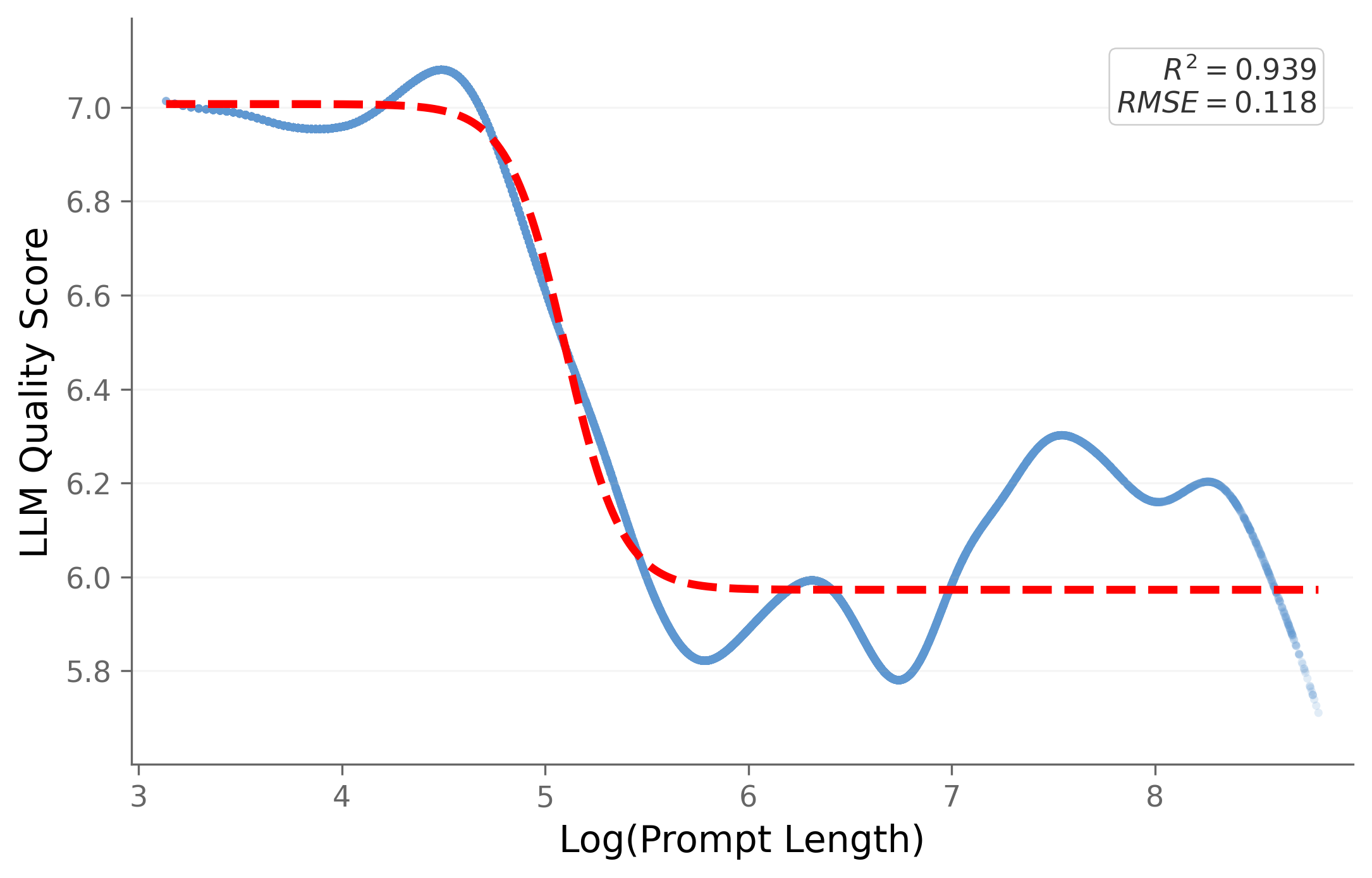}
        \caption{Vicuna-33B}
    \end{subfigure}\hfill
    \begin{subfigure}[t]{\mw\textwidth}
        \includegraphics[width=\linewidth]{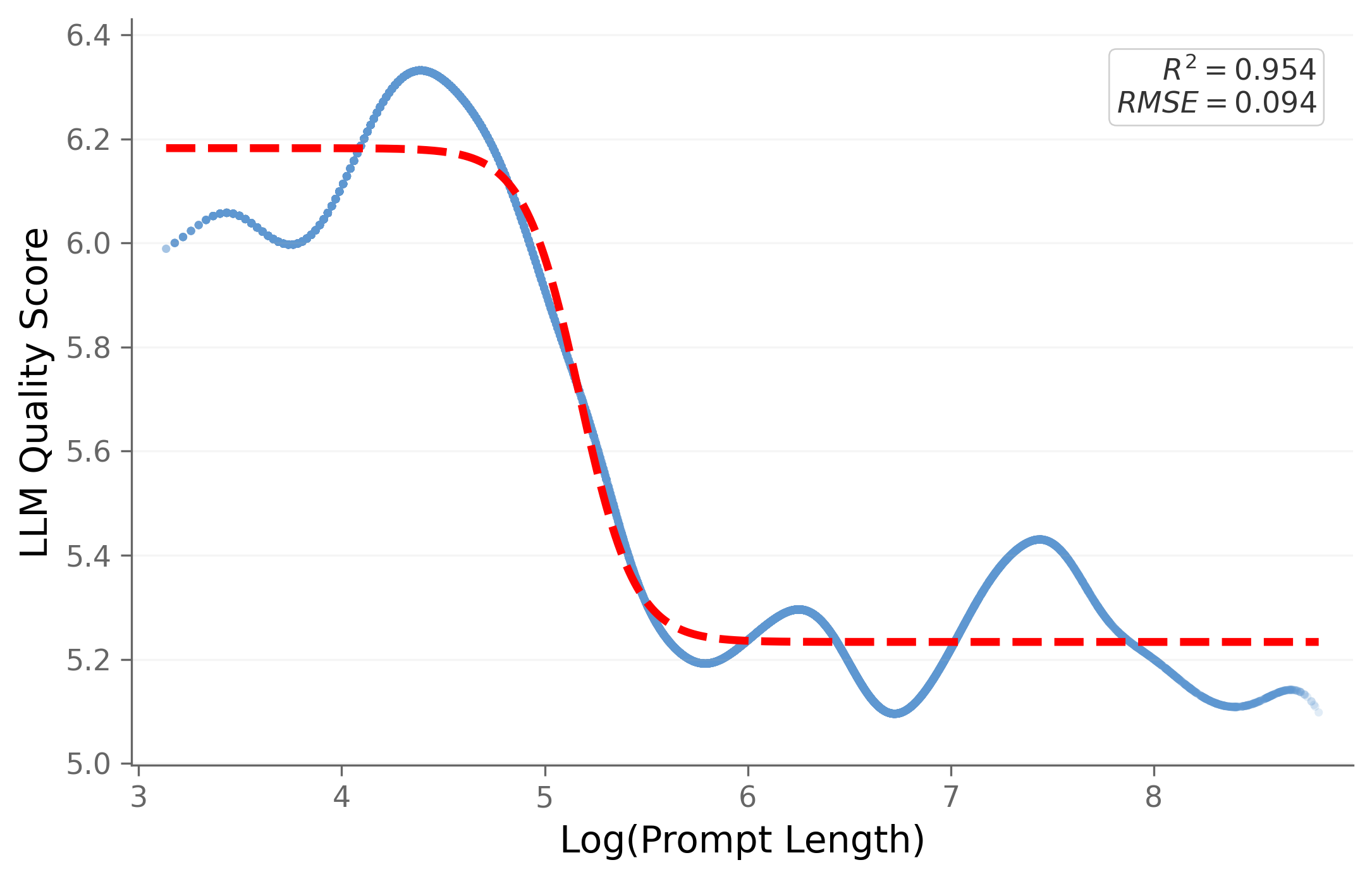}
        \caption{WizardLM-7B}
    \end{subfigure}\hfill
    \begin{subfigure}[t]{\mw\textwidth}
        \includegraphics[width=\linewidth]{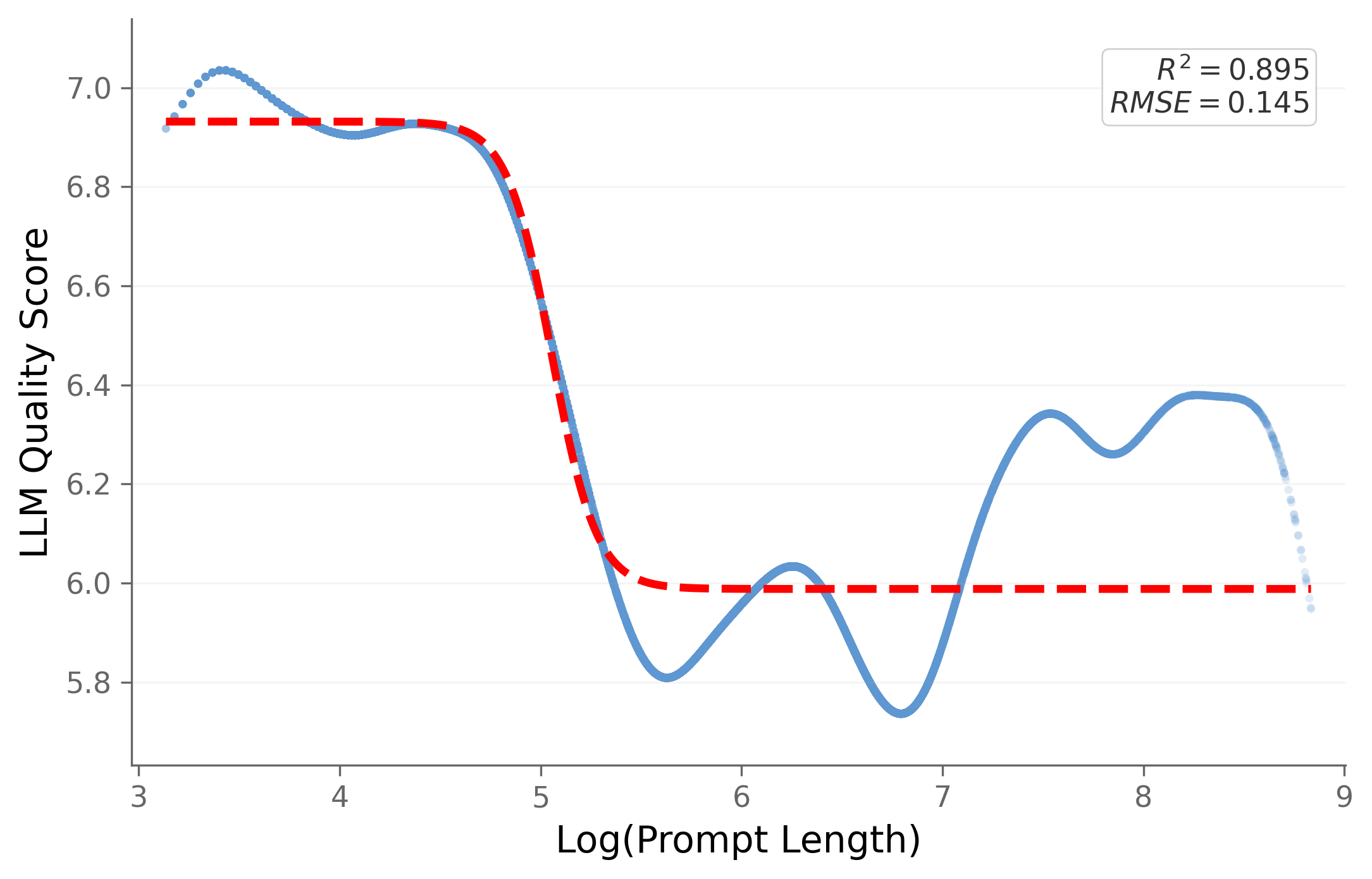}
        \caption{WizardLM-13B}
    \end{subfigure}
 
    \vspace{0.5em}
 
    % Row 5
    \begin{subfigure}[t]{\mw\textwidth}
        \includegraphics[width=\linewidth]{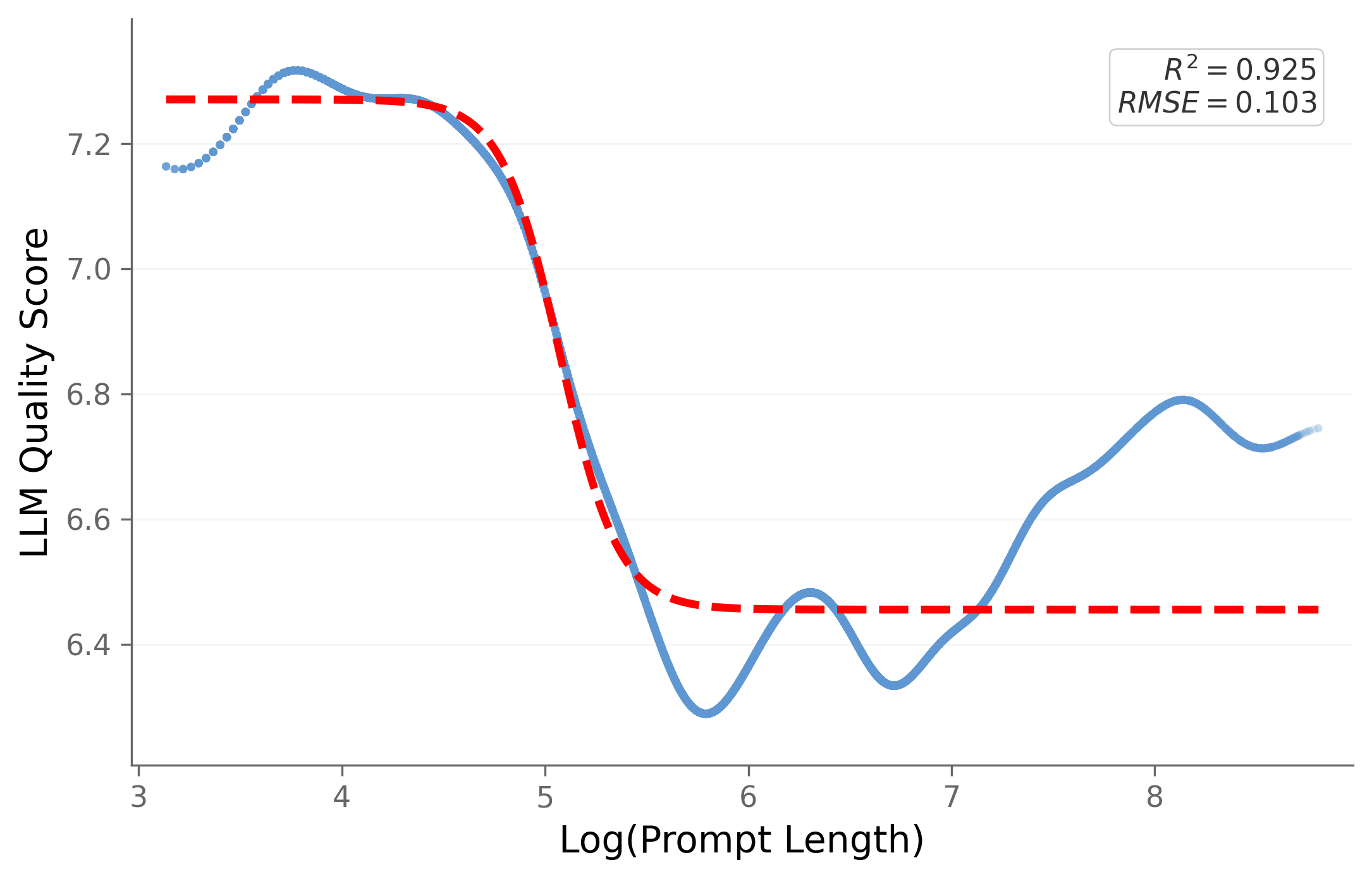}
        \caption{WizardLM-70B}
    \end{subfigure}
 
    \caption{Sigmoid best-fit curves for score (y-axis) vs log(length) baseline (x-axis) across all evaluated models. The blue curve is the scores in UltraFeedback after KDE smoothing.}
    \label{fig:bestfit_length}
\end{figure}

\begin{figure}[htbp]
    \centering
 
    % Row 1
    \begin{subfigure}[t]{\mw\textwidth}
        \includegraphics[width=\linewidth]{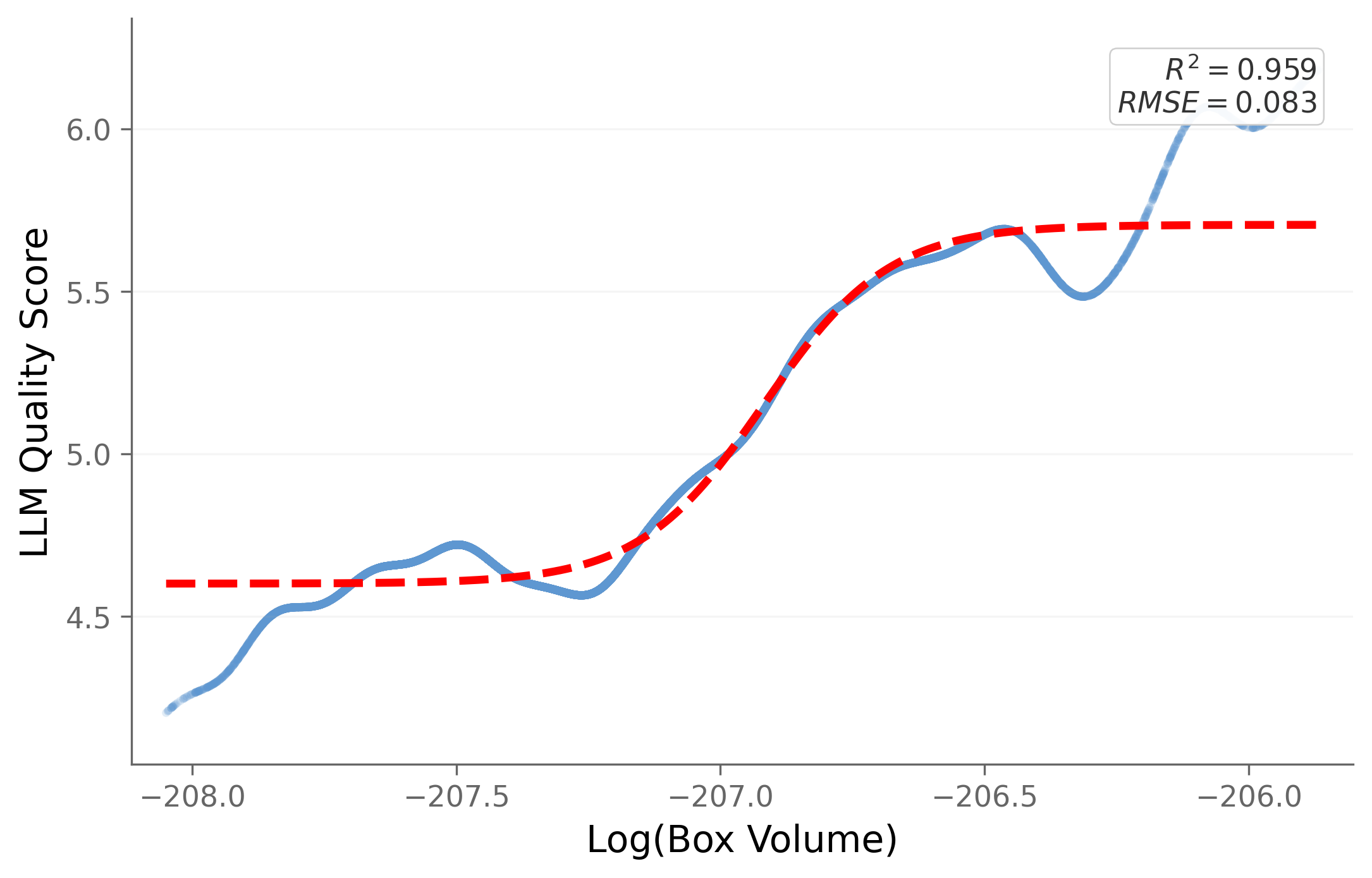}
        \caption{Alpaca-7B}
    \end{subfigure}\hfill
    \begin{subfigure}[t]{\mw\textwidth}
        \includegraphics[width=\linewidth]{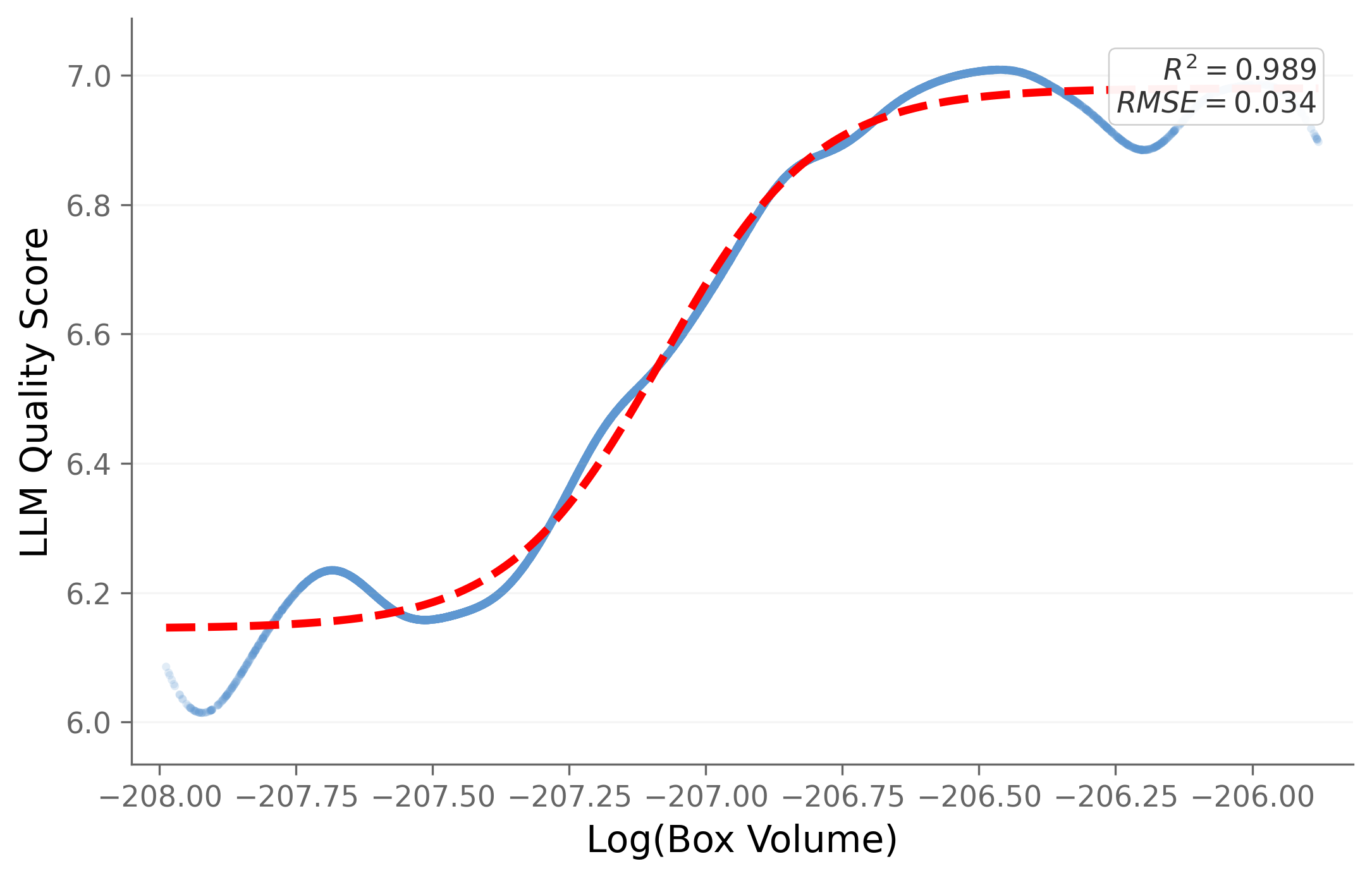}
        \caption{Bard}
    \end{subfigure}\hfill
    \begin{subfigure}[t]{\mw\textwidth}
        \includegraphics[width=\linewidth]{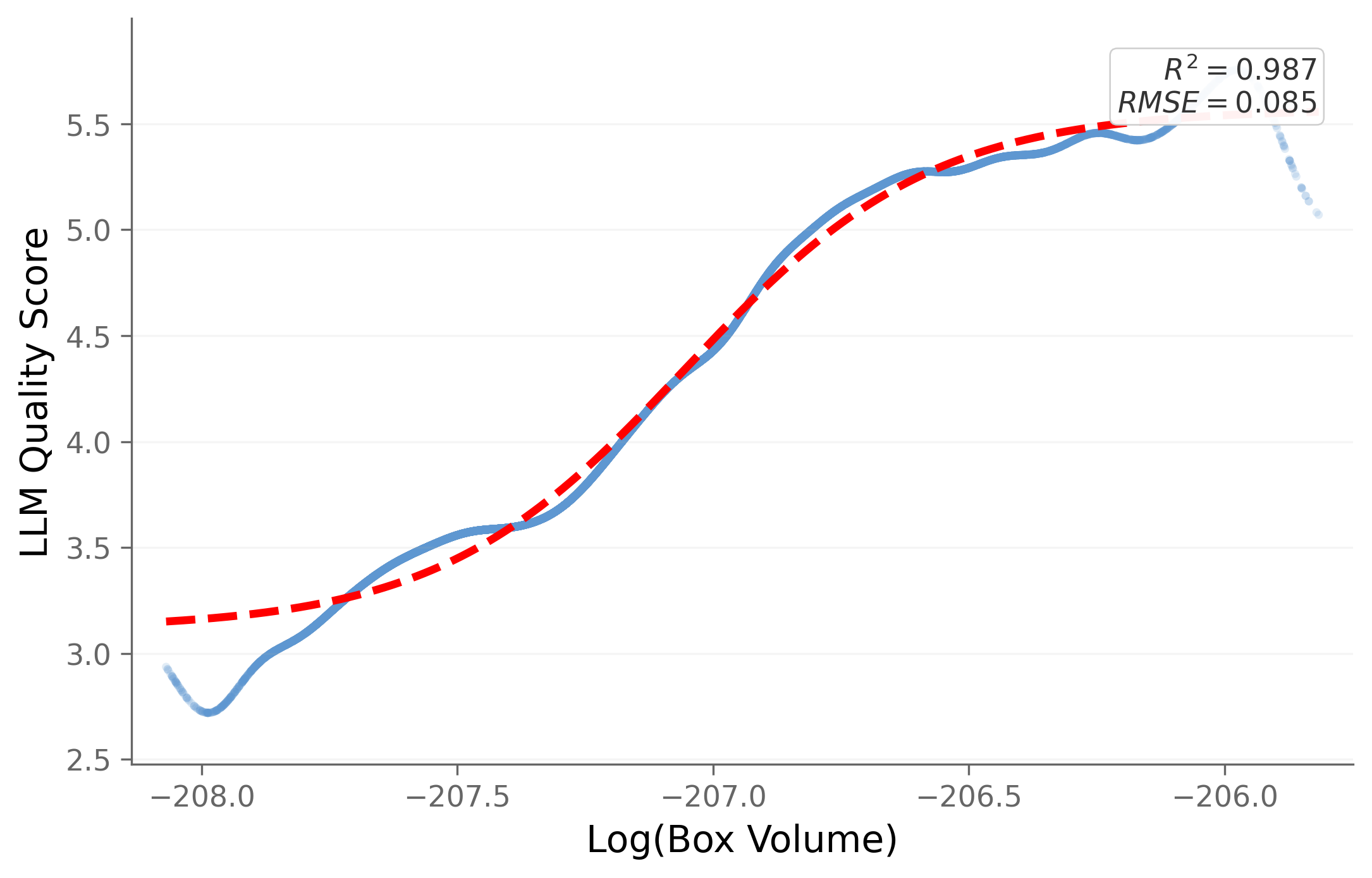}
        \caption{Falcon-40B-Instruct}
    \end{subfigure}\hfill
    \begin{subfigure}[t]{\mw\textwidth}
        \includegraphics[width=\linewidth]{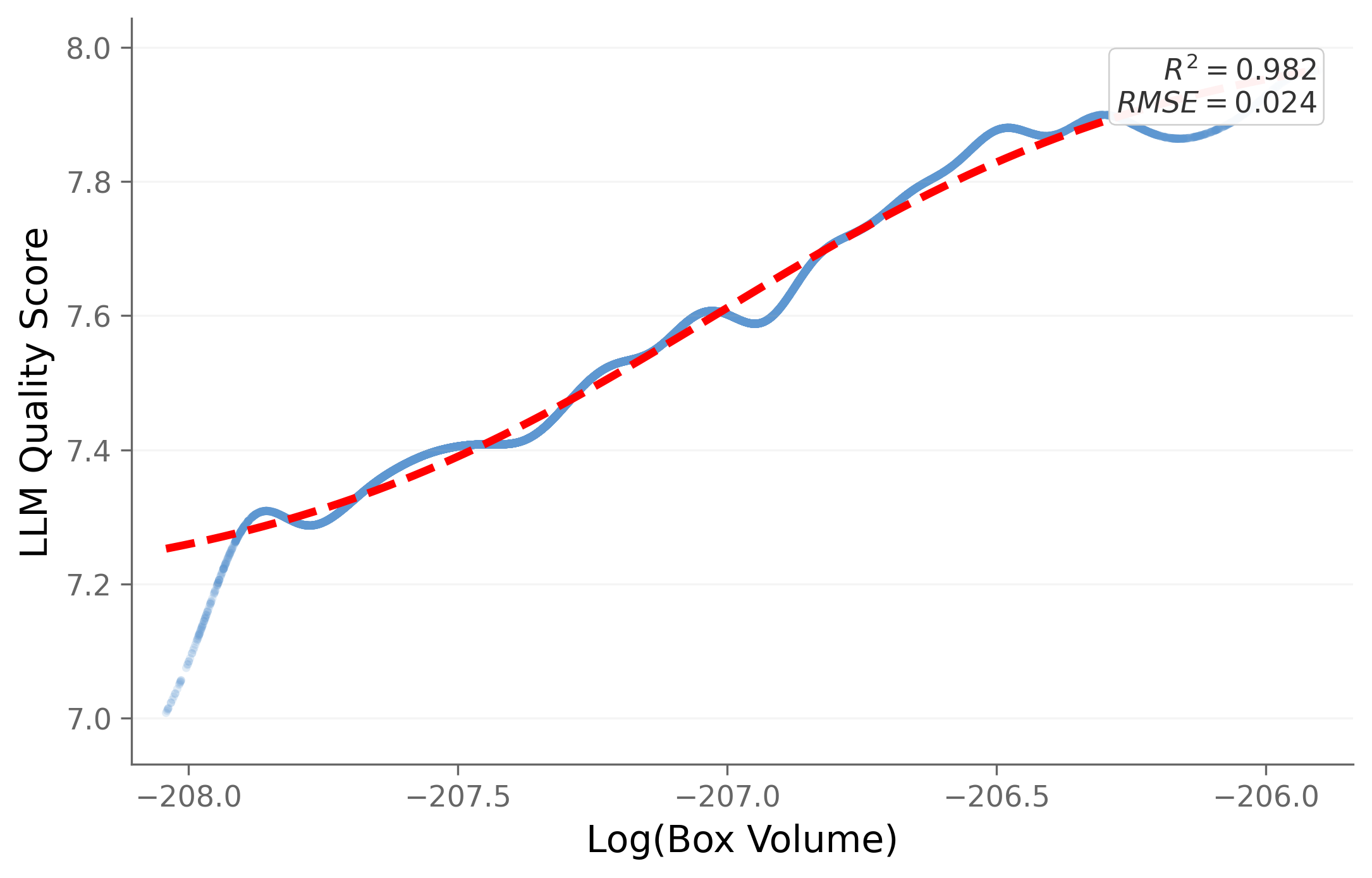}
        \caption{GPT-3.5-Turbo}
    \end{subfigure}
 
    \vspace{0.5em}
 
    % Row 2
    \begin{subfigure}[t]{\mw\textwidth}
        \includegraphics[width=\linewidth]{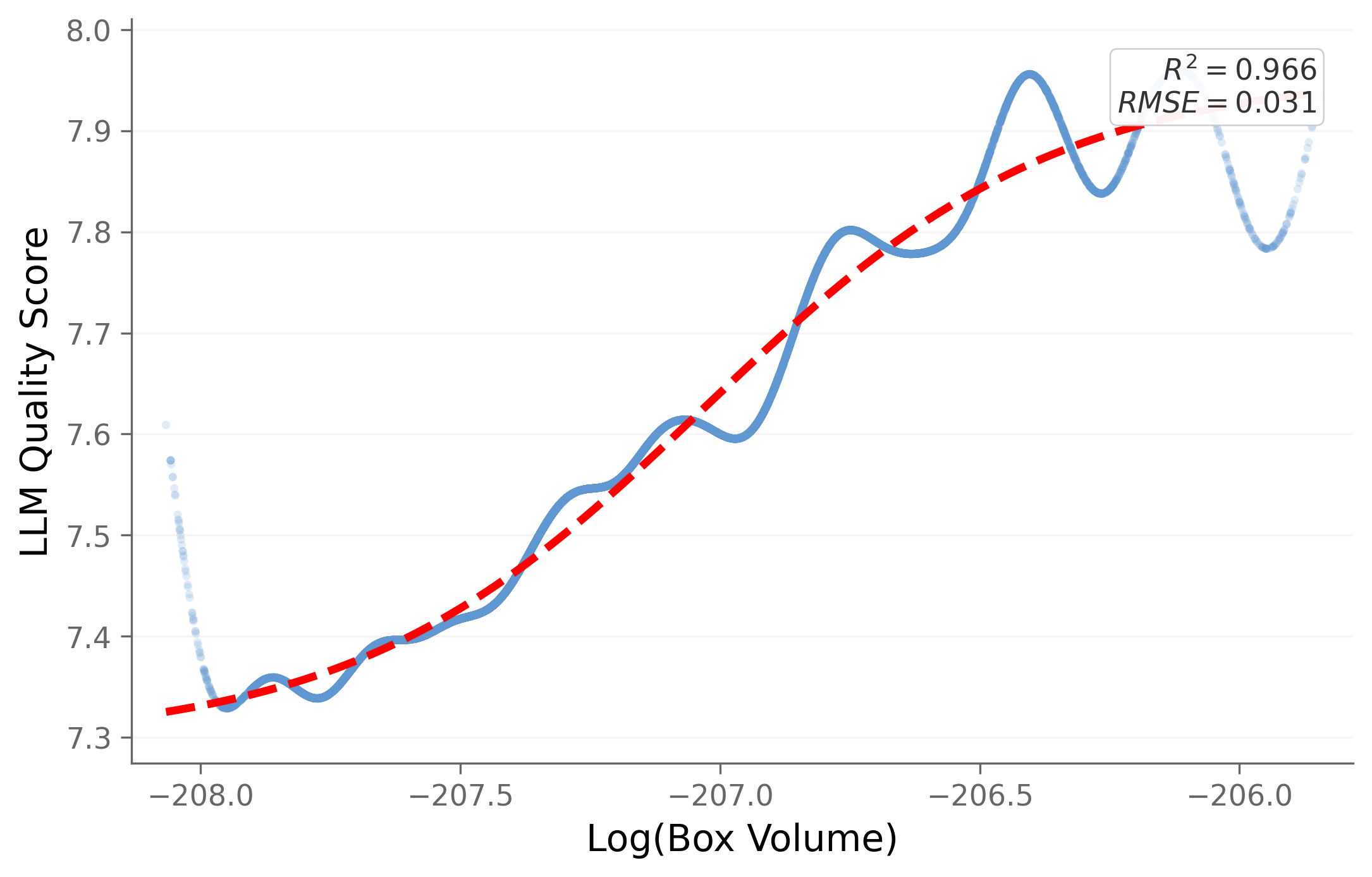}
        \caption{GPT-4}
    \end{subfigure}\hfill
    \begin{subfigure}[t]{\mw\textwidth}
        \includegraphics[width=\linewidth]{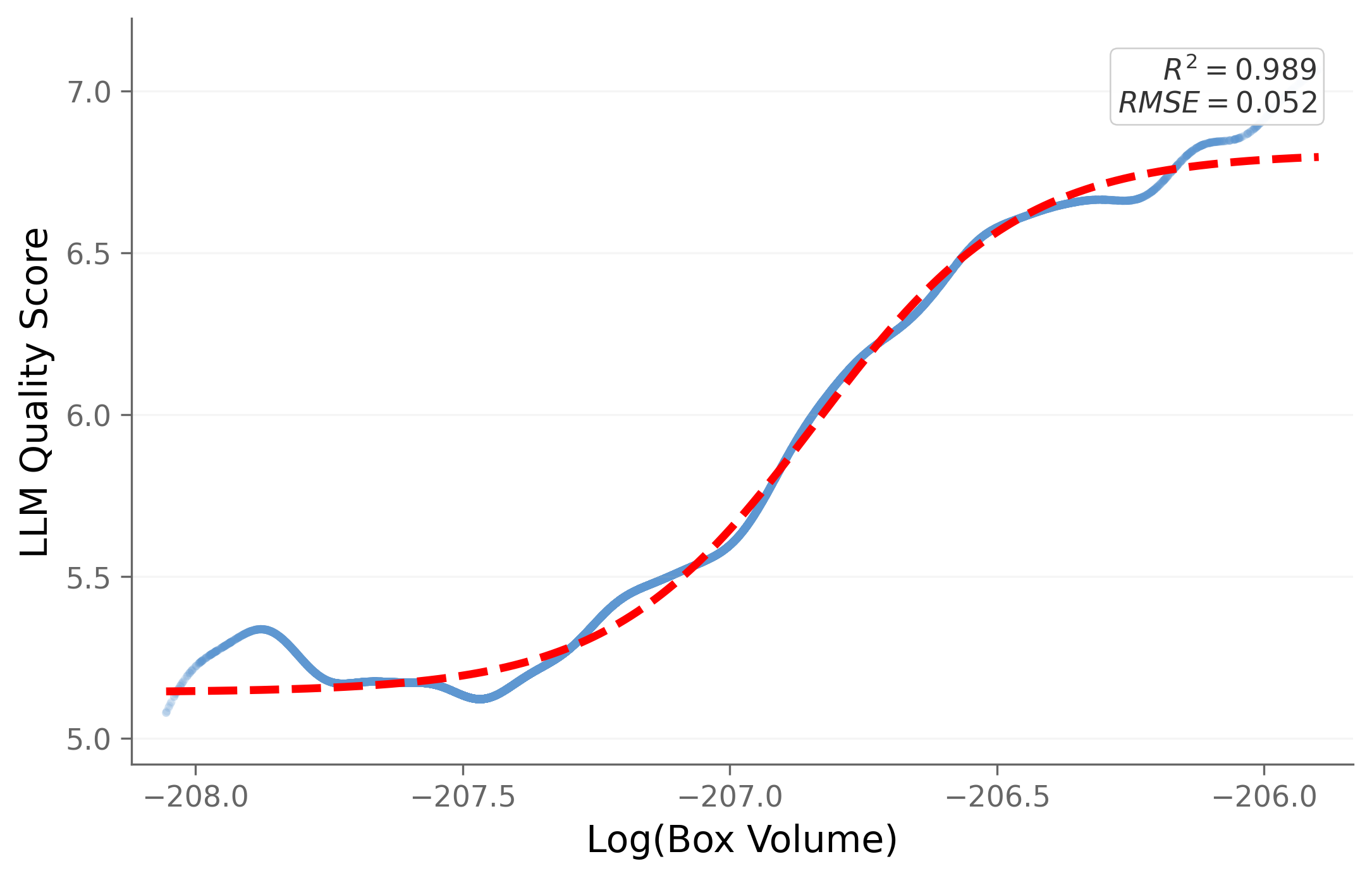}
        \caption{LLaMA-2-7B-Chat}
    \end{subfigure}\hfill
    \begin{subfigure}[t]{\mw\textwidth}
        \includegraphics[width=\linewidth]{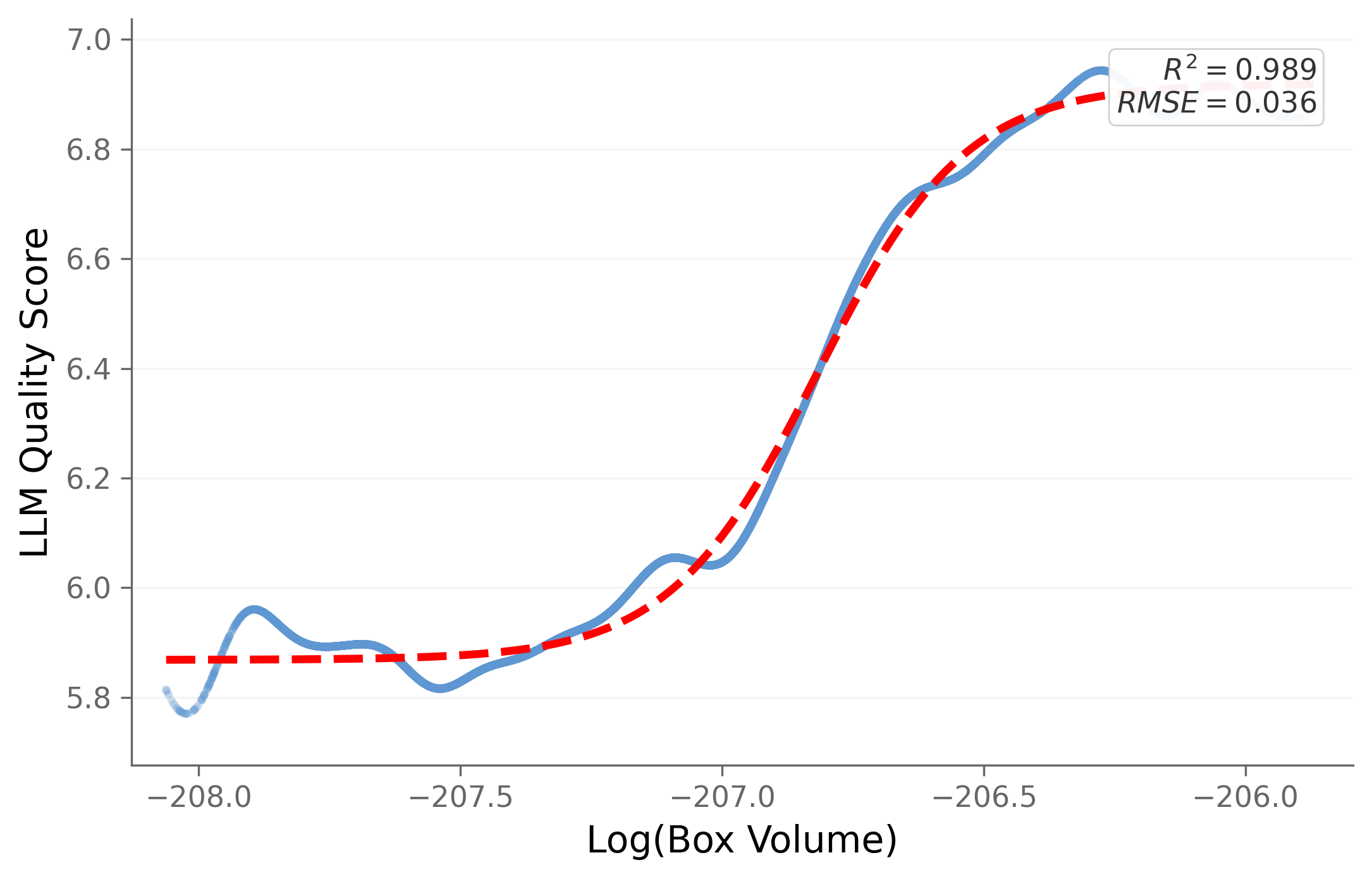}
        \caption{LLaMA-2-13B-Chat}
    \end{subfigure}\hfill
    \begin{subfigure}[t]{\mw\textwidth}
        \includegraphics[width=\linewidth]{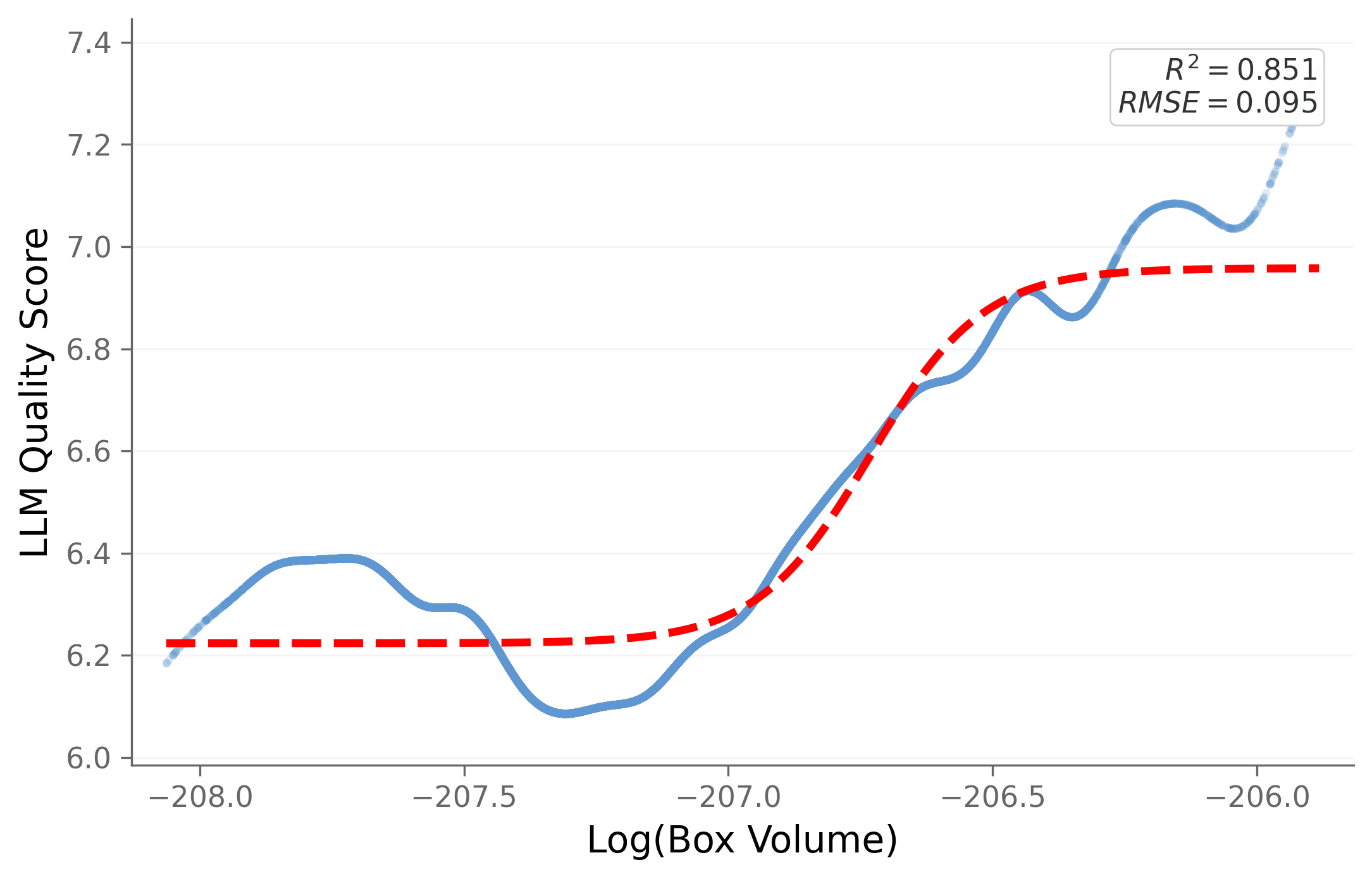}
        \caption{LLaMA-2-70B-Chat}
    \end{subfigure}
 
    \vspace{0.5em}
 
    % Row 3
    \begin{subfigure}[t]{\mw\textwidth}
        \includegraphics[width=\linewidth]{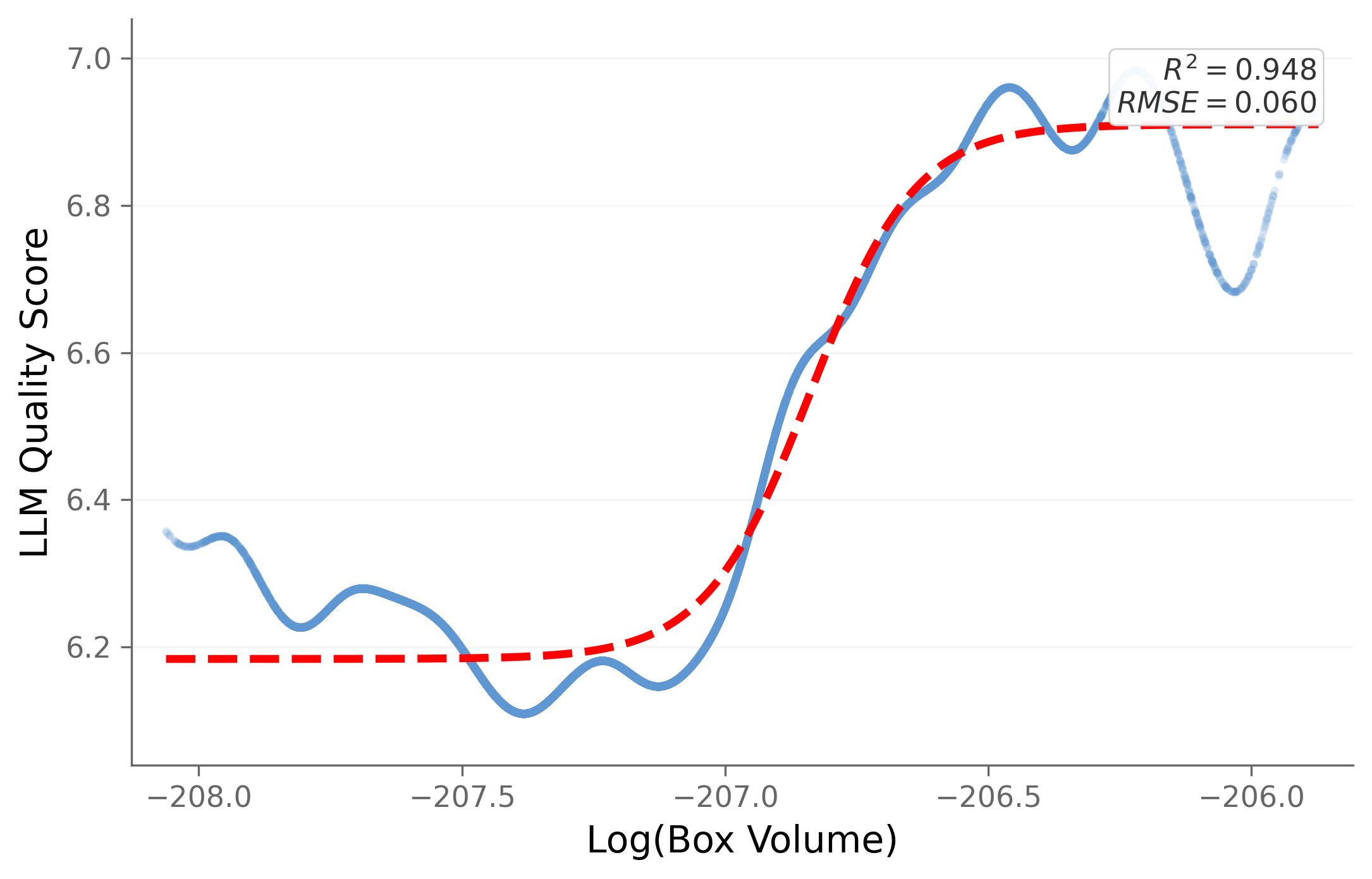}
        \caption{MPT-30B-Chat}
    \end{subfigure}\hfill
    \begin{subfigure}[t]{\mw\textwidth}
        \includegraphics[width=\linewidth]{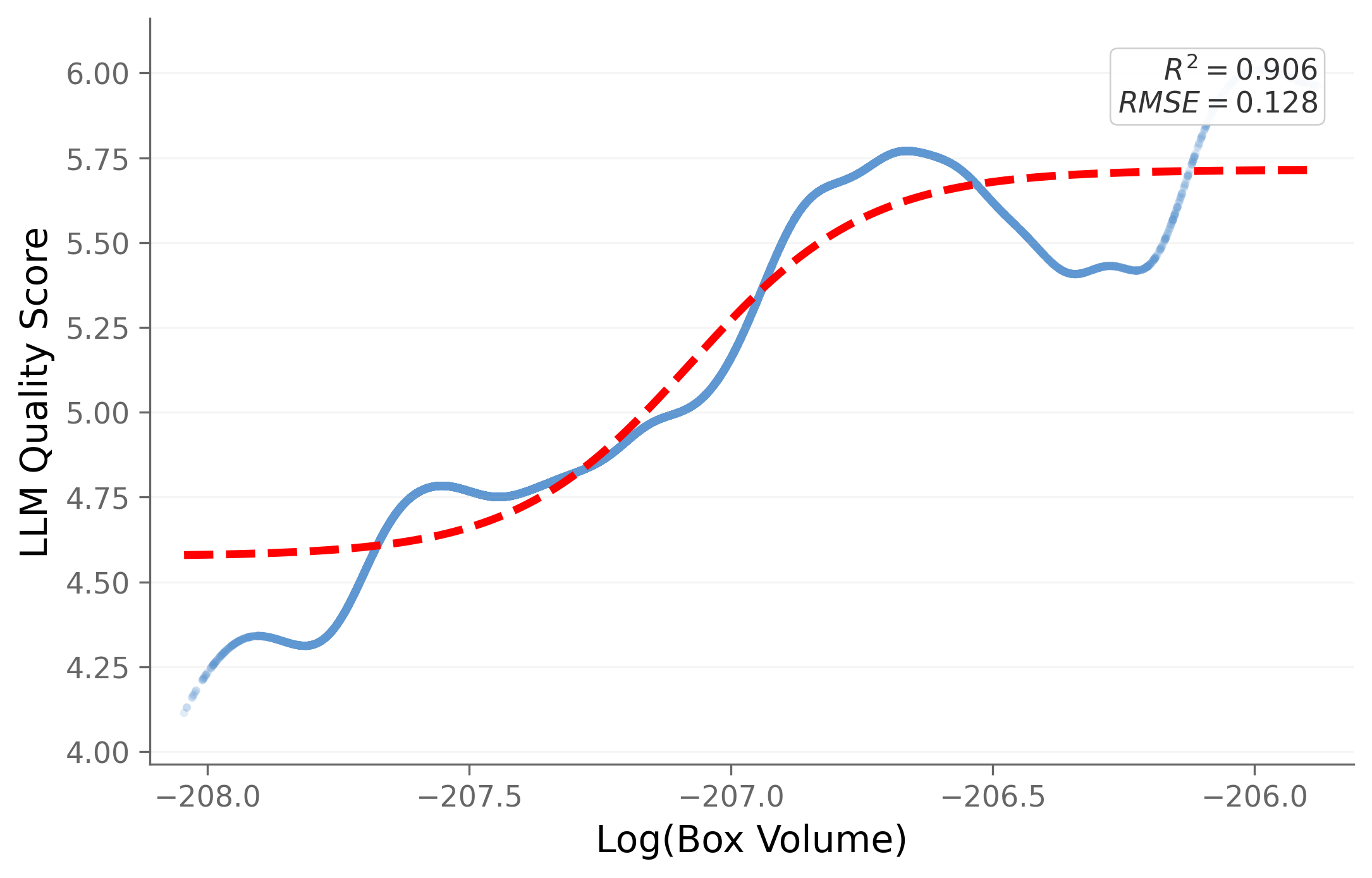}
        \caption{StarChat}
    \end{subfigure}\hfill
    \begin{subfigure}[t]{\mw\textwidth}
        \includegraphics[width=\linewidth]{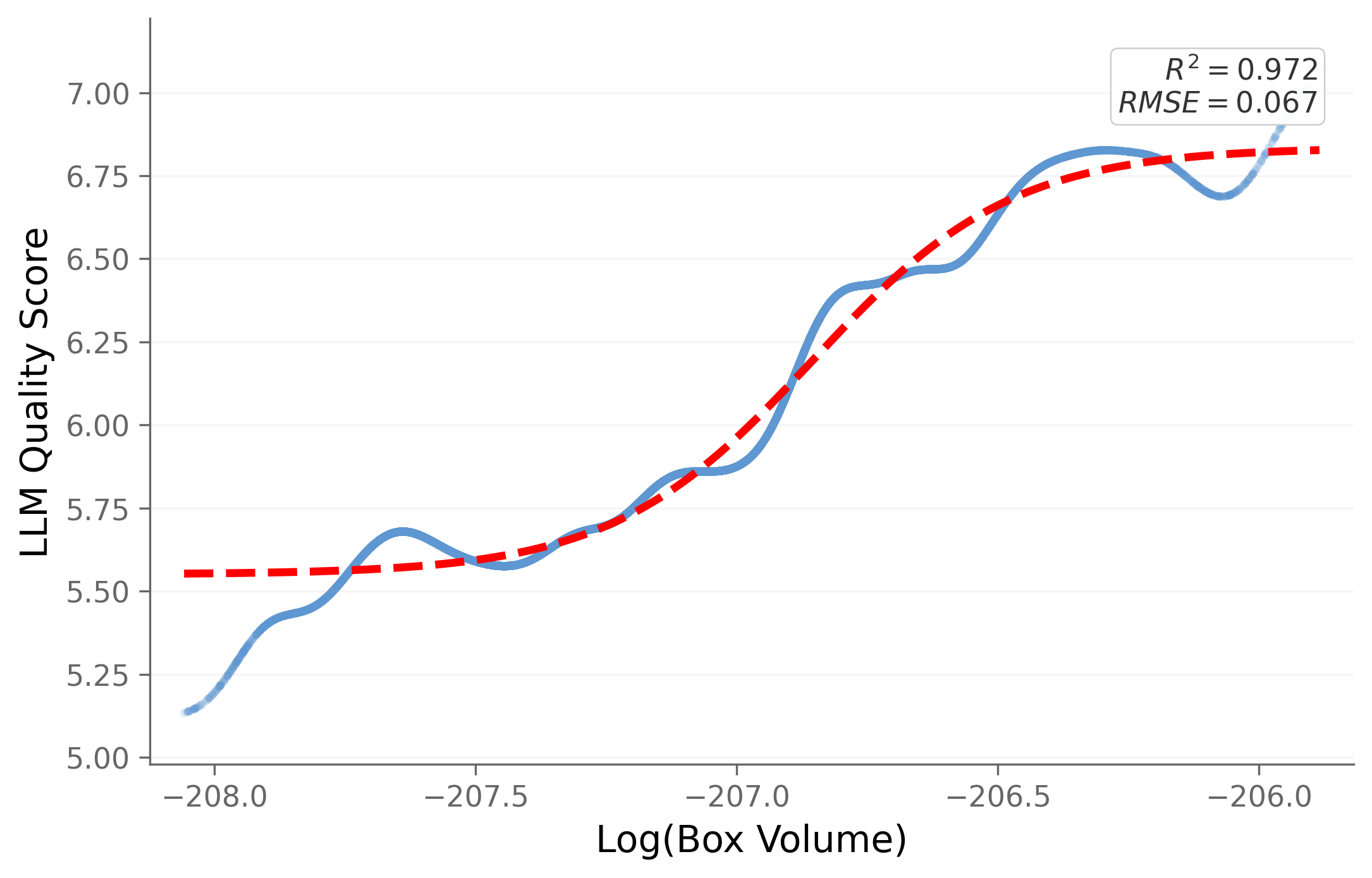}
        \caption{UltraLM-13B}
    \end{subfigure}
 
    \vspace{0.5em}
 
    % Row 4
    \begin{subfigure}[t]{\mw\textwidth}
        \includegraphics[width=\linewidth]{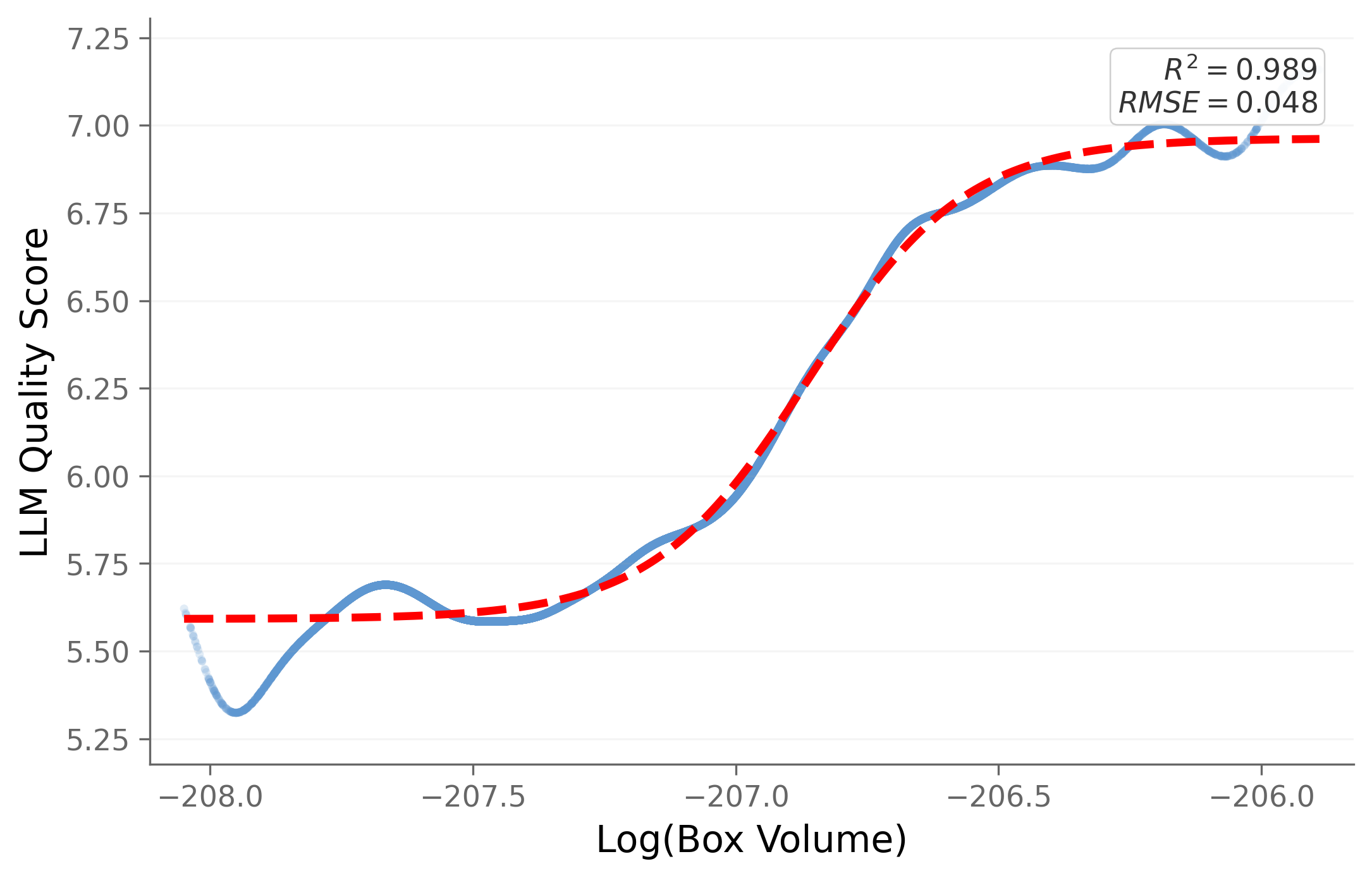}
        \caption{UltraLM-65B}
    \end{subfigure}\hfill
    \begin{subfigure}[t]{\mw\textwidth}
        \includegraphics[width=\linewidth]{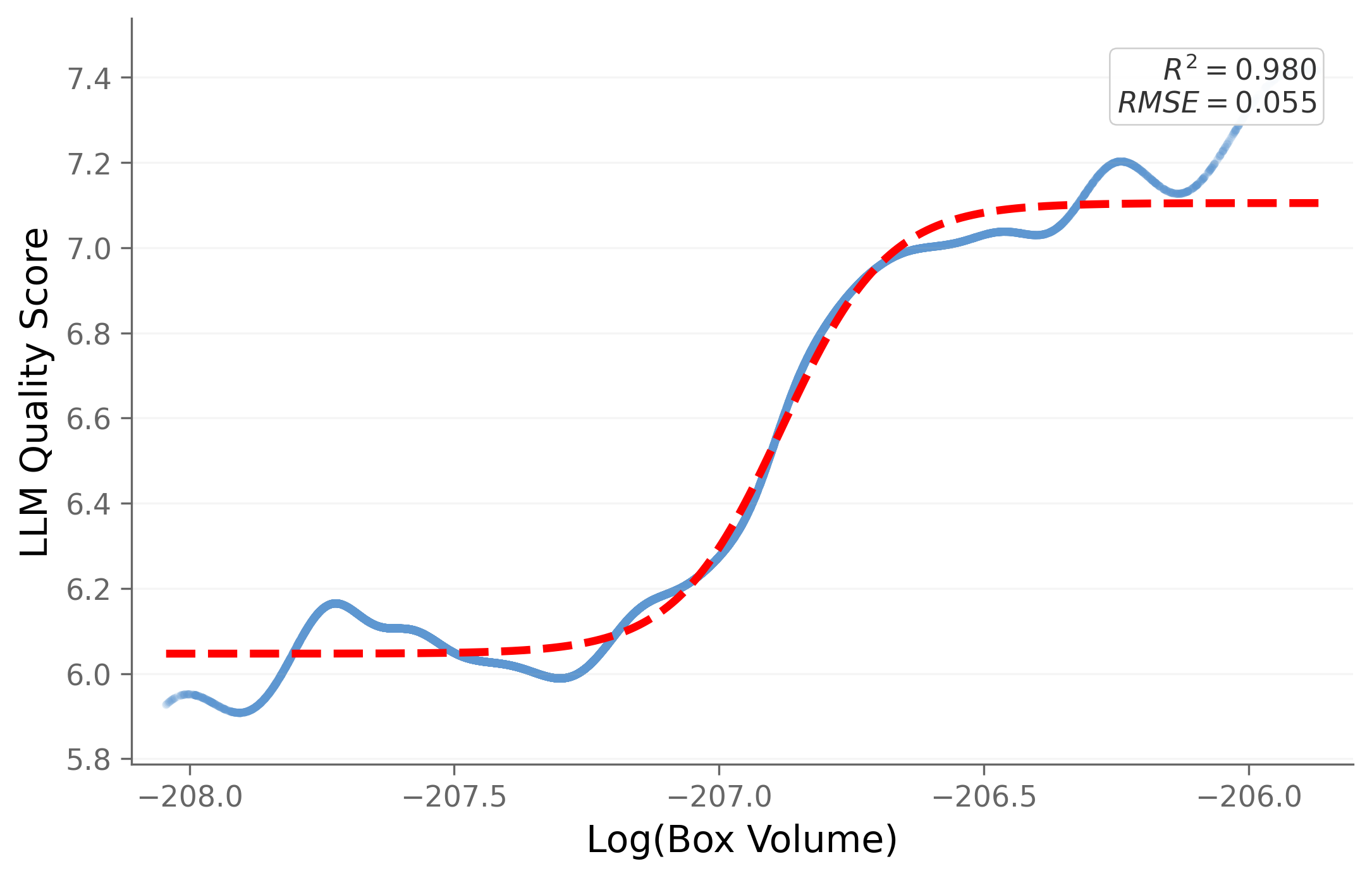}
        \caption{Vicuna-33B}
    \end{subfigure}\hfill
    \begin{subfigure}[t]{\mw\textwidth}
        \includegraphics[width=\linewidth]{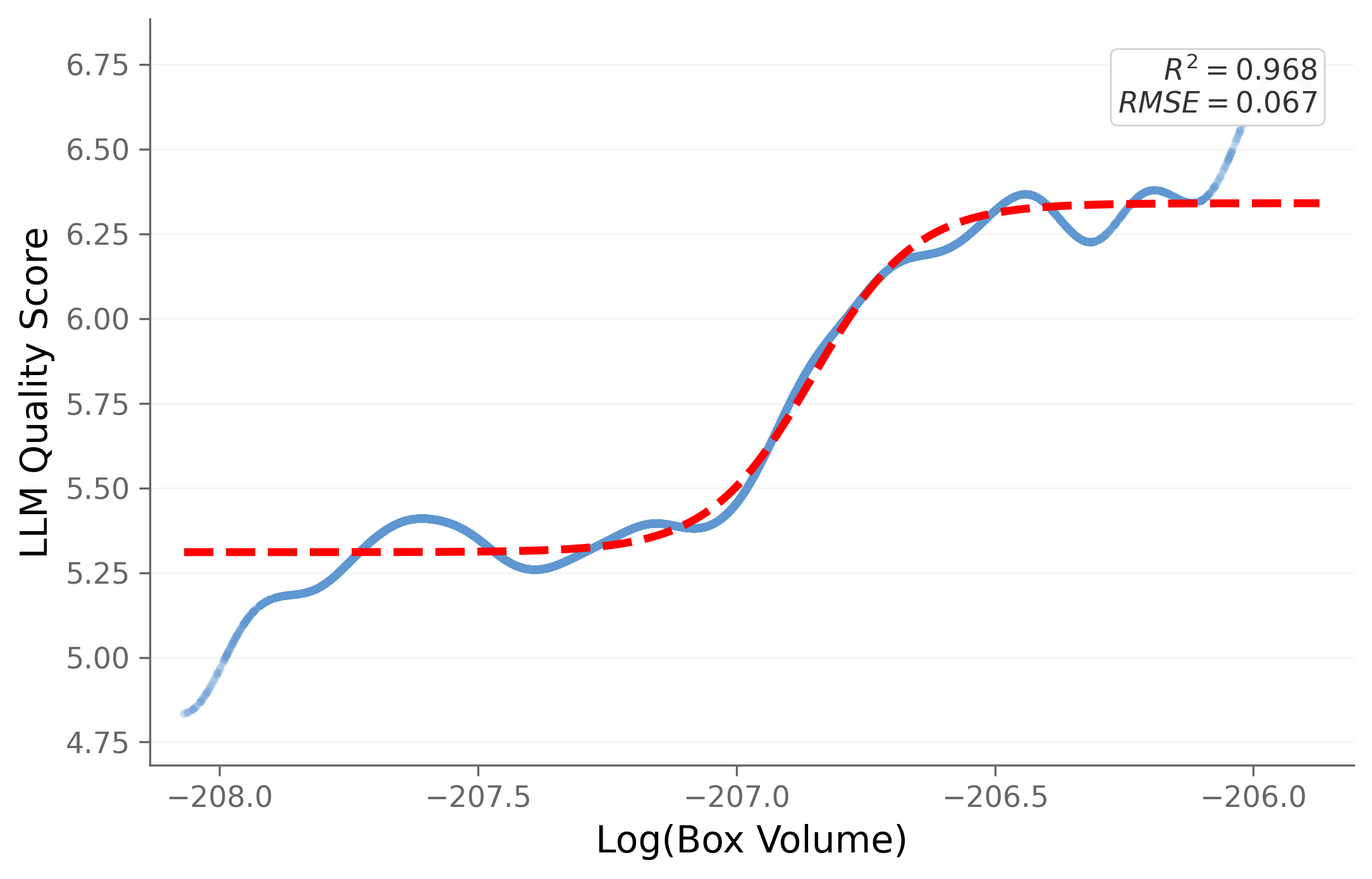}
        \caption{WizardLM-7B}
    \end{subfigure}\hfill
    \begin{subfigure}[t]{\mw\textwidth}
        \includegraphics[width=\linewidth]{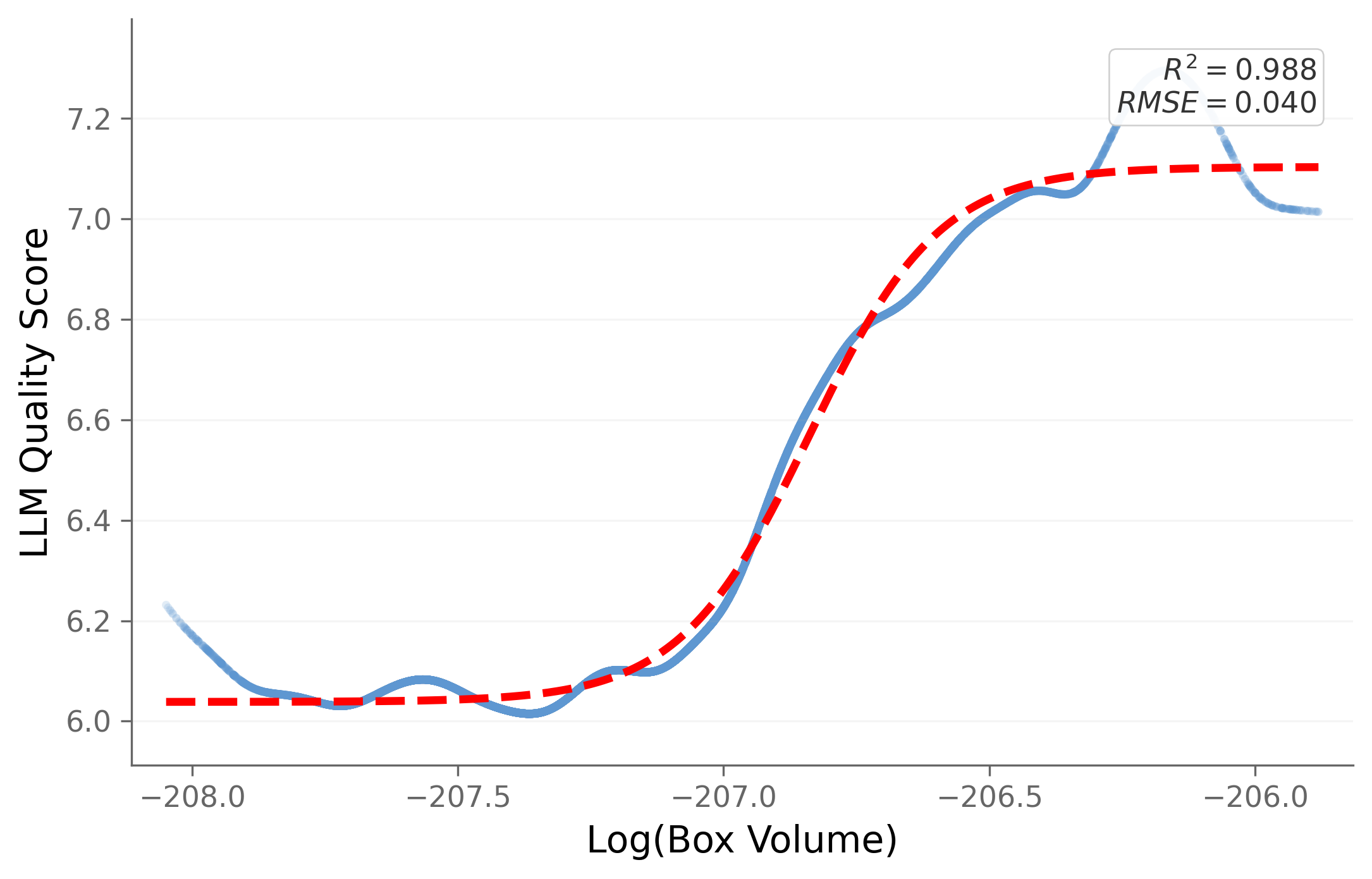}
        \caption{WizardLM-13B}
    \end{subfigure}
 
    \vspace{0.5em}
 
    % Row 5
    \begin{subfigure}[t]{\mw\textwidth}
        \includegraphics[width=\linewidth]{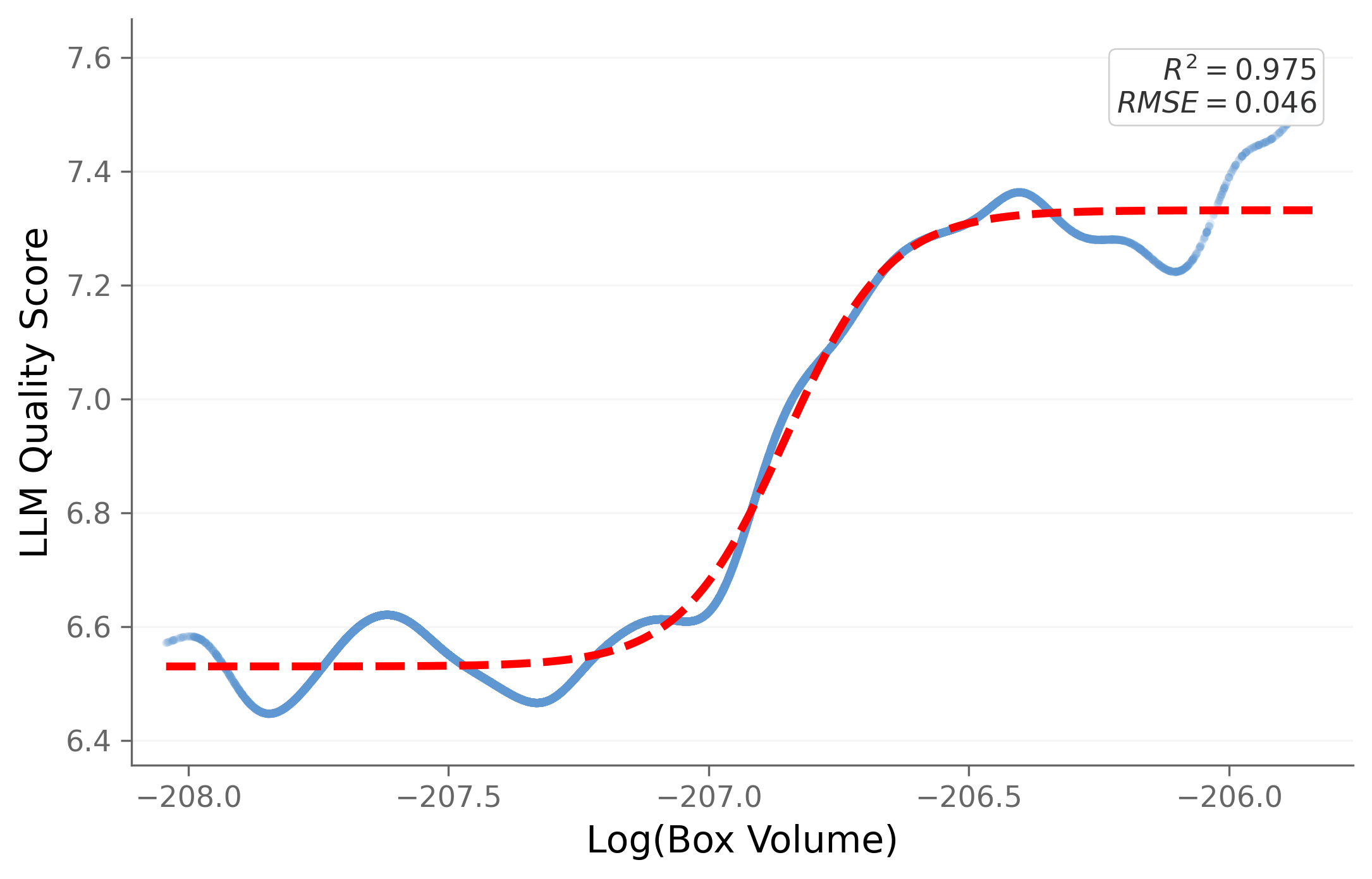}
        \caption{WizardLM-70B}
    \end{subfigure}
 
    \caption{Sigmoid best-fit curves for our \textbf{score} vs \textbf{log(box volume)} across all evaluated models in UltraFeedback.}
    \label{fig:bestfit_volume}
\end{figure}

\begin{table}[h]
\centering
\caption{RMSE comparison of volume-based versus length-based representation across LMSYS.}
\label{tab:rmse-diff}
\begin{tabular}{lccc}
\toprule
\textbf{Model} & \textbf{Vol RMSE} & \textbf{Len RMSE} & \textbf{Improvement} \\
\midrule
chatglm-6b                  & 0.2619 & 0.5363 & +51.2\% \\
oasst-pythia-12b            & 0.1513 & 0.2448 & +38.2\% \\
vicuna-13b                  & 0.1318 & 0.1984 & +33.6\% \\
fastchat-t5-3b              & 0.3518 & 0.4160 & +15.4\% \\
alpaca-13b                  & 0.1409 & 0.2032 & +30.7\% \\
wizardlm-13b                & 0.0824 & 0.1313 & +37.2\% \\
vicuna-33b                  & 0.1068 & 0.1550 & +31.1\% \\
mpt-7b-chat                 & 0.0851 & 0.1250 & +31.9\% \\
llama-13b                   & 0.1460 & 0.1753 & +16.7\% \\
koala-13b                   & 0.1288 & 0.1421 &  +9.4\% \\
stablelm-tuned-alpha-7b     & 0.0526 & 0.0642 & +18.1\% \\
llama-2-13b-chat            & 0.0769 & 0.0863 & +10.9\% \\
RWKV-4-Raven-14B            & 0.0652 & 0.0711 &  +8.3\% \\
dolly-v2-12b                & 0.0580 & 0.0558 &  -3.9\% \\
claude-1                    & 0.0801 & 0.0751 &  -6.7\% \\
vicuna-7b                   & 0.1507 & 0.1165 & -29.4\% \\
\midrule
\textbf{Average}            & \textbf{0.1325} & \textbf{0.1873} & \textbf{+18.3\%} \\
\bottomrule
\end{tabular}
\end{table}

\begin{table}[h]
\centering
\caption{RMSE comparison of volume-based versus length-based representation for ultrafeedback.}
\label{tab:rmse-diff2}
\begin{tabular}{lccc}
\toprule
\textbf{Model} & \textbf{Vol RMSE} & \textbf{Len RMSE} & \textbf{Improvement} \\
\midrule
wizardlm-13b         & 0.0404 & 0.1452 & +72.2\% \\
llama-2-7b-chat      & 0.0517 & 0.1393 & +62.9\% \\
starchat             & 0.1283 & 0.2082 & +38.4\% \\
mpt-30b-chat         & 0.0605 & 0.1345 & +55.0\% \\
bard                 & 0.0340 & 0.0989 & +65.6\% \\
vicuna-33b           & 0.0554 & 0.1176 & +52.9\% \\
wizardlm-70b         & 0.0462 & 0.1034 & +55.3\% \\
llama-2-13b-chat     & 0.0361 & 0.0859 & +58.0\% \\
llama-2-70b-chat     & 0.0951 & 0.1449 & +34.4\% \\
ultralm-13b          & 0.0670 & 0.1084 & +38.2\% \\
ultralm-65b          & 0.0481 & 0.0868 & +44.6\% \\
falcon-40b-instruct  & 0.0850 & 0.1139 & +25.4\% \\
wizardlm-7b          & 0.0666 & 0.0942 & +29.3\% \\
gpt-4                & 0.0308 & 0.0386 & +20.2\% \\
gpt-3.5-turbo        & 0.0238 & 0.0310 & +23.2\% \\
alpaca-7b        & 0.0830 & 0.1617 & +48.7\% \\
\midrule
\textbf{Average}     & \textbf{0.0595} & \textbf{0.1133} & \textbf{+45.3\%} \\
\bottomrule
\end{tabular}
\end{table}
%%%%%%%%%%%%%%%%%%%%%%%%%%%%%%%%%%%%%%%%%%%%%%%%%%%%%%%%%%%%%%%%%%%%%%%%%%%%%%%
%%%%%%%%%%%%%%%%%%%%%%%%%%%%%%%%%%%%%%%%%%%%%%%%%%%%%%%%%%%%%%%%%%%%%%%%%%%%%%%
%%%%%%%%%%%%%%%%%%%%%%%%%%%%%%%%%%%%%%%%%%%%%%%%%%%%%%%%%%%%%%%%%%%%%%%%%%%%%%%

\end{document}